\documentclass{article}
\usepackage{algorithmicx}
\usepackage[preprint]{iclr2027_conference}

\usepackage{times}
\usepackage{hyperref}
\usepackage{url}
\usepackage{graphicx}
\usepackage{subcaption}
\newcommand{\otherdomain}{MedBookVQA}
\usepackage{booktabs}
\usepackage{multirow}
\usepackage{amsmath}
\usepackage{amssymb}
\usepackage{amsthm}
\usepackage{enumitem}
\usepackage{xcolor}
\usepackage{algorithm}
\usepackage{algpseudocodex}
\usepackage{wrapfig}

\usepackage{xcolor}
\algrenewcommand{\algorithmiccomment}[1]{\hfill\textcolor{blue}{$\triangleright$ #1}}

\newtheorem{theorem}{Theorem}
\newtheorem{definition}{Definition}
\newtheorem{proposition}{Proposition}
\newtheorem{corollary}{Corollary}
\newtheorem{remark}{Remark}
\let\STATE\State

\setlist{itemsep=1pt, topsep=2pt, parsep=0pt}

\usepackage{fontawesome5}
\usepackage{lmodern}

\newcommand{\namel}{CFA}
\newcommand{\realname}{RDA}

\title{\namel: Mitigating Forgetting in RL Post-Training with Content-Free Anchors}

\author{
Ramnath Kumar\textsuperscript{1}\thanks{Work done while at Google.} \quad
Jiong Zhang\textsuperscript{2} \quad
Wei-Cheng Chang\textsuperscript{3} \quad
Fei Wang\textsuperscript{2} \\
\textbf{Hsiang-Fu Yu\textsuperscript{2}} \quad
\textbf{Inderjit Dhillon\textsuperscript{2,4}} \quad
\textbf{Cho-Jui Hsieh\textsuperscript{1,5}} \\[0.3em]
\textsuperscript{1}University of California, Los Angeles \quad
\textsuperscript{2}Google \quad
\textsuperscript{3}Google DeepMind \\
\textsuperscript{4}University of Texas, Austin \quad
\textsuperscript{5}Arena \\[0.3em]
\texttt{\{ramnathk, chohsieh\}@cs.ucla.edu}
}

\begin{document}

\maketitle

\begin{abstract}
RL post-training is widely assumed to forget little of what a model was not trained on. We show this assumption is incomplete. Across Qwen3-VL at 2B, 4B, and 8B scale, single-domain RL post-training silently collapses individual held-out capabilities, with worst-benchmark drops of up to 67\%, while suite-wide averages stay flat because unrelated benchmarks improve on the same runs. We introduce the worst-k curve to expose this tail. Existing defenses, such as parameter-importance penalties, replay, and KL regularization, assume that the anchor inputs used to constrain the model must represent the capabilities being protected, which is infeasible for general capability. We find this assumption is not load-bearing: anchors made of uniformly random tokens match, and at modest budgets outperform, a curated real coreset spanning five capability dimensions. We instantiate this as Content-Free Anchoring (CFA), a Fisher-weighted Maximum Mean Discrepancy penalty between current and reference next-token distributions in the model's embedding space, which we derive as a Gauss-Newton generalization of Elastic Weight Consolidation. We prove that anchors affect retention only through a transfer coefficient, and that anchor content can improve it over random anchors by at most a bounded factor that requires no held-out data. Across all nine domain-scale combinations, CFA limits worst-benchmark forgetting to between 0.1\% and 9.1\%, while every baseline loses at least 36\% on some benchmark. The result carries over unchanged to InternVL3.5 and Gemma 4, where CFA cuts the worst-benchmark drop of GSPO from 43-48\% to under 6\%, and CFA remains the most robust method on six additional out-of-domain benchmarks, all without any real, licensed, or curated data.
\end{abstract}

\section{Introduction}
\label{sec:intro}

RL post-training (RLHF, RLVR, and their variants) is the default recipe for adapting large language and vision-language models to new capabilities. It is usually deployed under an optimistic assumption: the tuned model forgets little of what it was not trained on. A growing body of work supports this view, arguing that on-policy RL stays close to the base policy in KL divergence~\citep{shenfeld2026rl}, that reward variance acts as an implicit regularizer~\citep{lai2025rft}, and that RL's mode-seeking dynamics preserve more of the prior task than SFT's mode-covering dynamics~\citep{chen2026retaining}. Together, these results suggest a rule of thumb: one can RL-tune a pretrained model on a new domain and ship it without worrying about catastrophic forgetting. \citet{luo2026rlforgets} complicate this picture. They argue that the benchmarks behind the optimistic literature are largely covered by pretraining, build a benchmark of recent and diverse tasks (MRCL), and show that RL post-training \emph{does} forget across a \emph{sequence} of tasks. Their fix, CPO, is a replay-free regularizer on the magnitude of parameter change that does not account for curvature or Fisher information.

We revisit the most common deployment pattern, post-training on a single new domain, where prior work predicts little forgetting. We RL-tune Qwen3-VL on three MRCL domains and evaluate on a common multimodal suite, and find the opposite: grounding and document understanding collapse. CountBenchQA, for instance, falls by up to $67\%$ relative to the base model, while STEM and reasoning benchmarks \emph{improve on the same runs}. A suite-wide average cancels the two and suggests little forgetting, which hides the damage (Section~\ref{sec:phenomenon}).

\textbf{Motivation.} Every existing defense against this kind of forgetting, including parameter-importance penalties (EWC; \citealp{kirkpatrick2017overcoming}), replay buffers~\citep{rolnick2019experience}, and KL regularization on reference prompts, shares an assumption about the \emph{anchor set}: the reference inputs used to constrain the model must represent the capabilities being protected, whether as a buffer of prior-task examples, a set of inducing points, or domain-relevant prompts. This assumption is easy to state and hard to satisfy, for two reasons of increasing severity. The first is cost: sourcing, licensing, and maintaining real data for even a handful of known capabilities is expensive, and grows with the number of protected domains. The second is scope. The goal is not to protect a few domains we happened to anticipate, but the model's \emph{general} capability, because an RL run optimized for one narrow reward can degrade competencies far outside its training distribution, and there is no way to know in advance which of them a given run will damage. Once the goal is general capability protection, a real anchor set stops being merely expensive and becomes infeasible by construction: the only anchor set guaranteed to cover whatever capability turns out to be at risk is one that spans essentially the model's entire pretraining distribution. No practitioner curates, licenses, and replays pretraining-scale data on every RL run, and the natural shortcut, anchoring on the evaluation suite itself, invites contamination and reward hacking. What is needed is a way to constrain a model's outputs relative to its own prior behavior, at the scale that general capability protection demands, without real data at that scale.

\begin{wrapfigure}{r}{0.50\linewidth}
\centering
\vspace{-12pt}
\includegraphics[width=\linewidth]{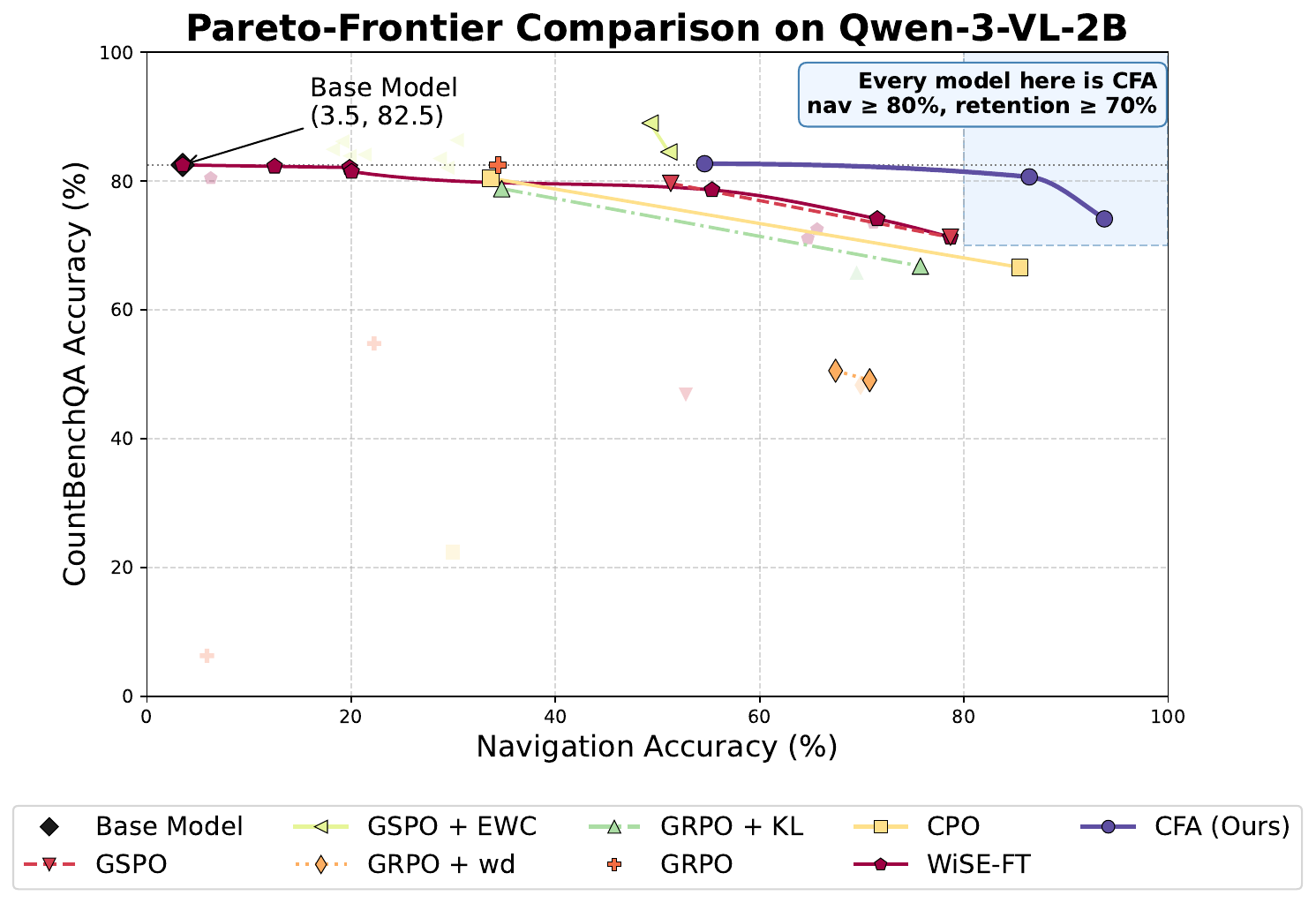}
\caption{\textbf{Pareto frontiers, in-domain vs.\ OOD retention.} Navigation accuracy vs.\ CountBenchQA accuracy, Qwen3-VL-2B. \namel{} reaches $\geq80\%$ Navigation while retaining $\geq70\%$ CountBenchQA, a region unavailable to other methods.}
\label{fig:pareto-intro}
\vspace{-8pt}
\end{wrapfigure}

\textbf{Overview.} We test whether curated anchors are load-bearing and find that they are not. Content-free anchors, random token sequences with no connection to any capability, trail a curated real coreset at small budgets but match or outperform it once given a modest budget (Section~\ref{sec:discussion-ablations}). We instantiate this as Content-Free Anchoring (\namel): a frozen reference policy is evaluated once on a fixed anchor set, and the current policy is penalized for diverging from it. The divergence is a Fisher-weighted Maximum Mean Discrepancy computed in the model's own token-embedding space rather than discrete KL over the raw vocabulary, since KL treats every token as an unrelated class and penalizes a benign paraphrase as heavily as a semantic error. To expose the tail that averages hide, we report the \emph{worst-$k$ curve}, the mean relative change over the $k$ most-affected out-of-domain benchmarks. Before scaling up, we validate both design choices, that functional anchoring is necessary and that anchor content is not, in a minimal, fully enumerable testbed (Section~\ref{sec:toy-main}). At scale, \namel{} keeps worst-benchmark forgetting under $10\%$ in all nine domain-scale combinations of Qwen3-VL (Section~\ref{sec:results-main}), and the result carries over beyond this grid: \namel{} again has the smallest worst-case forgetting on InternVL3.5-1B and Gemma~4~E2B, cuts SFT's worst-benchmark drop from $78.8\%$ to $9.1\%$, reduces the already small forgetting of four text-only models, and remains the most robust method on six additional out-of-domain benchmarks (Section~\ref{sec:other-families}). Figure~\ref{fig:pareto-intro} previews the resulting trade-off on Navigation at 2B: \namel{} attains both higher in-domain accuracy and higher retention than every baseline family we compare against, including weight interpolation, parameter-importance penalties, and KL regularization.

\textbf{Contributions.} We make the following contributions:
\begin{itemize}
\item \textbf{A hidden failure mode in the default RL post-training recipe.} Single-domain RL post-training silently collapses held-out capabilities while suite-wide averages hide it. We propose the \emph{worst-$k$ curve}, an empirical analogue of Conditional Value-at-Risk over out-of-domain benchmarks, as a replacement metric (Section~\ref{sec:metric}).
\item \textbf{A theory of what an anchor set must provide.} We unify EWC and \namel{} under a single Gauss-Newton template, prove that anchors protect a held-out capability exactly as far as a transfer coefficient allows, bound the most that anchor content can add over random anchors, derive a checkable budget crossover and the limits of random anchors, and give an explicit trust radius for EWC's second-order truncation (Section~\ref{sec:theory}, Appendix~\ref{app:proofs}).
\item \textbf{A data-free regularizer that outperforms curated alternatives.} Content-free random-token anchors match or outperform a curated real coreset once given a modest budget. The resulting method, \namel{}, substantially outperforms GRPO and GRPO+KL~\citep{shao2024deepseekmath}, GSPO~\citep{zheng2025groupsequencepolicyoptimization}, EWC~\citep{kirkpatrick2017overcoming}, and CPO~\citep{luo2026rlforgets} on worst-$k$ forgetting at comparable in-domain gain, keeping worst-case forgetting under $10\%$ in every domain and scale tested (Section~\ref{sec:results-main}).
\item \textbf{Evidence of generality.} The same configuration transfers unchanged to two further VLM families, to SFT, to four text-only models, and to six out-of-domain benchmarks outside the original suite (Section~\ref{sec:other-families}).
\end{itemize}

\section{Related Work}
\label{sec:related}

\textbf{Forgetting in RL post-training.} Catastrophic forgetting is a
long-standing continual-learning concern
\citep{mccloskey1989catastrophic,french1999catastrophic,kirkpatrick2017overcoming}
that has recently reappeared in LLM post-training
\citep{luo2023empirical,li2024revisiting,scialom2022continual}. A separate
thread reports RL as more conservative than SFT at matched
target-task performance, attributing this to low-KL on-policy optimization
\citep{shenfeld2026rl}, on-policy data \citep{chen2026retaining}, or RL's
mode-seeking dynamics \citep{lai2025rft} making average-case claims over target
tasks plausibly already seen during pretraining. We ask whether that average
conceals selective, worst-case collapse when the training domain is
genuinely unseen, in the spirit of \citet{luo2026rlforgets}'s critique. The
closest comparison is \citet{luo2026rlforgets}, which studies
\emph{inter-task} forgetting across a sequence of multimodal tasks and
proposes CPO, a replay-free parameter-movement regularizer. Our setting is
\emph{domain-to-general}: a single RL run with no task sequence, so a
capability can be damaged with no evident connection to the training
domain. CPO constrains parameter movement as a proxy for behavioral drift;
\namel{} instead evaluates the current and reference policies directly on
anchor inputs, targeting functional equivalence on held-out anchors while
learning the new task. Related mitigations include proactive pre-hardening
against anticipated shifts (FRPO; \citealp{sabbaghi2026frpo}) and sequential
preference regularization requiring stored historical query buffers (COPR;
\citealp{zhang2024copr}). Two further paradigms are widely deployed in
practice: mixing in pretraining or instruction-token loss during policy
gradient updates (PPO-ptx; \citealp{ouyang2022training}), which is effective
but requires complex data streaming pipelines; and post-hoc weight
interpolation between the finetuned and reference parameters
\citep{wortsman2022robust,ilharco2023editing,yadav2023ties,yu2024dare}, which
is training-free but assumes linear weight-space paths preserve non-linear
representation manifolds, often trading away target-task gains to recover
general skills. Most of these, like EWC rely on a second-order parameter-space truncation; we show that what an anchor set must supply is coverage of the directions post-training moves along, measured by a transfer coefficient, rather than semantic relevance to the protected domain.

\textbf{Classical continual learning and data-free distillation.}
Parameter-importance penalties \citep{kirkpatrick2017overcoming,
zenke2017continual}, gradient-projection methods
\citep{lopezpaz2017gem,chaudhry2019agem,farajtabar2020orthogonal,
saha2021gradient}, and replay of stored or generated examples
\citep{rolnick2019clear,buzzega2020derpp,scialom2022continual} all protect
prior behavior in some form; dark-experience replay \citep{buzzega2020derpp},
which stores soft targets from an earlier reference model, is closest in
spirit to our use of reference outputs. Data-free distillation similarly
transfers a teacher's behavior without original training data, but by
\emph{optimizing} synthetic inputs to be informative matching activation
statistics \citep{lopes2017datafree}, targeting class-conditional outputs
\citep{nayak2019zero}, or inverting a classifier \citep{deepinversion2020}.
All of these assume the input side carries information about the protected
distribution, an assumption that breaks down for a deployed model whose
prior capabilities are broad, proprietary, and impossible to enumerate in
advance. \namel's anchors are neither drawn from nor optimized toward any such
distribution: an anchor supplies only a \emph{location} at which to evaluate, and it is the frozen reference policy's output at that location that carries the behavior signal
(Section~\ref{sec:results}, Appendix~\ref{sec:toy-anchors}).

\section{Content-Free Anchoring: Method and Theory}
\label{sec:prelim}

\subsection{Setup and Notation}
\label{sec:notation}

We consider an autoregressive vision-language policy $\pi_\theta$ over a
vocabulary of size $V$, with token-embedding matrix $E \in
\mathbb{R}^{V\times d}$. Given a context $c=(I,x_{<t})$, $\pi_\theta(\cdot\mid c) \in \Delta^{V-1}$
denotes the next-token distribution. We write $\theta^*$ for the frozen
reference checkpoint (the model before RL post-training) and
$\pi^*(\cdot\mid c) := \pi_{\theta^*}(\cdot\mid c)$ for its induced
distribution. RL post-training optimizes $\theta$ from $\theta^*$ via a
policy-gradient objective $\mathcal{L}_{\text{RL}}$ (GRPO or GSPO) on a single target domain. Section~\ref{sec:theory} shows that forgetting tracks $\pi_\theta$'s induced behavior rather than the magnitude of parameter movement; we therefore constrain the output distribution directly rather than through a proxy
on $\theta$. Doing so requires (i) a reference distribution to compare against and (ii) a set of inputs on which to make that comparison. The
reference is simply the frozen pretrained policy $\pi^*$, already resident in memory for standard KL-regularized RL; Section~\ref{sec:anchor-construction}
addresses the harder question of what the input set should be. We formalize
the resulting mechanism as {Content-Free Anchoring (\namel)}: a fixed
anchor set is evaluated once under $\pi^*$, and $\pi_\theta$ is penalized
for diverging from the cached reference output.

\subsection{Measuring Forgetting: The Worst-k Curve}
\label{sec:metric}

Let $b\in\mathcal{B}$ index out-of-domain benchmarks and $\Delta_b =
(s_b^{\text{tuned}} - s_b^{\text{base}})/s_b^{\text{base}}$ be relative
change in score. The standard reported quantity is the suite-wide mean
$\bar\Delta = \frac{1}{|\mathcal{B}|}\sum_b \Delta_b$. We instead report
the \emph{worst-$k$ curve}
\begin{equation}
\label{eq:worstk}
\Delta_k \;=\; \frac{1}{k}\sum_{b\in\mathcal{B}_k}\Delta_b
       \;=\; \min_{\mathcal{S}\subseteq\mathcal{B},\;|\mathcal{S}|=k}\;
             \frac{1}{k}\sum_{b\in\mathcal{S}}\Delta_b, \qquad
\mathcal{B}_k = \text{the } k \text{ benchmarks with smallest } \Delta_b,
\end{equation}
swept over $k=1,\dots,|\mathcal{B}|$, so $\Delta_{|\mathcal{B}|} = \bar\Delta$
recovers the usual average at the right endpoint. Because $\mathcal{B}_k$
dynamically selects the $k$ most degraded benchmarks per method, the
worst-$k$ curve is an empirical extreme-value risk measure directly
analogous to Conditional Value-at-Risk (CVaR): the left endpoint ($k=1$)
diagnoses worst-case localized collapse, while the curvature and slope
measure how concentrated that collapse is across capabilities. The selected
sets are nested, $\mathcal{B}_1\subset\mathcal{B}_2\subset\cdots\subset
\mathcal{B}_{|\mathcal{B}|}=\mathcal{B}$, so $\Delta_k$ is non-decreasing in
$k$ by construction. Additional information about our metric of study and the motivation behind the same is provided in Appendix~\ref{app:metric}. 

\subsection{Anchor Set Construction}
\label{sec:anchor-construction}

We construct two anchor sets of fixed cardinality $N$: a curated set of
real image-context pairs, and a content-free set with no curation at all.

\textbf{Real anchors.} The candidate pool is assembled exclusively from
\emph{training} splits of public datasets spanning five capability
dimensions, so no evaluation benchmark can leak into the regularizer. A
size-$N$ subset is selected by greedy $k$-center traversal over a frozen
visual encoder, giving a 2-approximation to the optimal $N$-point covering
radius \citep{gonzalez1985clustering}; full pool and algorithm details are
in Appendix~\ref{app:anchors}. Because the traversal is greedy, the sets
used across budgets form a single nested sequence with non-increasing
covering radius rather than independent draws. This is a 
\emph{non}-i.i.d.\ construction, which is why the real anchor set requires its own guarantee (Proposition~\ref{prop:coreset-covering}) rather than the
i.i.d.\ argument covering content-free anchors.

\textbf{Content-free anchors.} These consist of synthetic token sequences drawn
i.i.d.\ and uniformly from the base vocabulary, at sequence length
$T=2048$, with no prompt engineering (full generation details in
Appendix~\ref{app:anchors}). Because this is an exact i.i.d.\ draw from a fixed distribution, the concentration guarantees of Section~\ref{sec:theory} apply directly. The anchors are text only and never pass through the vision encoder (Appendix~\ref{sec:limitations}).

\subsection{\namel{} Objective}
\label{sec:anchor-objective}

Let $\phi(v) := e_v/\sqrt{d}$ be the feature map induced by the model's own
token-embedding matrix, and $K:=\frac1d EE^\top$. Given an anchor set
$\{c^{(i)}\}_{i=1}^N$ of fixed contexts specified at construction time, we
evaluate the frozen reference policy in a single teacher-forced pass once
before training. Rather than caching dense categorical distributions, we cache
only the $d$-dimensional expected feature vectors $\mu_t^{*(i)} :=
\frac{1}{\sqrt d}E^\top\pi^*(\cdot\mid c^{(i)}_{<t})$ and scalar weights
$\tilde w_t^{(i)}$, requiring only $\mathcal{O}(NTd)$ storage. During training we penalize the active
policy for diverging from the cached reference embeddings:
\begin{equation}
\mathcal{L}_{\text{anchor}} = \frac{\lambda}{BT}\sum_{i=1}^{B}\sum_{t=1}^{T}
\tilde w_t^{(i)} \cdot \mathrm{MMD}_\phi^2\!\left(
\pi_\theta(\cdot \mid c^{(i)}_{<t}),\, \pi^*(\cdot \mid c^{(i)}_{<t})\right),
\label{eq:anchor}
\end{equation}
where $\mathrm{MMD}_\phi^2(\pi_\theta,\pi^*) := \|\mathbb{E}_{v\sim\pi_\theta}[\phi(v)]
- \mu_t^*\|_2^2 = (\pi_\theta-\pi^*)^\top K (\pi_\theta-\pi^*)$ is computable
in closed form, $\mathcal{O}(Vd)$ per position, with no sampling and $B$ is a mini-batch size we sample from the total anchor pool. The
weight $\tilde w_t^{(i)}$ is normalized reference varentropy,
$w_t = \mathrm{Var}_{v\sim\pi^*}[\log\pi^*(v\mid c_{<t})]$, which is maximal
at intermediate entropy where the reference distributes mass over a
small set of live candidate tokens (decision forks) and vanishes at both near-uniform and near-deterministic extremes (Appendix~\ref{app:fisher-weighting}). The total training objective is $\mathcal{L} =
\mathcal{L}_{\text{RL}} + \mathcal{L}_{\text{anchor}}$, adding only
$\sim11\%$ wall-clock overhead per step over GSPO, against $37\%$ for GRPO+KL despite more FLOPs (Appendices~\ref{sec:appendix_flops} and~\ref{sec:wallclock}). The frozen reference runs once, in a single no-grad pass over the anchor pool before training; we cache $\mu_t^{*(i)}$ and $\tilde w_t^{(i)}$ and release the reference from GPU memory. Each training step then needs one teacher-forced forward and backward pass of the policy on a mini-batch of anchors (Algorithm~\ref{alg:cfa}). 

\textbf{Naming convention.} We designate \namel{} (Content-Free Anchoring) for Eq.~\ref{eq:anchor} with
Fisher-weighted $\mathrm{MMD}_\phi^2$ on Random content-free anchors, and denote the curated real-data anchor baseline as \realname{} (Real-Data Anchoring). Ablations that change the divergence or weighting
(Fisher-KL, Unweighted-KL) are discussed in
Section~\ref{sec:discussion-ablations} and Appendix~\ref{app:full-tables}.

\subsection{Theoretical Grounding: \namel{} as a Generalized Fisher Penalty}
\label{sec:theory}

Every defense discussed in Section~\ref{sec:intro}, including EWC, KL regularization, and replay, shares an assumption about the anchor set that traces to a shared mathematical template. Near the reference policy, any twice-differentiable
divergence $D(\pi^*\|\pi)$ with $D(\pi^*\|\pi^*)=0$ reduces to a quadratic form $\frac12\Delta\pi^\top M_D\Delta\pi$ on the simplex tangent space
(Proposition~\ref{prop:quadratic-template}): discrete KL and \namel's embedding MMD are the same template, differing only in the metric $M_D$.
Discrete KL induces the Fisher-Rao metric $M_{\mathrm{KL}} = \mathrm{diag}(1/\pi^*)$ (Proposition~\ref{prop:kl-metric}); \namel's MMD is
already exactly quadratic, with $M_{\mathrm{MMD}} = 2K$ everywhere (Proposition~\ref{prop:mmd-metric}). Pulling back through the parameter Jacobian $J_\pi$ makes this precise: KL regularization toward the reference is, to leading order, exactly EWC's
Fisher-information penalty (Corollary~\ref{cor:kl-is-ewc}), while \namel{} is the same Gauss-Newton template with the Fisher-Rao metric replaced by $K$ and an adaptive varentropy reweighting (Corollary~\ref{cor:cfa-is-generalized-ewc}). EWC and \namel{} are thus the same recipe with a different metric. The template itself is only a local approximation, valid near $\theta^*$, with no built-in signal for when that stops holding. We make this checkable rather than qualitative via an explicit trust radius $\rho^*$ bounding the quadratic form's relative error (Proposition~\ref{prop:trust-radius}); tracking
$\|\Delta\theta_t\|_2$ along real RL training shows it leaves $\rho^*$ within about 120 gradient steps (Appendix~\ref{sec:beyond-second}), so a curvature matrix computed once and frozen, as in EWC, goes stale for most of a run. \namel{} sidesteps this rather than managing it: Eq.~\ref{eq:anchor} evaluates $\pi_\theta$ exactly at every step and never forms or truncates a curvature matrix.

Removing the Taylor truncation avoids the staleness problem, but invites an obvious objection: does the requirement that anchors represent the protected capabilities simply resurface in the choice of anchors? We separate two steps. \textbf{Estimation:} a finite anchor set reproduces its own population curvature $G_{\text{emb}}(\nu)$ at an $N^{-1/2}$ rate for \emph{any} i.i.d.\ anchor distribution (Propositions~\ref{prop:concentration}, \ref{prop:matrix-concentration}), which alone says nothing about retention. \textbf{Transfer:} retention on a held-out capability $\tau$ is governed by a \emph{transfer coefficient} $\kappa_S(\tau\|\nu)$, the largest ratio of $\tau$'s curvature to the anchors' curvature over the subspace $S$ in which training moves. An anchor loss of $\varepsilon$ guarantees, to second order, drift at most $\kappa_S\varepsilon$ on $\tau$'s real inputs for all capabilities at once, and $\kappa_S=\infty$ exposes a direction the anchors cannot see (Theorem~\ref{thm:transfer}). Content thus matters only through $\kappa_S$: curated anchors attain small $\kappa_S$ inside their pool's span and $\kappa_S=\infty$ outside it at every budget, while content can improve on random anchors by at most a bounded factor that needs no held-out data (Theorem~\ref{thm:content}) and isotropic anchors minimize the worst case (Proposition~\ref{prop:minimax}). This yields a budget crossover under a condition on measurable quantities (Corollary~\ref{cor:crossover}), and a limit: random anchors remove \emph{collateral} drift without content, but \emph{interference} with capabilities close to the target domain needs content (Theorem~\ref{thm:interference}). Statements, assumptions and proofs are in Appendix~\ref{app:proofs} (transfer: \ref{sec:transfer}; coverage: \ref{sec:bias-variance}; content and interference: \ref{sec:interference}), and a fully observable testbed confirms each prediction (\ref{sec:transfer-sim}).

\section{Validating the Approach in a Controlled Testbed}
\label{sec:toy-main}

Before scaling to Qwen3-VL, we resolve two design questions in a minimal, fully enumerable testbed: is functional anchoring against a reference necessary at all, and must the anchor set contain authentic prior-task data? We pretrain a 3-layer MLP on FashionMNIST (Task A), then post-train with GRPO on ParityMNIST (Task B); unmitigated GRPO reaches near-ceiling Task-B accuracy while Task-A collapses from $88.7\%$ to $16.7\%$, reproducing the single-domain collapse studied at scale.

\begin{wrapfigure}{r}{0.62\linewidth}
\centering
\vspace{-10pt}
\includegraphics[width=\linewidth]{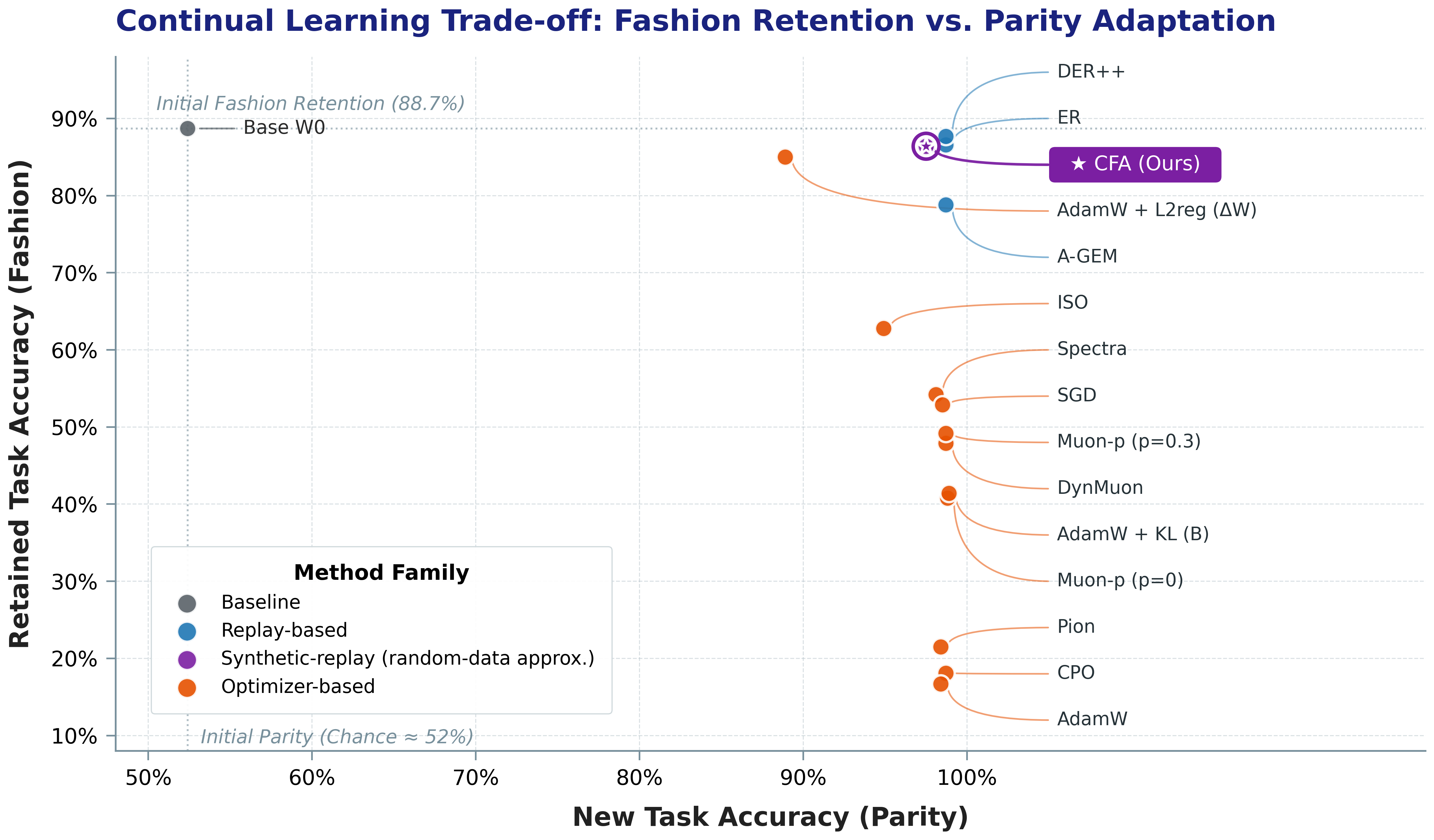}
\caption{Fashion retention vs.\ Parity acquisition.}
\label{fig:toy-buffer}
\vspace{-8pt}
\end{wrapfigure}

We sweep twelve optimizer- and parameter-space interventions spanning preconditioning (AdamW~\citep{loshchilov2017decoupled}, SGD), spectral shaping (Muon~\citep{jordan2024muon,liu2025muon} and its variants (Muon$^p$~\citealp{dong2026muonp}, DynMuon~\citealp{wu2026dynmuon}), SPECTRA~\citep{jiang2026spectra}, Pion~\citep{fan2026pion}, ISO~\citep{iso2026rlvr}), and movement penalties ($L_2$, CPO~\citep{luo2026rlforgets}). None resolve forgetting: the best retains only $47.9$-$54.2\%$ Fashion accuracy against an $88.7\%$ base, since constraining \emph{how} an update moves through weight space does not constrain the induced output distribution, which is a highly non-linear function of the parameters. Against these, real-data replay and content-free random anchoring (matching \namel's mechanism, Eq.~\ref{eq:anchor}) both retain Task-A accuracy near base ($86.6\%$ and $86.4\%$ respectively), while learning the new task to near perfection. What governs retention is whether the objective evaluates functional behavior against a reference, not whether real data is cached, or restricting parameter movement. See Appendix~\ref{sec:toy} for the full experimental setup and further discussion.

\section{Experiments and Results}
\label{sec:results}

\subsection{Experimental Setup}
\label{sec:exp-setup}

\textbf{Models.} We post-train Qwen3-VL-Instruct at $2$B, $4$B and $8$B scale on three domains drawn from MRCL's \citep{luo2026rlforgets}
task pool: Navigation, MedBookVQA, and Puzzle. Each run fine-tunes one base checkpoint on a single domain for 600 steps; we sweep the learning rate and, where present, the regularizer weight.

\textbf{Evaluation suite.} We use the common multimodal suite from
\citet{luo2026rlforgets}: STEM (MMMU-Pro, MathVista, MathVision,
MathVerse), general VQA (RealWorldQA, MMStar),
chart/document understanding (DocVQA, CharXiv-DQ, CharXiv-RQ), and
grounding (CountBenchQA). For evaluation, we sample trajectories stochastically with temperature $\tau=0.7$, nucleus sampling with top-$p=0.8$, top-$k=20$, and
max tokens $16384$, and compare against our own reproduced base-model eval
under identical settings. All reported $\pm$ values are bootstrap standard deviations over evaluation items (Appendix~\ref{app:hparams}).

\textbf{Weight decay.} \citet{luo2026rlforgets} use weight decay
throughout, which independently pulls weights toward the origin and
toward forgetting. Removing it improves retention modestly; we ablate
both configurations and report the no-weight-decay setting as our default.

\textbf{Baselines.} We compare plain policy-gradient training with and
without weight decay (GRPO wd$\in\{0,0.1\}$, GSPO wd$=0$); a domain-prompt
KL penalty (GRPO, $\beta \in \{0.01, 0.005, 0.001\}$); parameter-importance regularization
(GSPO~+~EWC with real curated anchor set; \citealp{kirkpatrick2017overcoming}); full-parameter SFT and LoRA on the same domain; and CPO \citep{luo2026rlforgets}, adapted to a
single-domain run by partitioning each training trajectory into $k{=}3$
equal segments and applying CPO's parameter-movement penalty between consecutive segments. Against these we report \realname{} and \namel{} at the anchor budgets of Section~\ref{sec:anchor-construction},
and, on Navigation specifically, the full Fisher-KL/Unweighted-KL
ablation grid across budgets $N\in\{320,640,1280,2560\}$.

\textbf{Reporting criterion.} LoRA, SFT and CPO do not always learn the target domain as well as RL-based methods, and comparing forgetting across methods that differ this much in whether they learned the task conflates two failure modes. We therefore include a baseline in the headline forgetting comparison only if its in-domain gain reaches at least $80\%$ of the median gain of the RL baselines (GRPO with $\text{wd}\in\{0,0.1\}$, GRPO+KL, GSPO, and GSPO+EWC) on that domain. The threshold depends only on in-domain accuracy, never on out-of-domain results; complete metrics for every run, qualifying or not, appear in Appendix~\ref{app:full-tables}. Each configuration is one run; three seeds on Navigation-2B leave every conclusion unchanged (Appendix~\ref{app:seeds}).

\begin{figure}[hbt!]
\centering
\includegraphics[width=0.95\linewidth]{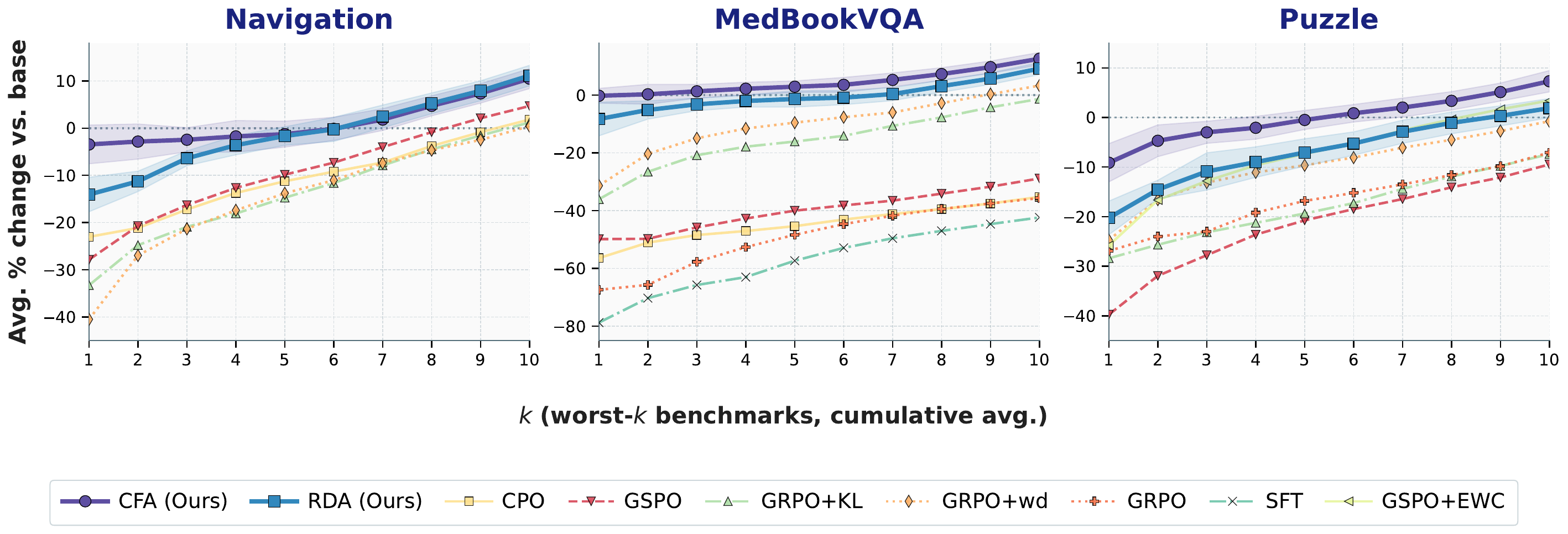}
\caption{\textbf{Qwen3-VL-2B.} Worst-$k$ forgetting curves on Navigation, MedBookVQA, and Puzzle. Each panel plots the cumulative average \% change relative to the base model over the $k$ most-affected out-of-domain benchmarks, alongside \realname{} and \namel{} at $N=1280$. The right endpoint, $k=10$, is the suite-wide average that is conventionally reported on its own; the gap between it and the left endpoint is the forgetting the average conceals. \namel{} substantially reduces tail degradation across all qualifying baselines in all three domains.}
\label{fig:results-2b}
\end{figure}

\subsection{Results}
\label{sec:results-main}
\label{sec:phenomenon}

Figure~\ref{fig:results-2b} shows the worst-$k$ curves at 2B. On Navigation, \namel{} reduces the worst-benchmark change to $-3.5\%$, against $-40.5\%$ for GRPO (wd$=0.1$), $-23.0\%$ for CPO, the strongest qualifying baseline, and $-14.1\%$ for \realname{}, while also posting the highest in-domain gain of any method. WiSE-FT (Appendix~\ref{app:wiseft}) reduces forgetting relative to unregularized RL, but trades away target-task gains and still concedes a larger tail than \namel{}. On MedBookVQA, unmitigated GRPO (wd$=0$) shows the most severe collapse in this study, $-67.4\%$ despite $99.8\%$ in-domain accuracy; \namel{} holds it to $-0.3\%$. On Puzzle, \namel{} again reduces worst-$k$ forgetting relative to every baseline that learns the domain comparably well, from $-24.8\%$ (GRPO, wd$=0.1$) to $-9.1\%$. CPO's apparent CountBenchQA retention on Puzzle comes with an in-domain gain of only $+39.0$ points (final $62.8\%$), $63\%$ of the median RL gain ($+62.2$), so it falls below the reporting threshold (Table~\ref{tab:puz-2b-full}).

\begin{figure}[hbt!]
\centering
\includegraphics[width=0.95\linewidth]{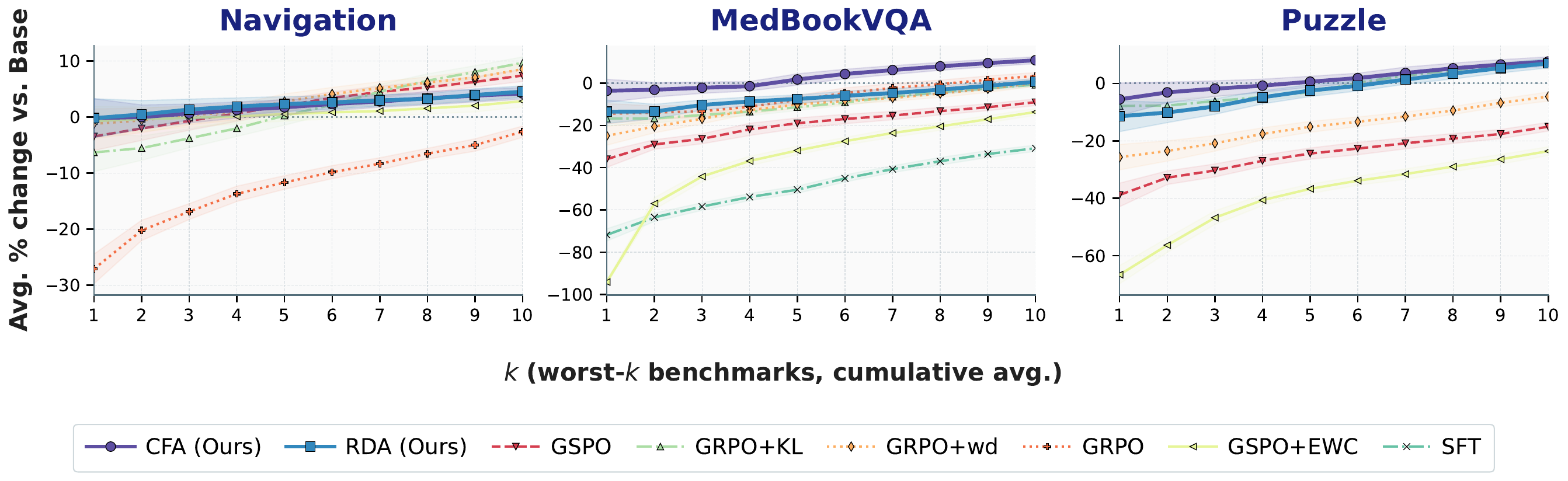}
\caption{\textbf{Qwen3-VL-4B.} Worst-$k$ forgetting curves on the same
three domains and under the same reporting criterion as
Figure~\ref{fig:results-2b}. MedBookVQA and Puzzle reproduce the 2B
ordering, with \namel{} and \realname{} again separating cleanly
from every baseline. Navigation shows an atypical pattern at this scale
specifically, discussed in Appendix~\ref{app:domain-scale-mechanisms}.}
\label{fig:results-4b}
\end{figure}

\begin{figure}[hbt!]
\centering
\includegraphics[width=0.95\linewidth]{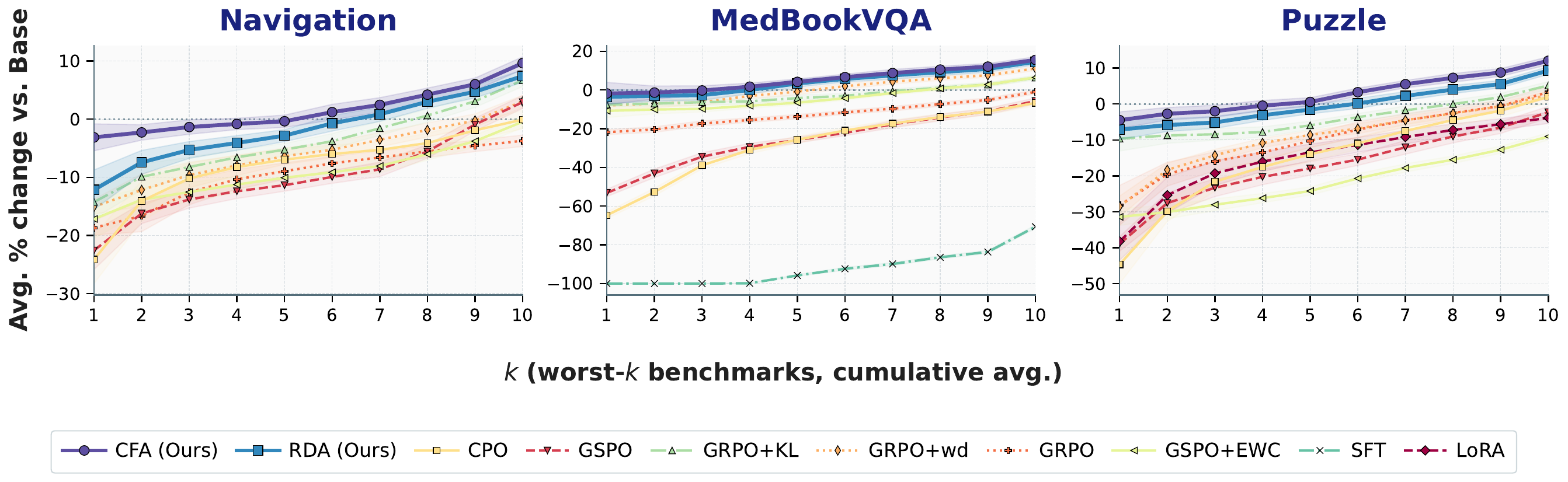}
\caption{\textbf{Qwen3-VL-8B.} Worst-$k$ forgetting curves on the same
three domains as Figures~\ref{fig:results-2b}-\ref{fig:results-4b}, at
the largest scale tested. \realname{} and \namel{} remain the two
best-retaining methods across all three domains, and the gap to the
strongest baseline does not shrink with scale.}
\label{fig:results-8b}
\end{figure}

The ordering is stable across scale. At 4B (Figure~\ref{fig:results-4b}),
\namel{} outperforms every qualifying baseline, reaching $-0.1\%$/$-3.6\%$/$-5.5\%$ on Navigation/MedBookVQA/Puzzle; \realname{} does so everywhere except Puzzle, where GRPO+KL ($-7.8\%$) beats it ($-11.5\%$). GSPO~+~EWC fails badly here ($-94.1\%$ on MedBookVQA, DocVQA collapsing $64.4\to3.8$), evidence that a curvature matrix frozen at
$\theta^*$ can actively mislead once training leaves its valid region. At 8B (Figure~\ref{fig:results-8b}),
the ordering holds: \namel{} posts the smallest worst-benchmark change in all three domains ($-3.1\%$/$-1.8\%$/$-4.4\%$), \realname{} second-best throughout ($-12.1\%$/$-3.5\%$/$-7.1\%$), against the strongest qualifying baseline (GRPO+KL: $-14.2\%$/$-7.6\%$/$-9.6\%$); CPO degrades sharply at
this scale (worst-1 accuracy of $-64.7\%$ MedBookVQA, and $-44.6\%$ Puzzle), performing worse than plain GRPO on both. The Real-vs-Content-free gap narrows slightly from 2B to 8B on Navigation ($10.6\to9.0$ points), consistent with content-free anchors needing less relative budget as capacity grows.

Across all nine domain-scale combinations, \namel{} keeps the worst-benchmark change within $10\%$ of base (range $-0.1\%$ to $-9.1\%$). No single baseline is reliably safe: the best baseline differs from cell to cell, and every baseline run on the full grid collapses somewhere, with worst cases ranging from $-36.0\%$ (GRPO+KL) to $-94.1\%$ (GSPO+EWC) over the cells where it qualifies (Table~\ref{tab:worst-per-method}). GSPO+EWC matches \namel{} on MedBookVQA-2B yet loses $94.1\%$ on MedBookVQA-4B, and GRPO+KL is close to \namel{} on Puzzle-4B yet loses $36.0\%$ on MedBookVQA-2B, so picking the best baseline per cell would require knowing the forgetting outcome in advance. The failure is qualitative as well as quantitative: on GRPO's most severe case (MedBookVQA, $-67.4\%$), post-RL models retain fluent, correctly structured reasoning while the correspondence between that trace and the actual image content degrades into three recurring failure modes: hallucinated instances, shortcut answers, and loss of verification.\label{sec:qualitative} Full examples of each are in Appendix~\ref{app:qualitative}.

\subsection{Anchor Content, Budget, and Divergence-Measure Ablations}
\label{sec:discussion-ablations}

\begin{figure}[hbt!]
\centering
\includegraphics[width=0.95\linewidth]{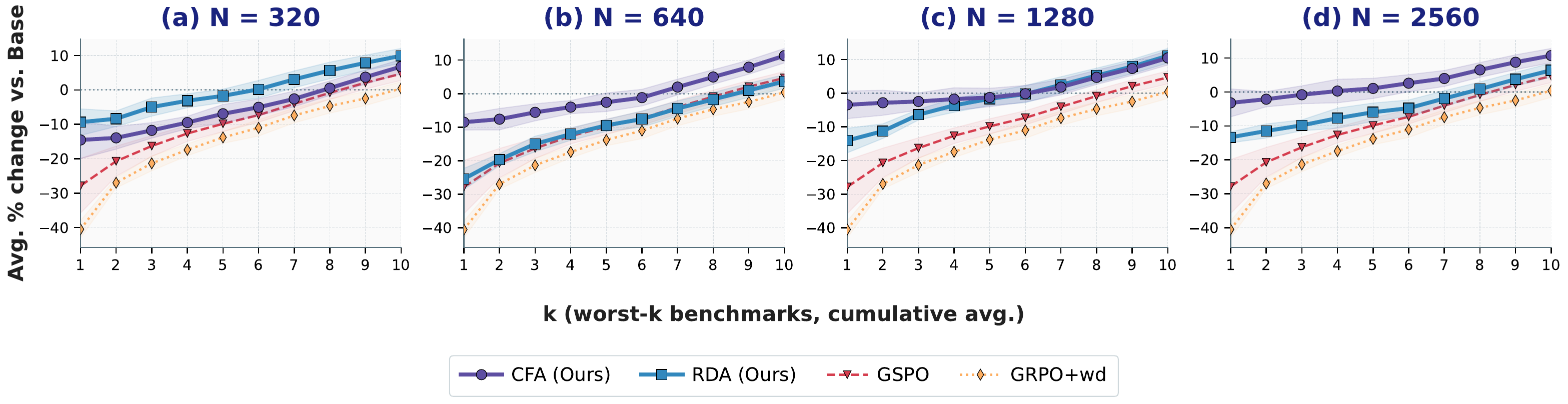}
\caption{\textbf{Anchor budget and content.} Worst-$k$ forgetting curves
on Navigation (2B) for \realname{}, \namel, GSPO, and GRPO
(wd=$0.1$), at four anchor budgets $N\in\{320,640,1280,2560\}$. Real anchors are budget-efficient and essentially flat past
$N{=}320$; Content-free anchors are budget-hungry, matching real anchors by
$N{=}640$ and exceeding them on the worst-1 metric, while
real anchors degrade slightly relative to their own $N{=}320$ peak. This
is the pattern Corollary~\ref{cor:crossover} predicts.}
\label{fig:results_anchor_size}
\end{figure}

Figure~\ref{fig:results_anchor_size} shows a clear difference in budget efficiency between real and Content-free anchors. Real anchors reach essentially their full benefit by $N{=}320$ and then plateau, with larger budgets providing little or slightly negative improvement. Content-free anchors require more samples: they are weaker at $N{=}320$, match real anchors by $N{=}640$, and plateau by $N{=}2560$. This crossover is what Appendix~\ref{sec:bias-variance} predicts: a curated coreset's coverage is capped by the span of its pool at every budget, so its retention guarantee saturates, whereas content-free anchors need a larger budget before their empirical curvature becomes well conditioned, after which their worst-case transfer coefficient is lower. Corollary~\ref{cor:crossover} formalizes this behavior. Because padding is masked, a real anchor contributes about $4.2\times$ fewer scored tokens than a content-free one; the crossover persists at matched token budgets on Qwen3-VL (Appendix~\ref{app:matched-tokens}) and in the testbed, where token counts match exactly (Appendix~\ref{sec:transfer-sim}). Mixing the two anchor types inherits the better of both on every capability, up to a harmonic-mean bound (Proposition~\ref{prop:hybrid}); a 50/50 hybrid in the testbed retains nearly as well as \namel{} at higher target accuracy.

\begin{wrapfigure}{r}{0.42\linewidth}
\centering
\includegraphics[width=\linewidth]{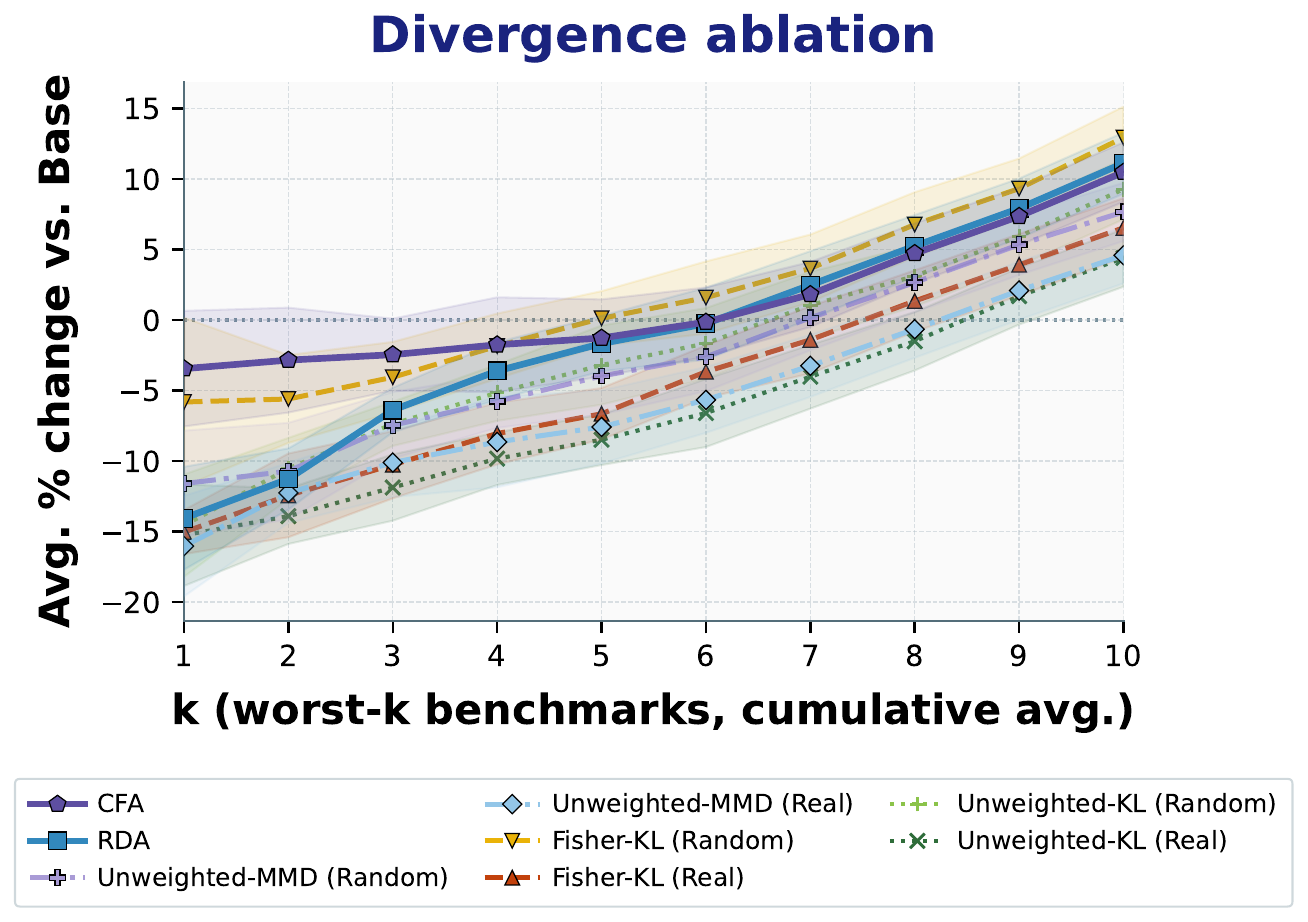}
\caption{\textbf{Divergence ablation.} Worst-$k$ forgetting, Navigation
(2B), $N{=}1280$. Full results in Table~\ref{tab:divergence_ablation_full}.}
\label{fig:results_divergence}
\end{wrapfigure}
Figure~\ref{fig:results_divergence} isolates the effect of the divergence
and weighting choices at the deployed anchor budget ($N{=}1280$). Within
real-anchor content, the predicted ordering holds cleanly at every $k$:
\realname{} outperforms Fisher-KL (Real), which in turn outperforms
Unweighted-KL (Real). This isolates the value of the embedding metric $K$
over the discrete Fisher-Rao metric, and, within the Fisher-Rao metric, the
added value of varentropy weighting over uniform weighting. Within the
content-free anchor set, the same ordering holds for the most part. For
completeness, we also report unweighted MMD: under both divergences and
both anchor types, removing varentropy weighting lowers retention at every
$k$, confirming that weighting provides a consistent benefit independent of
anchor content. Finally, Appendix~\ref{app:anchor-sampling} varies how content-free tokens are sampled: uniform sampling over the full vocabulary, our default, gives the best retention, while restricted or reweighted vocabularies give no consistent benefit and can degrade it substantially.

\subsection{Generalization Beyond the Main Grid}
\label{sec:other-families}

\begin{figure}[hbt!]
\centering
\includegraphics[width=0.95\linewidth]{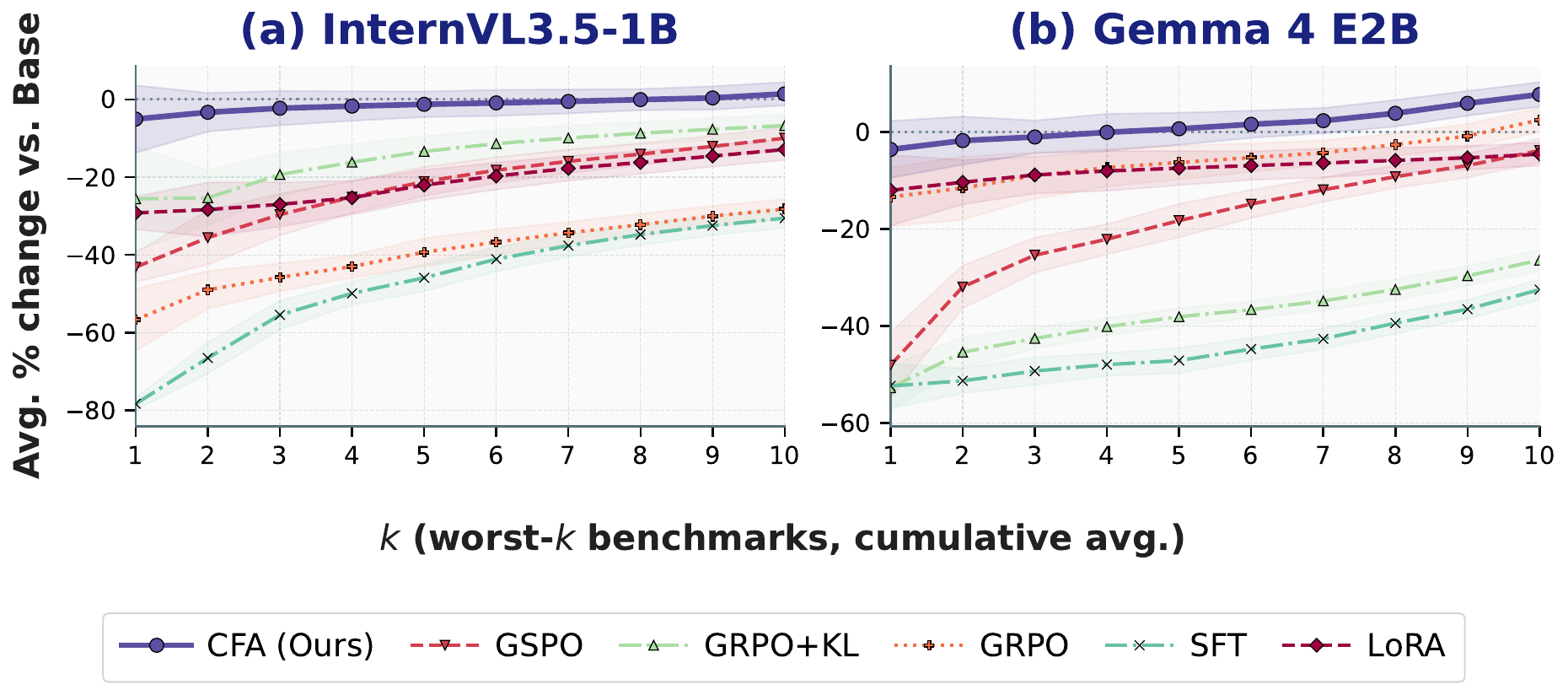}
\caption{\textbf{Robustness to architecture.} Worst-$k$ forgetting curves for InternVL3.5-1B and Gemma~4~E2B trained on \otherdomain{} for 600 steps, over the ten benchmarks of the common suite. All trained methods are shown; no in-domain qualifying threshold is applied. Full results in Tables~\ref{tab:internvl-full} and~\ref{tab:gemma-full}.}
\label{fig:other-families}
\end{figure}

\textbf{Other model families.} We apply \namel{} unchanged ($N=1280$, $T=2048$) to two VLM families with different vision encoders and language backbones, InternVL3.5-1B and Gemma~4~E2B. Every method is trained on \otherdomain{} for 600 steps and evaluated on the common suite (Figure~\ref{fig:other-families}). Without the compute for per-method sweeps, we use one learning rate per training type ($5\times10^{-6}$ for all RL runs, $10^{-5}$ for SFT, $5\times10^{-5}$ for LoRA), so in-domain accuracy could likely be improved slightly by tuning. On both models, \namel{} has the smallest worst-$k$ forgetting at every $k$: its worst-1 change is $-5.1\%$ on InternVL3.5-1B and $-3.6\%$ on Gemma~4~E2B, against at best $-25.6\%$ (GRPO+KL) and $-12.0\%$ (LoRA) for any other method. Relative to GSPO, the algorithm it regularizes, \namel{} reduces worst-1 forgetting from $-43.1\%$ to $-5.1\%$ and from $-48.0\%$ to $-3.6\%$, while matching or exceeding its in-domain accuracy ($78.0$ vs.\ $77.9$ and $77.0$ vs.\ $68.7$); on InternVL3.5-1B it reaches the highest in-domain accuracy of any method. As on Qwen3-VL, the strongest baseline changes from model to model: GRPO+KL is runner-up on InternVL3.5-1B but loses $52.7\%$ on Gemma~4~E2B, while GRPO forgets little on Gemma~4~E2B ($-13.4\%$) but loses $56.6\%$ on InternVL3.5-1B. Per-benchmark results are in Appendix~\ref{app:other-families-full}.

\textbf{Other training algorithms and modalities.} Because the anchor term depends only on the model's outputs on anchors, it can be added unchanged to any training loss. Added to SFT, the most destructive method in our study, \namel{} cuts worst-1 forgetting on \otherdomain-2B from $-78.8\%$ to $-9.1\%$ while keeping most of the target gain (Table~\ref{tab:med-2b-sft-lora}). On four text-only models, where RL forgets far less than in the vision-language setting, it reduces the remaining forgetting further at unchanged target accuracy (Appendix~\ref{app:text-only}).

\textbf{Other benchmarks.} Finally, we evaluate the Navigation-2B checkpoints of every qualifying method on six benchmarks outside the common suite: HallusionBench, OCRBench, AI2D, HRBench-4K, ScreenSpot, and EmbSpatialBench (Appendix~\ref{app:extended-ood}). \namel{} has the smallest worst-1 ($-7.4\%$) and worst-3 ($-2.5\%$) change on the new benchmarks, against $-29.9\%$ for \realname{} and $-11.0\%$ to $-37.0\%$ for the RL baselines, and remains best over all sixteen benchmarks combined. The two anchor types fail in complementary places, as Theorem~\ref{thm:content} predicts: \realname{} retains OCRBench, whose text-and-document skills its curated pool represents, but loses AI2D, which the pool does not.

\section{Conclusion}
\label{sec:conclusion}

Single-domain RL post-training silently collapses held-out capabilities of Qwen3-VL at every scale we tested, with drops of up to $67\%$ hidden behind flat suite-wide averages. Every existing defense assumes that the anchor set must represent the capabilities it protects; we have shown that this is unnecessary. Recasting anchoring as exact Gauss-Newton functional regularization shows that an anchor set protects a capability only through a transfer coefficient, which anchor content can improve over random anchors by at most a bounded factor: anchors must supply coverage, not relevance, except along directions the target task itself needs. \namel{}, built from uniformly random tokens, matches a curated five-dimension coreset at a modest budget, keeps worst-benchmark forgetting under $10\%$ in every domain and scale tested, and carries over unchanged to InternVL3.5 and Gemma~4, to SFT, to text-only models, and to benchmarks outside the original suite, with no real, licensed, or curated data anywhere in the loop. Appendix~\ref{sec:limitations} discusses the limits of these guarantees and the settings our evaluation does not cover.

\subsection*{AI use statement}
In this work, we used generative AI tools exclusively for assistance with grammar checking, text formatting, and LaTeX typesetting. We have not used generative AI tools for scientific data generation, model implementation, or empirical experiment execution. We take full responsibility for the final content of this work, including all text, mathematical derivations, empirical claims, and conclusions.

\subsection*{Reproducibility Statement}

Our implementation makes minimal modifications to the open-source Continual Policy Optimization (CPO) codebase of \citet{luo2026rlforgets}\footnote{\url{https://github.com/MaolinLuo/CPO}}. From this base we inherit the GRPO/GSPO trainer, dataset loaders, and evaluation harness; \namel{} is implemented as a trainer subclass on top of it, with the core mechanism given as Algorithm~\ref{alg:cfa} in Appendix~\ref{app:anchors}. The full hyperparameter search space swept for every method reported in the paper (anchor budget $N$, regularizer strength $\lambda$, divergence family, learning rate, KL coefficient $\beta$, and number of optimization steps) is given in Appendix~\ref{app:hparams}.

\paragraph{Datasets.} Continual post-training and in-domain evaluation use
the five MRCL benchmark tasks (MedBookVQA, Navigation, We-Math2, Puzzle, and
FinMME), released at \url{https://huggingface.co/datasets/MaolinLuo/MRCL}.
General-capability (out-of-domain) retention is evaluated on nine external
benchmarks spanning STEM reasoning, general VQA, chart/document
understanding, and grounding: MMMU-Pro, MathVerse, MathVision,
MathVista, RealWorldQA, MMStar, DocVQA, CharXiv, and CountBenchQA.

\paragraph{Compute.} All experiments were run on nodes of 8 NVIDIA H100 GPUs
and 8 NVIDIA B40 GPUs. We use DeepSpeed ZeRO Stage 2 for Qwen3-VL-2B, and Qwen3-VL-4B and ZeRO
Stage 3 with CPU offload for Qwen3-VL-8B.

\subsection*{Broader Impact}
This work studies a failure mode of RL post-training and a mitigation for
it. To the best of our knowledge, we do not foresee a plausible dual-use concern specific to either
the diagnosis or the method. We show that \namel's anchor sets can be built entirely from synthetic noise, removing a data-provenance and licensing burden that otherwise falls on practitioners
trying to protect a model's prior capabilities. We view this as a
net-positive access effect rather than a risk. Forgetting
mitigation is not safety alignment: \namel{} constrains behavioral drift on
capabilities present in the reference policy, and does not itself
constrain or evaluate whether those reference behaviors were safe or
desirable to begin with. We do not view this as unique to our method as
every regularizer discussed in Section~\ref{sec:related} that constrains drift toward a reference has the same property.

\bibliography{references}
\bibliographystyle{iclr2027_conference}

\newpage
\appendix

\section{A Toy Testbed for Forgetting}
\label{sec:toy}

Before running expensive experiments at Qwen3-VL scale
(Section~\ref{sec:results}), we construct a minimal, fully enumerable
testbed to resolve two foundational questions motivating
\namel{}'s design. First, do mitigations that constrain
the optimizer, update geometry, or parameter movement perform as well as
explicitly evaluating functional behavior against a reference anchor set?
Second, does an anchor set require authentic prior-task data, or can
completely unstructured noise, bearing zero resemblance to the
pretraining data distribution, fulfill the same behavioral-anchoring
purpose?

\subsection{Setup}
\label{sec:toy-setup}

We use a 3-layer MLP pretrained via supervised fine-tuning on FashionMNIST
(Task A, $60$K examples, predicting apparel category), then GRPO
post-trained on ParityMNIST (Task B, $60$K examples, predicting digit
parity) for 20 epochs. This mirrors the single-domain RL transition
studied at scale in Section~\ref{sec:prelim}: one established prior
capability, one distinct downstream RL objective, no interleaved task
sequence. The base checkpoint scores $88.7\%$ on FashionMNIST and $52.4\%$
on ParityMNIST. AdamW post-training with no mitigation reaches $98.7\%$
ParityMNIST accuracy while degrading FashionMNIST to $16.7\%$ - a
$-82\%$ relative loss of prior-task capability, reproducing the
single-domain collapse in a testbed small enough for full spectral and
representational profiling.

\subsection{Optimizer-Level Fixes Are Not Enough}
\label{sec:toy-optimizers}

A natural first hypothesis is that this is an optimization artifact: a
more geometrically principled update rule, spectral regularizer, or
parameter-space movement constraint should prevent representation
collapse while still letting the model master the new task. We test this
by sweeping four families of optimizer-level and parameter-space
interventions under identical initializations, learning-rate budgets, and
schedules: (i) element-wise preconditioning - SGD with momentum and
AdamW \citep{loshchilov2017decoupled}; (ii) spectral whitening and
fractional shaping - Muon \citep{jordan2024muon,liu2025muon}, which
projects momentum onto the polar factor $UV^\top$ via Newton-Schulz
iteration; fractional spectral powers (Muon$^p$, \citealp{dong2026muonp})
interpolating steepest descent across Schatten-$q$ norms; and
curvature-adaptive spectral scheduling (DynMuon,
\citealp{wu2026dynmuon}); (iii) spectral filtering and manifold
constraints - spectral-norm clipping (SPECTRA, \citealp{jiang2026spectra}),
high-pass Newton-Schulz filtering for low-SNR RL regimes (Pion,
\citealp{fan2026pion}), and RLVR-native frame optimization with frozen
base spectra (ISO, \citealp{iso2026rlvr}); and (iv) parameter-space
movement penalties - an explicit $L_2$ penalty on $\Delta W=W-W_0$, and
CPO \citep{luo2026rlforgets}.

None of these interventions resolve catastrophic forgetting
(Figure~\ref{fig:toy-buffer}). Every method reaches near-ceiling
ParityMNIST accuracy ($96.5$-$98.8\%$) while sacrificing most of Task-A
performance: unconstrained preconditioning (AdamW, standard Muon) retains
only $16.7$-$40.8\%$ Fashion accuracy; spectral shaping and tail
suppression (DynMuon, Muon$^p$ at $p{=}0.3$, SPECTRA, Pion) modestly
improve retention to $47.9$-$54.2\%$, still far below the $88.7\%$ base;
frame-frozen manifold optimization (ISO) still suffers catastrophic
collapse ($62.8\%$); and explicit distance penalties (CPO, $L_2$ parameter
damping) mitigate weight drift but still fail to protect Task-A
classification boundaries ($18.1\%$ retention). SGD with momentum is a
partial exception, retaining Fashion accuracy noticeably better than
AdamW or standard Muon at a comparable Parity acquisition level.
This toy-scale advantage is not, on its own, evidence of a viable
mitigation strategy at scale. The toy network has a single weight scale
and few layers, so a single global learning rate adequately characterizes
its optimization dynamics. Qwen3-VL has no such uniformity: attention
projections, MLP expansion layers, embedding tables, and normalization
parameters each operate at markedly different effective weight and
gradient scales, and this heterogeneity grows with depth. We therefore do
not treat SGD's toy-scale behavior as informative about mitigation
strategies at the Qwen3-VL scale and exclude it from the large-scale
comparisons in Section~\ref{sec:results}.

These findings align with \citet{iso2026rlvr}, who report that RLVR
barely perturbs singular values ($\sim3\%$ spectral residual vs.\
$\sim35\%$ in SFT), localizing updates primarily in the singular frames
$(U,V)$. Yet as demonstrated here, weight-space geometric constraints do
not guarantee behavioral preservation: because the mapping from parameter
matrices to representation manifolds is non-injective and highly
non-linear, constraining \emph{how} an update navigates weight space does
not prevent catastrophic output drift.

\subsection{Random Anchors Match Real Replay}
\label{sec:toy-anchors}

Figure~\ref{fig:toy-buffer} contrasts these optimizer-level methods
against two functional reference-matching paradigms: real-data replay
(distillation on a stored buffer of genuine Task-A exemplars; e.g., Experience Replay
\citealp{chaudhry2019tiny} and A-GEM, \citealp{chaudhry2019agem}, and
synthetic anchoring, matching the model's behavior on purely random
tokens against the frozen reference $W_0$ via the same
Fisher-weighted MMD used at scale (\namel, Eq.~\ref{eq:anchor}).\footnote{The
3-layer MLP has no token-embedding matrix $E$; we substitute the frozen
reference model's final-layer classifier weight matrix
$W_{\text{out}}^{\text{ref}}\in\mathbb{R}^{C\times d}$, which plays an
analogous linear-projection role. We note an important dimensional rank inversion
between the two settings: in the toy MLP, the number of classes is small ($C=10 \le d$),
so the linear Gram matrix $K_{\text{toy}} = \frac{1}{d} W_{\text{out}}^{\text{ref}} (W_{\text{out}}^{\text{ref}})^\top \in \mathbb{R}^{C \times C}$
has full rank $\operatorname{rank}(K_{\text{toy}}) = C = 10$. In contrast, for VLMs,
the vocabulary dimension vastly exceeds the hidden embedding dimension ($d \approx 2048 \ll V \approx 152{,}000$),
rendering $K = \frac{1}{d} E E^\top$ severely rank-deficient ($\operatorname{rank}(K) \le d$,
with null-space dimension $> 147{,}000$; see Section~\ref{sec:theory}). Corollary~\ref{cor:cfa-is-generalized-ewc}'s
algebraic pullback holds for any fixed linear projection, but the VLM setting induces
a semantic low-rank projection rather than full categorical coverage.}
Furthermore, in single-step classification where sequence length $T=1$, the sequence-normalized varentropy weight $\tilde{w}_t = w_t / (\sum_{s=1}^T w_s + \epsilon)$ collapses identically to $\tilde{w}_1 \equiv 1.0$ regardless of logit values. The toy benchmark therefore isolates the geometric impact of unweighted functional MMD anchoring from temporal entropy scheduling.

Real Task-A replay
achieves one of the best retention scores ($86.6\%$ Fashion), and
synthetic noise anchoring ($86.4\%$) sits within noise of it, both near
base performance, while every optimizer-level and parameter-space method
collapses. What governs retention is not the algorithmic complexity of
the optimizer, the rank of the update, or whether real domain data is
cached, but whether the objective explicitly evaluates and anchors
functional behavior against a reference.

\section{Metric Motivation}
\label{app:metric}

\textbf{The suite average hides the failure.} Averaging is not merely less
sensitive than the worst case here but is actively misleading. Across the
$68$ qualifying runs in this study, $20$ post a suite-wide mean above
$-5\%$, which by the standard metric reads as essentially no forgetting,
while simultaneously losing more than $20\%$ on their single worst
benchmark. CPO on Puzzle at 8B averages $+2.0\%$ yet drops $-44.6\%$ at
$k=1$; GRPO (wd$=0.1$) on Navigation at 2B averages $+0.4\%$ while
CountBenchQA falls $-40.5\%$. The suite mean lets positive transfer on
reasoning benchmarks pay for the destruction of a narrow capability, and gives the false security of minimal capability degradation. A practitioner reading only
$\bar\Delta$ would ship both models. Concretely, a method can appear competitive at $k=|\mathcal{B}|$ while silently dropping by 30 to 60 points at $k=1$, which is exactly what our metric worst-$k$ curve is meant to expose.

\textbf{On comparing different benchmarks at the same $k$.} Because $\mathcal{B}_k$ is chosen per method, two methods at a given $k$ might be scored on different benchmarks. It is also empirically the common case rather than an edge case:
across our qualifying runs, $7$ of the $10$ common-suite benchmarks appear
as some method's worst, the most frequent (CountBenchQA) accounts for only $32\%$ of them, and two methods within the same panel agree on $\mathcal{B}_1$ for just $25\%$ of pairs. We argue this is a property the metric must have, for three reasons.

\emph{(i) The rule is identical for every method, and it is self-penalizing.}
By the second equality in Eq.~\eqref{eq:worstk}, $\Delta_k$ is the minimum
over \emph{all} size-$k$ subsets. Every method is therefore evaluated under
the same rule and at its own least favorable slice; none is scored on a
subset chosen to flatter it. A method that wins at a given $k$ does so
despite being measured where it is individually weakest.

\emph{(ii) A fixed, pre-chosen subset would be less fair, not more.} Which
capability collapses is method-dependent as highlighted by the  $32\%$ figure above. Committing in advance to a single probe would turn the comparison into a test of one arbitrary capability: fixing on
CountBenchQA would have missed GSPO~+~EWC's $-94.1\%$ collapse on DocVQA at 4B (Section~\ref{sec:results-main}), and fixing on DocVQA would have missed
the CountBenchQA collapses that motivate this work. The quantity a practitioner cares about is whether \emph{any} retained capability was destroyed, not whether a particular one was.

\emph{(iii) Nothing is discarded, so the curves compare distributions rather
than subsets.} The map $\{\Delta_b\} \mapsto \{\Delta_k\}_{k=1}^{|\mathcal{B}|}$
is a bijection onto the sorted score vector: reporting the whole curve is
equivalent to reporting the entire empirical distribution of per-benchmark
change, just re-indexed so that the tail is read first. Consequently, if one
method's curve lies above another's at \emph{every} $k$, its loss profile
dominates in the majorization (second-order stochastic dominance) sense, and
is preferred under every risk-averse aggregation of $\{\Delta_b\}$ and not merely at the particular $k$.

That last condition is what we actually observe, so our conclusions do not rest on a choice of $k$: \namel{} dominates the qualifying baselines at every
$k$ in $54$ of $59$ method comparisons ($92\%$), and in $56$ of $59$ ($95\%$)
over the worst half of the suite, $k\leq5$. All five exceptions occur in the
two Navigation panels at points where \emph{both} curves are non-negative,
i.e.\ where neither method is forgetting and the comparison has become one of average positive transfer rather than of tail damage. Per-benchmark
numbers are reported in full in Appendix~\ref{app:full-tables} for better comparison.

To avoid conflating failure to learn the target task with capability retention, we exclude baselines that fail to reach an in-domain gain threshold from headline retention comparisons, reporting them in full only in
Appendix~\ref{app:full-tables}.

\section{Full Result Tables}
\label{app:full-tables}

This section provides the complete evaluation results across all post-training configurations, baseline algorithms, and model scales (Qwen3-VL-2B, 4B, and 8B) on the 10-benchmark common evaluation suite.
For every setting, the worst-$k$ retention curves ($k=1,\ldots,10$) illustrated in Section~\ref{sec:results} (Figures~\ref{fig:results-2b}-\ref{fig:results-8b}) are derived directly from these per-benchmark accuracies: for a checkpoint with scores $s_b$ and base scores $s_b^{\text{base}}$, the per-benchmark relative shifts $\Delta_b = (s_b - s_b^{\text{base}})/s_b^{\text{base}}$ are sorted in ascending order $\Delta_{(1)} \le \Delta_{(2)} \le \dots \le \Delta_{(10)}$, and the worst-$k$ metric is computed as $\mathcal{W}_k = \frac{1}{k}\sum_{j=1}^k \Delta_{(j)}$. Specifically, Navigation evaluations across scales are reported in Tables~\ref{tab:nav-2b-full}, \ref{tab:nav-4b-full}, and \ref{tab:nav-8b-full}; MedBookVQA evaluations in Tables~\ref{tab:med-2b-full}, \ref{tab:medbook-4b-full}, and \ref{tab:medbook-8b-full}; and Puzzle evaluations in Tables~\ref{tab:puz-2b-full}, \ref{tab:puzzle-4b-full}, and \ref{tab:puzzle-8b-full}. Table~\ref{tab:wiseft-nav-2b} evaluates post-hoc weight merging (WiSE-FT; \citealp{wortsman2022robust}) across mixing coefficients $\alpha \in \{0.1,\dots,0.9\}$ on Navigation. Table~\ref{tab:anchor_budget_ablation_full} details the full anchor budget scaling sweep ($N \in \{320, 640, 1280, 2560\}$), and Table~\ref{tab:divergence_ablation_full} compares functional divergence criteria. Appendix~\ref{app:seeds} reports variation across training seeds.

\begin{table}[t]
\centering
\caption{\textbf{Navigation, Qwen3-VL-2B.} Common-suite accuracy (mean $\pm$ bootstrap std over evaluation items) for every baseline and for \namel{} at anchor budget $N=1280$ (Real/Content-Free). \textbf{Bold} marks the best method per column; \underline{underline} marks the second best. ``Target'' is final in-domain (target-task) accuracy.}
\label{tab:nav-2b-full}
\resizebox{\textwidth}{!}{%
\begin{tabular}{lccccccccccc}
\toprule
 & \multicolumn{4}{c}{STEM} & \multicolumn{2}{c}{Gen. VQA} & \multicolumn{3}{c}{Chart/Doc.} & \multicolumn{1}{c}{Ground.} &  \\
\cmidrule(lr){2-5} \cmidrule(lr){6-7} \cmidrule(lr){8-10} \cmidrule(lr){11-11}
Method & MMMU & MVi & MVn & MVe & RWQA & MMS & DocVQA & CX-DQ & CX-RQ & CBQA & Target \\
\midrule
Base & 48.4\,$\pm$\,2.4 & 56.0\,$\pm$\,3.1 & 21.1\,$\pm$\,1.4 & 37.4\,$\pm$\,1.5 & 58.4\,$\pm$\,3.5 & 46.7\,$\pm$\,2.5 & 68.6\,$\pm$\,1.2 & 60.8\,$\pm$\,1.5 & 28.0\,$\pm$\,2.8 & \underline{82.5\,$\pm$\,3.4} & 3.5 \\
GRPO (wd=0) & 44.9\,$\pm$\,2.3 & 55.6\,$\pm$\,3.1 & 19.9\,$\pm$\,1.4 & 36.2\,$\pm$\,1.5 & 49.5\,$\pm$\,3.5 & 58.7\,$\pm$\,2.5 & \textbf{71.6\,$\pm$\,1.2} & \underline{60.9\,$\pm$\,1.5} & 28.3\,$\pm$\,2.8 & 80.7\,$\pm$\,3.5 & 34.4$^{\dagger}$ \\
GRPO (wd=0.1) & 48.7\,$\pm$\,2.4 & 64.2\,$\pm$\,3.0 & 24.3\,$\pm$\,1.5 & 42.7\,$\pm$\,1.5 & 52.5\,$\pm$\,3.5 & 59.0\,$\pm$\,2.5 & 59.4\,$\pm$\,1.3 & 57.4\,$\pm$\,1.5 & \underline{28.8\,$\pm$\,2.8} & 49.1\,$\pm$\,4.4 & 70.8 \\
GRPO+KL & 47.8\,$\pm$\,2.4 & 64.2\,$\pm$\,3.0 & 25.8\,$\pm$\,1.6 & 44.6\,$\pm$\,1.6 & \underline{60.6\,$\pm$\,3.5} & 58.8\,$\pm$\,2.5 & 45.8\,$\pm$\,1.3 & 52.8\,$\pm$\,1.6 & 25.3\,$\pm$\,2.7 & 69.0\,$\pm$\,4.1 & 75.7 \\
GSPO & 51.0\,$\pm$\,2.4 & \underline{64.8\,$\pm$\,3.0} & 27.0\,$\pm$\,1.6 & 45.4\,$\pm$\,1.6 & 59.2\,$\pm$\,3.5 & 58.7\,$\pm$\,2.5 & 67.4\,$\pm$\,1.3 & 56.2\,$\pm$\,1.5 & 20.2\,$\pm$\,2.8 & 71.3\,$\pm$\,4.0 & 78.7 \\
GSPO+EWC & 47.8\,$\pm$\,2.4 & 58.5\,$\pm$\,3.0 & 21.3\,$\pm$\,1.4 & 37.0\,$\pm$\,1.5 & 48.5\,$\pm$\,3.5 & 56.3\,$\pm$\,2.5 & \underline{71.4\,$\pm$\,1.2} & 57.5\,$\pm$\,1.5 & 26.2\,$\pm$\,2.7 & \textbf{87.6\,$\pm$\,2.9} & 51.1$^{\dagger}$ \\
LoRA & 46.9\,$\pm$\,2.4 & 57.2\,$\pm$\,3.1 & 18.2\,$\pm$\,1.4 & 37.3\,$\pm$\,1.5 & 55.8\,$\pm$\,3.5 & 39.1\,$\pm$\,2.5 & 65.1\,$\pm$\,1.3 & 58.6\,$\pm$\,1.5 & 25.3\,$\pm$\,2.7 & 68.8\,$\pm$\,4.1 & 3.9$^{\dagger}$ \\
SFT & 46.2\,$\pm$\,2.4 & 42.3\,$\pm$\,3.1 & 14.4\,$\pm$\,1.2 & 31.9\,$\pm$\,1.5 & 57.0\,$\pm$\,3.5 & 35.9\,$\pm$\,2.4 & 48.0\,$\pm$\,1.3 & 56.9\,$\pm$\,1.5 & 19.8\,$\pm$\,2.5 & 59.7\,$\pm$\,4.3 & 3.9$^{\dagger}$ \\
CPO & 48.9\,$\pm$\,2.4 & 58.4\,$\pm$\,3.0 & 26.6\,$\pm$\,1.6 & 45.9\,$\pm$\,1.6 & 57.6\,$\pm$\,3.5 & 56.3\,$\pm$\,2.5 & 52.8\,$\pm$\,1.3 & 55.0\,$\pm$\,1.5 & 27.1\,$\pm$\,2.8 & 66.6\,$\pm$\,4.2 & 85.5 \\
\midrule
\realname{} & \textbf{51.4\,$\pm$\,2.4} & \textbf{66.6\,$\pm$\,2.9} & \textbf{29.6\,$\pm$\,1.6} & \underline{46.6\,$\pm$\,1.6} & \textbf{61.2\,$\pm$\,3.5} & \textbf{60.3\,$\pm$\,2.5} & 70.9\,$\pm$\,1.2 & 55.6\,$\pm$\,1.5 & \textbf{29.9\,$\pm$\,2.8} & 70.9\,$\pm$\,4.0 & \underline{85.7} \\
\namel & \underline{51.1\,$\pm$\,2.4} & 63.6\,$\pm$\,3.0 & \underline{29.3\,$\pm$\,1.6} & \textbf{46.8\,$\pm$\,1.6} & 57.1\,$\pm$\,3.5 & \underline{59.9\,$\pm$\,2.5} & 67.5\,$\pm$\,1.3 & \textbf{61.1\,$\pm$\,1.5} & 28.1\,$\pm$\,2.8 & 79.6\,$\pm$\,3.6 & \textbf{87.0} \\
\bottomrule
\multicolumn{12}{l}{\footnotesize $^{\dagger}$ Below the per-domain in-domain-gain threshold; excluded from the headline forgetting comparison (Section~\ref{sec:exp-setup}).}
\end{tabular}%
}
\end{table}

\subsection{Post-Hoc Weight Merging (WiSE-FT)}
\label{app:wiseft}

\begin{figure}[t]
\centering
\includegraphics[width=0.72\linewidth]{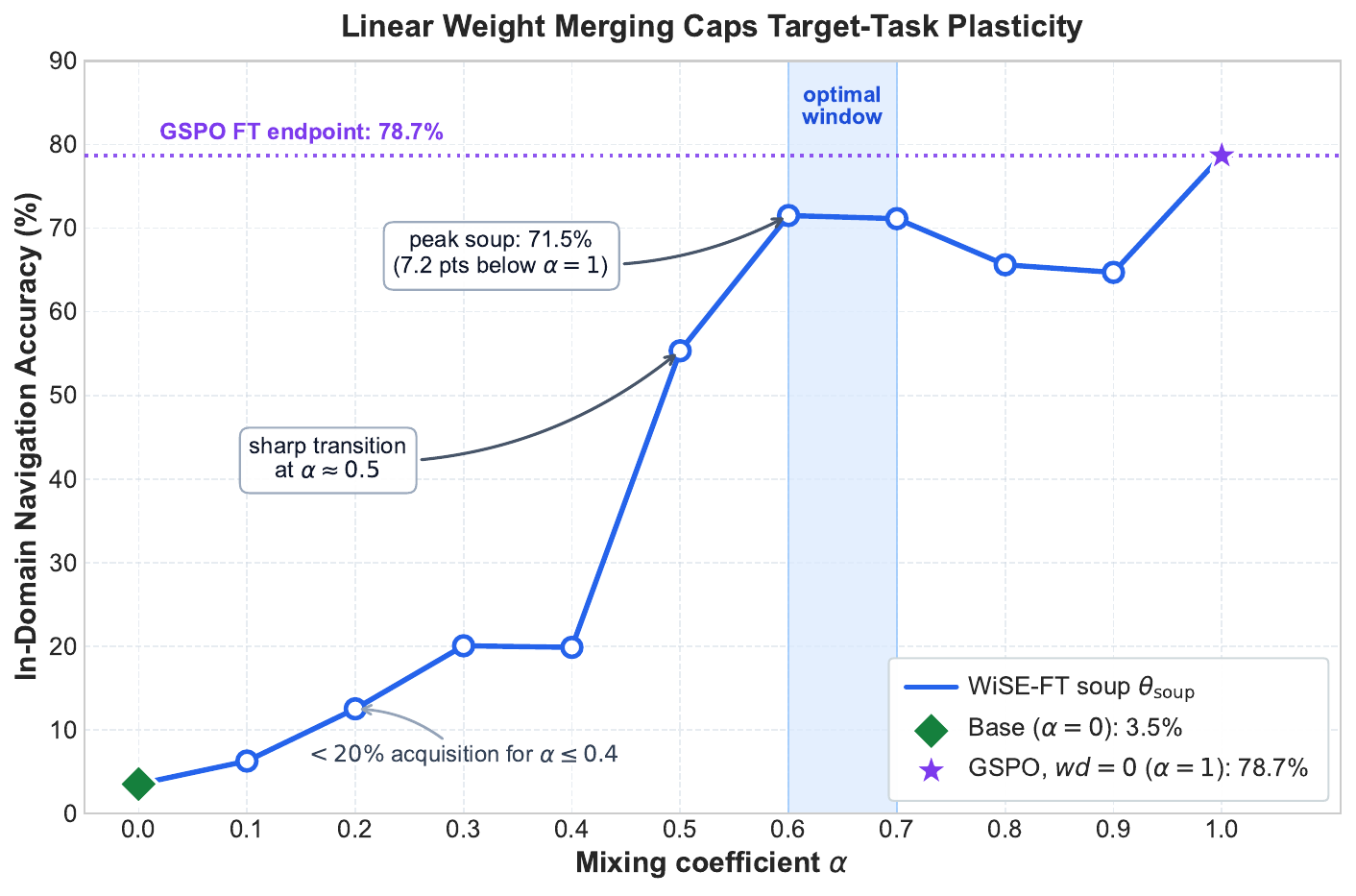}
\caption{\textbf{WiSE-FT In-Domain Navigation Accuracy vs.\ Mixing Coefficient $\alpha$.} Linear weight interpolation $\theta_{\text{soup}} = (1-\alpha)\theta_{\text{base}} + \alpha\theta_{\text{FT}}$ on Qwen3-VL-2B Navigation. The model exhibits minimal target-task acquisition for $\alpha \le 0.4$ ($<20\%$), undergoes a sharp transition at $\alpha \approx 0.5$, and plateaus at an optimal specialization window of $\alpha \in [0.6, 0.7]$ with $71.5\%$ accuracy which is still well below the $78.7\%$ reached by GSPO ($\alpha=1$), the \texttt{FT} model used in the model-souping. The dotted horizontal line marks that GSPO endpoint; no soup along the linear path reaches it.}
\label{fig:wiseft-indomain}
\end{figure}

We evaluate linear post-hoc weight interpolation (WiSE-FT; \citealp{wortsman2022robust}) on Navigation with Qwen3-VL-2B. Given the frozen base checkpoint $\theta_{\text{base}}$ and the unregularized fine-tuned policy $\theta_{\text{FT}}$ (GSPO, $wd=0$, step 600), we construct model soups along the linear path $\theta_{\text{soup}} = (1-\alpha)\theta_{\text{base}} + \alpha\theta_{\text{FT}}$ sweeping $\alpha \in \{0.1, 0.2, \dots, 0.9\}$. Figure~\ref{fig:wiseft-indomain} traces in-domain task acquisition across the sweep, and Table~\ref{tab:wiseft-nav-2b} details the full 10-benchmark evaluation profile alongside final target accuracy.

\begin{figure}[t]
\centering
\includegraphics[width=0.82\linewidth]{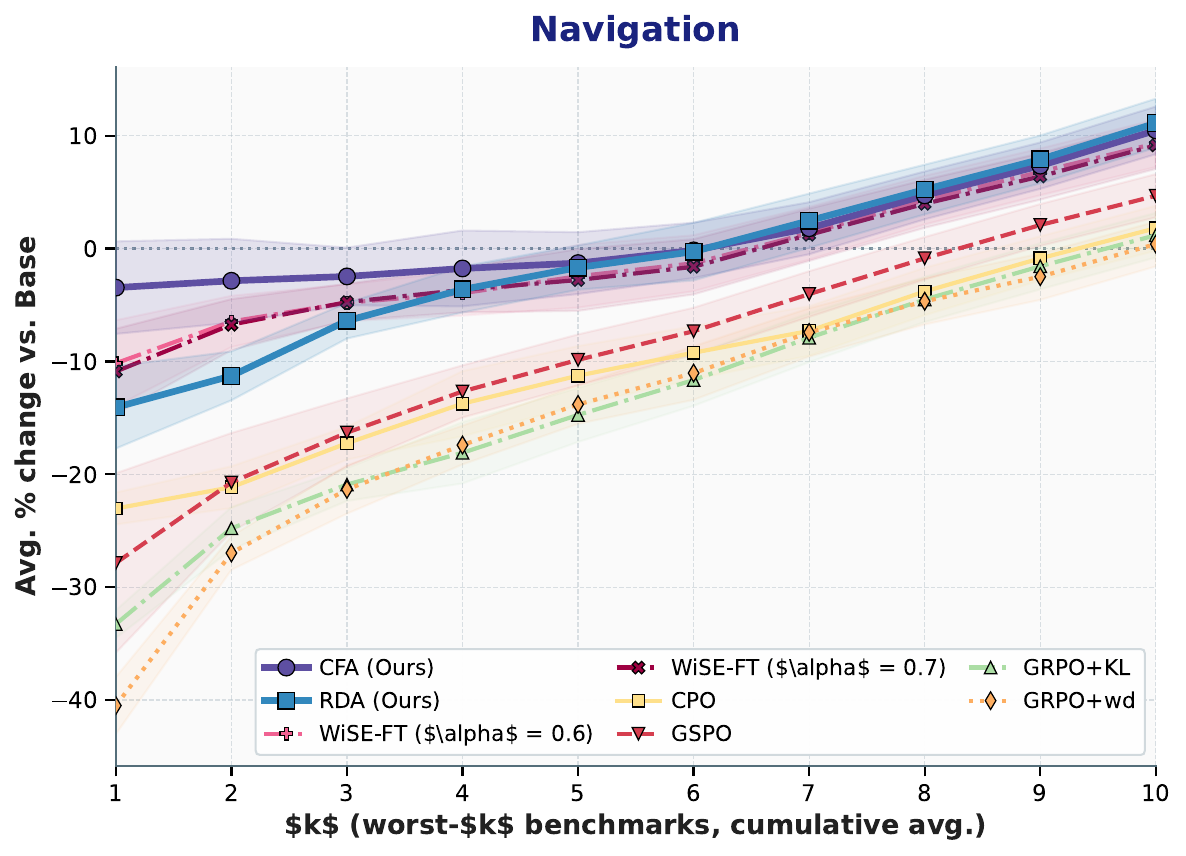}
\caption{\textbf{Worst-$k$ forgetting on Navigation (Qwen3-VL-2B): weight merging versus representation anchoring.} Mean relative change against the base model over the $k$ worst-affected out-of-domain benchmarks. At its two best mixing coefficients ($\alpha \in \{0.6, 0.7\}$), WiSE-FT improves markedly over unregularized GSPO and the parameter-movement baselines, but still concedes roughly $-10\%$ at $k=1$; \namel{} holds the same quantity to $-3.5\%$ while reaching higher in-domain accuracy (Table~\ref{tab:wiseft-nav-2b}). Shaded bands are bootstrap confidence intervals over benchmark resamples.}
\label{fig:wiseft-forgetting}
\end{figure}

\paragraph{Why Weight Merging is Suboptimal.}
While post-hoc weight merging is appealing because it incurs no training-time regularization overhead, our results demonstrate why it remains fundamentally suboptimal compared to dynamic representation anchoring:

\begin{enumerate}
    \item \textbf{Impaired Target-Domain Plasticity:} As illustrated in Figure~\ref{fig:wiseft-indomain}, weight interpolation exhibits near-zero learning of the new task for $\alpha \le 0.4$ ($<20\%$ accuracy), followed by a steep phase transition at $\alpha \approx 0.5$, before plateauing at an upper ceiling of $71.51\%$ ($\alpha=0.6$) and $71.14\%$ ($\alpha=0.7$). Linear interpolation severely truncates the model's plastic capacity on the target task, trailing representation anchoring by more than $15$ percentage points ($86.95\%$ for \namel).
    \item \textbf{Rigid Zero-Sum Parameter Trade-Off:} Linearly averaging parameter vectors implicitly assumes a convex, Euclidean landscape between pre-trained and fine-tuned checkpoints. On non-linear vision-language policy manifolds, however, linear weight mixtures perturb shared representations uniformly, enforcing an unavoidable trade-off: any parameter shift toward the base model directly dismantles newly acquired target capabilities.
    \item \textbf{Suboptimal Tail-Risk Retention:} Even at its peak specialization window ($\alpha=0.6$), WiSE-FT still incurs a $-10.12\%$ worst-1 relative drop on CountBenchQA, compared to only $-3.46\%$ for \namel{} (Figure~\ref{fig:wiseft-forgetting}). Dynamic representation anchoring evaluates the model's induced functional geometry on anchor points during training, selectively penalizing divergence in feature space while granting full optimization freedom along task-relevant gradient subspaces.
\end{enumerate}

\begin{table}[t]
\centering
\caption{\textbf{Weight Merging (WiSE-FT), Qwen3-VL-2B Navigation.} Common-suite accuracy (mean $\pm$ bootstrap std over evaluation items) for linear weight interpolation $\theta_{\mathrm{soup}} = (1-\alpha)\theta_{\mathrm{base}} + \alpha\theta_{\mathrm{FT}}$ across mixing coefficients $\alpha \in \{0.1,\dots,0.9\}$ between the base model and the GSPO ($wd=0$) checkpoint, compared with Base, unregularized GSPO, and \namel{} ($N=1280$). \textbf{Bold} marks the best method per column; \underline{underline} marks the second best. ``Target'' is final in-domain (target-task) accuracy.}
\label{tab:wiseft-nav-2b}
\resizebox{\textwidth}{!}{%
\begin{tabular}{lccccccccccc}
\toprule
 & \multicolumn{4}{c}{STEM} & \multicolumn{2}{c}{Gen. VQA} & \multicolumn{3}{c}{Chart/Doc.} & \multicolumn{1}{c}{Ground.} &  \\
\cmidrule(lr){2-5} \cmidrule(lr){6-7} \cmidrule(lr){8-10} \cmidrule(lr){11-11}
Method & MMMU & MVi & MVn & MVe & RWQA & MMS & DocVQA & CX-DQ & CX-RQ & CBQA & Target \\
\midrule
Base & 48.4\,$\pm$\,2.4 & 56.0\,$\pm$\,3.1 & 21.1\,$\pm$\,1.4 & 37.4\,$\pm$\,1.5 & 58.4\,$\pm$\,3.5 & 46.7\,$\pm$\,2.5 & 68.6\,$\pm$\,1.2 & 60.8\,$\pm$\,1.5 & 28.0\,$\pm$\,2.8 & \textbf{82.5\,$\pm$\,3.4} & 3.5 \\
GSPO (wd=0) & 51.0\,$\pm$\,2.4 & 64.8\,$\pm$\,3.0 & 27.0\,$\pm$\,1.6 & 45.4\,$\pm$\,1.6 & 59.2\,$\pm$\,3.5 & 58.7\,$\pm$\,2.5 & 67.4\,$\pm$\,1.3 & 56.2\,$\pm$\,1.5 & 20.2\,$\pm$\,2.8 & 71.3\,$\pm$\,4.0 & 78.7 \\
\midrule
WiSE-FT ($\alpha=0.1$) & 50.0\,$\pm$\,2.4 & 59.2\,$\pm$\,3.0 & 23.6\,$\pm$\,1.6 & 39.4\,$\pm$\,1.6 & 57.9\,$\pm$\,3.5 & 51.4\,$\pm$\,2.5 & 69.8\,$\pm$\,1.2 & 63.5\,$\pm$\,1.5 & 26.4\,$\pm$\,2.8 & 80.5\,$\pm$\,1.8 & 6.2 \\
WiSE-FT ($\alpha=0.2$) & 51.0\,$\pm$\,2.4 & 61.2\,$\pm$\,3.0 & 24.9\,$\pm$\,1.6 & 40.9\,$\pm$\,1.6 & 57.4\,$\pm$\,3.5 & 54.8\,$\pm$\,2.5 & \underline{70.2\,$\pm$\,1.2} & 63.1\,$\pm$\,1.5 & 27.8\,$\pm$\,2.8 & \underline{82.3\,$\pm$\,1.7} & 12.5 \\
WiSE-FT ($\alpha=0.3$) & 48.7\,$\pm$\,2.4 & 61.4\,$\pm$\,3.0 & 27.8\,$\pm$\,1.7 & 44.2\,$\pm$\,1.6 & 52.8\,$\pm$\,3.5 & 60.6\,$\pm$\,2.4 & 70.0\,$\pm$\,1.2 & \textbf{64.2\,$\pm$\,1.5} & \underline{30.0\,$\pm$\,2.9} & 81.5\,$\pm$\,1.8 & 20.0 \\
WiSE-FT ($\alpha=0.4$) & 49.1\,$\pm$\,2.4 & 61.0\,$\pm$\,3.0 & 27.5\,$\pm$\,1.7 & 44.3\,$\pm$\,1.6 & 51.2\,$\pm$\,3.5 & 60.1\,$\pm$\,2.4 & 69.3\,$\pm$\,1.2 & \textbf{64.2\,$\pm$\,1.5} & 29.7\,$\pm$\,2.9 & 82.1\,$\pm$\,1.7 & 19.9 \\
WiSE-FT ($\alpha=0.5$) & 50.5\,$\pm$\,2.4 & \textbf{68.4\,$\pm$\,2.9} & \textbf{30.2\,$\pm$\,1.7} & \textbf{47.0\,$\pm$\,1.6} & 38.3\,$\pm$\,3.4 & \textbf{61.5\,$\pm$\,2.4} & 69.1\,$\pm$\,1.2 & 62.1\,$\pm$\,1.5 & 29.0\,$\pm$\,2.9 & 78.6\,$\pm$\,1.9 & 55.3 \\
WiSE-FT ($\alpha=0.6$) & 50.6\,$\pm$\,2.4 & 65.7\,$\pm$\,2.9 & 27.9\,$\pm$\,1.7 & 46.7\,$\pm$\,1.6 & 57.6\,$\pm$\,3.5 & 59.2\,$\pm$\,2.5 & 67.6\,$\pm$\,1.3 & 59.1\,$\pm$\,1.6 & 29.0\,$\pm$\,2.9 & 74.1\,$\pm$\,2.0 & 71.5 \\
WiSE-FT ($\alpha=0.7$) & 50.5\,$\pm$\,2.4 & 66.2\,$\pm$\,2.9 & 28.3\,$\pm$\,1.7 & 46.1\,$\pm$\,1.6 & 58.2\,$\pm$\,3.5 & 58.7\,$\pm$\,2.5 & 68.1\,$\pm$\,1.2 & 59.2\,$\pm$\,1.6 & 28.2\,$\pm$\,2.8 & 73.5\,$\pm$\,2.0 & 71.1 \\
WiSE-FT ($\alpha=0.8$) & 50.6\,$\pm$\,2.4 & 66.1\,$\pm$\,2.9 & 27.8\,$\pm$\,1.7 & \underline{46.9\,$\pm$\,1.6} & 59.1\,$\pm$\,3.5 & 59.0\,$\pm$\,2.5 & 67.3\,$\pm$\,1.3 & 57.8\,$\pm$\,1.6 & \textbf{30.6\,$\pm$\,2.9} & 72.5\,$\pm$\,2.0 & 65.6 \\
WiSE-FT ($\alpha=0.9$) & 48.8\,$\pm$\,2.4 & 65.5\,$\pm$\,2.9 & 28.1\,$\pm$\,1.7 & 46.5\,$\pm$\,1.6 & \underline{59.5\,$\pm$\,3.5} & 59.1\,$\pm$\,2.5 & 67.2\,$\pm$\,1.3 & 57.1\,$\pm$\,1.6 & 26.8\,$\pm$\,2.8 & 71.1\,$\pm$\,2.0 & 64.7 \\
\midrule
\realname{} & \textbf{51.4\,$\pm$\,2.4} & \underline{66.6\,$\pm$\,2.9} & \underline{29.6\,$\pm$\,1.6} & 46.6\,$\pm$\,1.6 & \textbf{61.2\,$\pm$\,3.5} & \underline{60.3\,$\pm$\,2.5} & \textbf{70.9\,$\pm$\,1.2} & 55.6\,$\pm$\,1.5 & 29.9\,$\pm$\,2.8 & 70.9\,$\pm$\,4.0 & \underline{85.7} \\
\namel & \underline{51.1\,$\pm$\,2.4} & 63.6\,$\pm$\,3.0 & 29.3\,$\pm$\,1.6 & 46.8\,$\pm$\,1.6 & 57.1\,$\pm$\,3.5 & 59.9\,$\pm$\,2.5 & 67.5\,$\pm$\,1.3 & \underline{61.1\,$\pm$\,1.5} & 28.1\,$\pm$\,2.8 & 79.6\,$\pm$\,3.6 & \textbf{87.0} \\
\bottomrule
\end{tabular}%
}
\end{table}

\begin{table}[t]
\centering
\caption{\textbf{MedBookVQA, Qwen3-VL-2B.} Common-suite accuracy (mean $\pm$ bootstrap std over evaluation items) for every baseline and for \namel{} at anchor budget $N=1280$ (Real/Content-Free). \textbf{Bold} marks the best method per column; \underline{underline} marks the second best. ``Target'' is final in-domain (target-task) accuracy.}
\label{tab:med-2b-full}
\resizebox{\textwidth}{!}{%
\begin{tabular}{lccccccccccc}
\toprule
 & \multicolumn{4}{c}{STEM} & \multicolumn{2}{c}{Gen. VQA} & \multicolumn{3}{c}{Chart/Doc.} & \multicolumn{1}{c}{Ground.} &  \\
\cmidrule(lr){2-5} \cmidrule(lr){6-7} \cmidrule(lr){8-10} \cmidrule(lr){11-11}
Method & MMMU & MVi & MVn & MVe & RWQA & MMS & DocVQA & CX-DQ & CX-RQ & CBQA & Target \\
\midrule
Base & 48.4\,$\pm$\,2.4 & 56.0\,$\pm$\,3.1 & 21.1\,$\pm$\,1.4 & 37.4\,$\pm$\,1.5 & 58.4\,$\pm$\,3.5 & 46.7\,$\pm$\,2.5 & 68.6\,$\pm$\,1.2 & 60.8\,$\pm$\,1.5 & 28.0\,$\pm$\,2.8 & 82.5\,$\pm$\,3.4 & 49.4 \\
GRPO (wd=0) & 37.3\,$\pm$\,2.3 & 45.4\,$\pm$\,3.1 & 15.6\,$\pm$\,1.3 & 28.4\,$\pm$\,1.4 & 40.1\,$\pm$\,3.5 & 36.2\,$\pm$\,2.4 & 24.7\,$\pm$\,1.2 & 38.1\,$\pm$\,1.5 & 16.3\,$\pm$\,2.3 & 26.9\,$\pm$\,3.9 & \textbf{99.8} \\
GRPO (wd=0.1) & 47.8\,$\pm$\,2.4 & 58.1\,$\pm$\,3.1 & 27.4\,$\pm$\,1.6 & 44.8\,$\pm$\,1.6 & \underline{59.5\,$\pm$\,3.5} & 58.5\,$\pm$\,2.5 & 65.7\,$\pm$\,1.3 & 55.1\,$\pm$\,1.5 & 27.6\,$\pm$\,2.8 & 56.6\,$\pm$\,4.4 & 94.7 \\
GRPO+KL & 44.1\,$\pm$\,2.3 & \underline{63.2\,$\pm$\,3.0} & 26.4\,$\pm$\,1.6 & 41.2\,$\pm$\,1.5 & 53.2\,$\pm$\,3.5 & 57.7\,$\pm$\,2.5 & 43.9\,$\pm$\,1.3 & 50.4\,$\pm$\,1.6 & 26.8\,$\pm$\,2.8 & 74.8\,$\pm$\,3.9 & \underline{99.2} \\
GSPO & 34.4\,$\pm$\,2.2 & 49.1\,$\pm$\,3.1 & 17.5\,$\pm$\,1.4 & 27.4\,$\pm$\,1.4 & 38.8\,$\pm$\,3.5 & 45.1\,$\pm$\,2.5 & 34.4\,$\pm$\,1.3 & 37.8\,$\pm$\,1.5 & 14.1\,$\pm$\,2.2 & 58.5\,$\pm$\,4.4 & 99.0 \\
GSPO+EWC & \underline{49.2\,$\pm$\,2.4} & 60.6\,$\pm$\,3.0 & 25.0\,$\pm$\,1.5 & 39.0\,$\pm$\,1.5 & 58.3\,$\pm$\,3.5 & \underline{59.4\,$\pm$\,2.5} & 68.5\,$\pm$\,1.2 & \underline{61.1\,$\pm$\,1.5} & \textbf{30.3\,$\pm$\,2.9} & \underline{84.3\,$\pm$\,3.2} & 84.5$^{\dagger}$ \\
LoRA & 49.1\,$\pm$\,2.4 & 58.3\,$\pm$\,3.1 & 23.8\,$\pm$\,1.5 & 38.9\,$\pm$\,1.5 & \textbf{61.6\,$\pm$\,3.5} & 48.7\,$\pm$\,2.5 & 66.0\,$\pm$\,1.3 & \textbf{62.8\,$\pm$\,1.5} & 27.8\,$\pm$\,2.8 & 79.4\,$\pm$\,3.6 & 50.8$^{\dagger}$ \\
SFT & 37.9\,$\pm$\,2.3 & 36.6\,$\pm$\,3.0 & 9.5\,$\pm$\,1.0 & 16.2\,$\pm$\,1.1 & 41.4\,$\pm$\,3.5 & 34.7\,$\pm$\,2.4 & 26.2\,$\pm$\,1.2 & 42.8\,$\pm$\,1.5 & 19.4\,$\pm$\,2.5 & 17.5\,$\pm$\,3.4 & 95.1 \\
CPO & 33.1\,$\pm$\,2.2 & 43.6\,$\pm$\,3.1 & 14.8\,$\pm$\,1.3 & 21.2\,$\pm$\,1.3 & 35.8\,$\pm$\,3.4 & 39.9\,$\pm$\,2.5 & 29.9\,$\pm$\,1.2 & 34.7\,$\pm$\,1.5 & 15.2\,$\pm$\,2.2 & 59.9\,$\pm$\,4.3 & 96.8 \\
\midrule
\realname{} & 49.1\,$\pm$\,2.4 & 60.3\,$\pm$\,3.0 & \textbf{29.6\,$\pm$\,1.6} & \textbf{45.8\,$\pm$\,1.6} & 53.6\,$\pm$\,3.5 & 59.1\,$\pm$\,2.5 & \underline{69.1\,$\pm$\,1.2} & 59.5\,$\pm$\,1.5 & 28.4\,$\pm$\,2.8 & 83.7\,$\pm$\,3.3 & 96.2 \\
\namel & \textbf{50.8\,$\pm$\,2.4} & \textbf{64.7\,$\pm$\,3.0} & \underline{29.3\,$\pm$\,1.6} & \underline{45.4\,$\pm$\,1.6} & 58.9\,$\pm$\,3.5 & \textbf{60.1\,$\pm$\,2.5} & \textbf{70.9\,$\pm$\,1.2} & 60.6\,$\pm$\,1.5 & \underline{29.9\,$\pm$\,2.8} & \textbf{87.2\,$\pm$\,3.0} & 98.5 \\
\bottomrule
\multicolumn{12}{l}{\footnotesize $^{\dagger}$ Below the per-domain in-domain-gain threshold; excluded from the headline forgetting comparison (Section~\ref{sec:exp-setup}).}
\end{tabular}%
}
\end{table}

\begin{table}[t]
\centering
\caption{\textbf{Puzzle, Qwen3-VL-2B.} Common-suite accuracy (mean $\pm$ bootstrap std over evaluation items) for every baseline and for \namel{} at anchor budget $N=1280$ (Real/Content-Free). \textbf{Bold} marks the best method per column; \underline{underline} marks the second best. ``Target'' is final in-domain (target-task) accuracy. This table illustrates the averaging problem of Section~\ref{sec:metric}: GRPO (wd$=0.1$) posts a suite-average change of $-0.7\%$ while losing $-24.8\%$ on its worst benchmark, and GSPO~+~EWC averages $+3.5\%$ while losing $-25.8\%$; \namel{} is the only method whose worst benchmark stays in single digits ($-9.1\%$, suite average $+7.3\%$). Note also that CPO's apparently strong retention here is accompanied by an in-domain gain of only $+39.0$ points against a median RL gain of $+62.2$ ($63\%$), placing it below the per-domain gain threshold of Section~\ref{sec:exp-setup}.}
\label{tab:puz-2b-full}
\resizebox{\textwidth}{!}{%
\begin{tabular}{lccccccccccc}
\toprule
 & \multicolumn{4}{c}{STEM} & \multicolumn{2}{c}{Gen. VQA} & \multicolumn{3}{c}{Chart/Doc.} & \multicolumn{1}{c}{Ground.} &  \\
\cmidrule(lr){2-5} \cmidrule(lr){6-7} \cmidrule(lr){8-10} \cmidrule(lr){11-11}
Method & MMMU & MVi & MVn & MVe & RWQA & MMS & DocVQA & CX-DQ & CX-RQ & CBQA & Target \\
\midrule
Base & \underline{48.4\,$\pm$\,2.4} & 56.0\,$\pm$\,3.1 & 21.1\,$\pm$\,1.4 & 37.4\,$\pm$\,1.5 & 58.4\,$\pm$\,3.5 & 46.7\,$\pm$\,2.5 & \underline{68.6\,$\pm$\,1.2} & 60.8\,$\pm$\,1.5 & 28.0\,$\pm$\,2.8 & \textbf{82.5\,$\pm$\,3.4} & 23.8 \\
GRPO (wd=0) & 44.9\,$\pm$\,2.3 & 58.5\,$\pm$\,3.0 & 16.7\,$\pm$\,1.3 & 34.5\,$\pm$\,1.5 & \underline{59.4\,$\pm$\,3.5} & 54.9\,$\pm$\,2.5 & 54.1\,$\pm$\,1.3 & 56.5\,$\pm$\,1.5 & 27.1\,$\pm$\,2.8 & 60.3\,$\pm$\,4.3 & 85.0 \\
GRPO (wd=0.1) & 46.1\,$\pm$\,2.4 & 62.4\,$\pm$\,3.0 & 21.1\,$\pm$\,1.4 & 39.6\,$\pm$\,1.5 & 54.8\,$\pm$\,3.5 & 54.6\,$\pm$\,2.5 & 51.6\,$\pm$\,1.3 & \underline{64.8\,$\pm$\,1.5} & 26.9\,$\pm$\,2.8 & 75.6\,$\pm$\,3.8 & 82.2 \\
GRPO+KL & 45.1\,$\pm$\,1.2 & 57.6\,$\pm$\,1.6 & 22.3\,$\pm$\,0.8 & 40.1\,$\pm$\,0.8 & 47.8\,$\pm$\,1.8 & 53.3\,$\pm$\,1.3 & 52.9\,$\pm$\,0.7 & 51.3\,$\pm$\,0.8 & 24.7\,$\pm$\,1.4 & 59.1\,$\pm$\,2.2 & 86.0 \\
GSPO & 46.4\,$\pm$\,2.4 & 58.1\,$\pm$\,3.1 & 17.0\,$\pm$\,1.3 & 38.4\,$\pm$\,1.5 & 51.9\,$\pm$\,3.5 & 53.3\,$\pm$\,2.5 & 52.1\,$\pm$\,1.3 & 56.5\,$\pm$\,1.5 & 25.4\,$\pm$\,2.7 & 49.7\,$\pm$\,4.4 & 87.0 \\
GSPO+EWC & 48.4\,$\pm$\,2.4 & \underline{63.7\,$\pm$\,3.0} & \underline{25.2\,$\pm$\,1.5} & \underline{43.0\,$\pm$\,1.6} & \textbf{59.5\,$\pm$\,3.5} & 55.1\,$\pm$\,2.5 & 65.1\,$\pm$\,1.3 & 63.8\,$\pm$\,1.5 & 25.9\,$\pm$\,2.7 & 61.2\,$\pm$\,4.3 & \underline{88.2} \\
LoRA & 46.3\,$\pm$\,2.4 & 52.2\,$\pm$\,3.1 & 17.3\,$\pm$\,1.4 & 33.3\,$\pm$\,1.5 & 58.4\,$\pm$\,3.5 & 49.3\,$\pm$\,2.5 & 57.3\,$\pm$\,1.3 & 62.9\,$\pm$\,1.5 & 23.4\,$\pm$\,2.6 & 44.4\,$\pm$\,4.4 & 48.0$^{\dagger}$ \\
SFT & 46.0\,$\pm$\,2.4 & 39.7\,$\pm$\,3.0 & 10.9\,$\pm$\,1.1 & 29.6\,$\pm$\,1.4 & 47.6\,$\pm$\,3.5 & 38.8\,$\pm$\,2.5 & 27.5\,$\pm$\,1.2 & 59.1\,$\pm$\,1.5 & 20.7\,$\pm$\,2.5 & 14.7\,$\pm$\,3.1 & 33.0$^{\dagger}$ \\
CPO & \textbf{53.3\,$\pm$\,2.4} & \textbf{76.8\,$\pm$\,2.6} & \textbf{30.7\,$\pm$\,1.6} & \textbf{46.4\,$\pm$\,1.6} & 59.2\,$\pm$\,3.5 & \textbf{63.8\,$\pm$\,2.4} & 67.4\,$\pm$\,1.3 & \textbf{65.5\,$\pm$\,1.5} & \textbf{43.6\,$\pm$\,3.1} & \underline{82.1\,$\pm$\,3.4} & 62.8$^{\dagger}$ \\
\midrule
\realname{} & 46.8\,$\pm$\,2.4 & 62.4\,$\pm$\,3.0 & 23.5\,$\pm$\,1.5 & 41.8\,$\pm$\,1.5 & 58.8\,$\pm$\,3.5 & 54.1\,$\pm$\,2.5 & 62.5\,$\pm$\,1.3 & 63.1\,$\pm$\,1.5 & 27.0\,$\pm$\,2.8 & 65.8\,$\pm$\,4.2 & 82.5 \\
\namel & 48.3\,$\pm$\,2.4 & 63.2\,$\pm$\,3.0 & 23.0\,$\pm$\,1.4 & 40.2\,$\pm$\,1.5 & 58.8\,$\pm$\,3.5 & \underline{55.7\,$\pm$\,2.5} & \textbf{68.9\,$\pm$\,1.3} & 64.3\,$\pm$\,1.6 & \underline{35.5\,$\pm$\,2.7} & 75.0\,$\pm$\,3.8 & \textbf{90.8} \\
\bottomrule
\multicolumn{12}{l}{\footnotesize $^{\dagger}$ Below the per-domain in-domain-gain threshold; excluded from the headline forgetting comparison (Section~\ref{sec:exp-setup}).}
\end{tabular}%
}
\end{table}

\begin{table*}[t]
\centering
\caption{{Anchor-Budget Ablation across Out-of-Domain Benchmarks (Navigation, Qwen3-VL-2B).}
Evaluation of \namel{} with curated real-image coreset anchors (from external public training splits) versus random pseudo-inputs (Content-Free) across scaling budgets $N \in \{320, 640, 1280, 2560\}$.
We report accuracy (mean $\pm$ bootstrap std over evaluation items) across all 10 common-suite evaluation benchmarks alongside final target in-domain accuracy.
\textbf{Bold} indicates the best performance per benchmark; \underline{underline} indicates the second best.}
\label{tab:anchor_budget_ablation_full}
\resizebox{\textwidth}{!}{%
\begin{tabular}{llccccccccccc}
\toprule
 & & \multicolumn{4}{c}{{STEM}} & \multicolumn{2}{c}{{Gen. VQA}} & \multicolumn{3}{c}{{Chart/Doc.}} & \multicolumn{1}{c}{{Ground.}} & \\
\cmidrule(lr){3-6} \cmidrule(lr){7-8} \cmidrule(lr){9-11} \cmidrule(lr){12-12}
Budget ($N$) & Anchors & MMMU & MVi & MVn & MVe & RWQA & MMS & DocVQA & CX-DQ & CX-RQ & CBQA & Target \\
\midrule
\multicolumn{2}{l}{\textit{Base}} & 48.4\,$\pm$\,2.4 & 56.0\,$\pm$\,3.1 & 21.1\,$\pm$\,1.4 & 37.4\,$\pm$\,1.5 & 58.4\,$\pm$\,3.5 & 46.7\,$\pm$\,2.5 & 68.6\,$\pm$\,1.2 & 60.8\,$\pm$\,1.5 & 28.0\,$\pm$\,2.8 & 82.5\,$\pm$\,3.4 & 3.5 \\
\multirow{2}{*}{$N=320$}
 & \realname{} & 50.4\,$\pm$\,2.4 & 67.5\,$\pm$\,2.9 & 26.1\,$\pm$\,1.6 & 47.0\,$\pm$\,1.6 & 59.5\,$\pm$\,3.5 & 59.9\,$\pm$\,2.5 & 70.2\,$\pm$\,1.2 & 56.3\,$\pm$\,1.5 & \textbf{30.7\,$\pm$\,2.9} & 74.8\,$\pm$\,3.8 & 87.3 \\
 & \namel & 50.6\,$\pm$\,2.4 & 62.7\,$\pm$\,3.0 & 28.4\,$\pm$\,1.6 & 45.8\,$\pm$\,1.6 & 49.9\,$\pm$\,3.5 & 60.3\,$\pm$\,2.5 & 63.5\,$\pm$\,1.3 & 59.2\,$\pm$\,1.5 & 28.9\,$\pm$\,2.8 & 71.5\,$\pm$\,4.0 & 83.8 \\
\addlinespace
\multirow{2}{*}{$N=640$}
 & \realname{} & 48.9\,$\pm$\,2.4 & 64.1\,$\pm$\,3.0 & 25.9\,$\pm$\,1.6 & 43.9\,$\pm$\,1.6 & 55.2\,$\pm$\,3.5 & 59.6\,$\pm$\,2.5 & 59.0\,$\pm$\,1.3 & 58.8\,$\pm$\,1.5 & 28.5\,$\pm$\,2.8 & 61.5\,$\pm$\,4.3 & 77.6 \\
 & \namel & 51.2\,$\pm$\,2.4 & \textbf{67.7\,$\pm$\,2.9} & \textbf{30.0\,$\pm$\,1.6} & \textbf{47.1\,$\pm$\,1.6} & 54.5\,$\pm$\,3.5 & \textbf{61.3\,$\pm$\,2.5} & 69.1\,$\pm$\,1.2 & 55.6\,$\pm$\,1.5 & 28.9\,$\pm$\,2.8 & 81.3\,$\pm$\,3.5 & 86.4 \\
\addlinespace
\multirow{2}{*}{$N=1280$}
 & \realname{} & \textbf{51.4\,$\pm$\,2.4} & 66.6\,$\pm$\,2.9 & 29.6\,$\pm$\,1.6 & 46.6\,$\pm$\,1.6 & \textbf{61.2\,$\pm$\,3.5} & 60.3\,$\pm$\,2.5 & \textbf{70.9\,$\pm$\,1.2} & 55.6\,$\pm$\,1.5 & \underline{29.9\,$\pm$\,2.8} & 70.9\,$\pm$\,4.0 & 85.7 \\
 & \namel & 51.1\,$\pm$\,2.4 & 63.6\,$\pm$\,3.0 & 29.3\,$\pm$\,1.6 & 46.8\,$\pm$\,1.6 & 57.1\,$\pm$\,3.5 & 59.9\,$\pm$\,2.5 & 67.5\,$\pm$\,1.3 & 61.1\,$\pm$\,1.5 & 28.1\,$\pm$\,2.8 & 79.6\,$\pm$\,3.6 & 87.0 \\
\addlinespace
\multirow{2}{*}{$N=2560$}
 & \realname{} & 49.0\,$\pm$\,2.4 & 64.5\,$\pm$\,3.0 & 27.5\,$\pm$\,1.6 & 44.9\,$\pm$\,1.6 & 59.1\,$\pm$\,3.5 & 59.3\,$\pm$\,2.5 & 59.5\,$\pm$\,1.3 & 56.8\,$\pm$\,1.5 & 27.6\,$\pm$\,2.8 & 74.5\,$\pm$\,3.8 & 82.7 \\
 & \namel & \underline{52.4\,$\pm$\,2.4} & \underline{65.7\,$\pm$\,2.9} & \underline{29.3\,$\pm$\,1.6} & \underline{47.9\,$\pm$\,1.6} & \underline{59.7\,$\pm$\,3.5} & \underline{60.1\,$\pm$\,2.5} & \underline{71.2\,$\pm$\,1.2} & \textbf{67.0\,$\pm$\,1.5} & 29.0\,$\pm$\,2.8 & \textbf{79.8\,$\pm$\,3.6} & \textbf{87.5} \\
\bottomrule
\end{tabular}%
}
\end{table*}

\begin{table*}[t]
\centering
\caption{{Divergence-Measure Ablation across Out-of-Domain Benchmarks
(Navigation, Qwen3-VL-2B, $N=1280$).} Comparison of regularizer
objectives at the anchor budget used throughout this paper: our
proposed {\namel}/\realname{} (varentropy-weighted MMD), {unweighted MMD},
{Fisher-weighted KL divergence}, and {unweighted reference KL divergence},
crossed with Real vs.\ Content-Free (Random) pseudo-inputs.
\textbf{Bold} indicates top performance per column; \underline{underline}
indicates the second best.}
\label{tab:divergence_ablation_full}
\resizebox{\textwidth}{!}{%
\begin{tabular}{llccccccccccc}
\toprule
 & & \multicolumn{4}{c}{{STEM}} & \multicolumn{2}{c}{{Gen. VQA}} & \multicolumn{3}{c}{{Chart/Doc.}} & \multicolumn{1}{c}{\textbf{Ground.}} & \\
\cmidrule(lr){3-6} \cmidrule(lr){7-8} \cmidrule(lr){9-11} \cmidrule(lr){12-12}
{Divergence} & {Anchors} & {MMMU} & {MVi} & {MVn} & {MVe} & {RWQA} & {MMS} & {DocVQA} & {CX-DQ} & {CX-RQ} & {CBQA} & {Target} \\
\midrule
\multicolumn{2}{l}{\textit{Base}} & 48.4\,$\pm$\,2.4 & 56.0\,$\pm$\,3.1 & 21.1\,$\pm$\,1.4 & 37.4\,$\pm$\,1.5 & 58.4\,$\pm$\,3.5 & 46.7\,$\pm$\,2.5 & 68.6\,$\pm$\,1.2 & 60.8\,$\pm$\,1.5 & 28.0\,$\pm$\,2.8 & 82.5\,$\pm$\,3.4 & 3.5 \\
\midrule
\textbf{\realname{} (Ours)} & Real & \textbf{51.4\,$\pm$\,2.4} & \textbf{66.6\,$\pm$\,2.9} & \underline{29.6\,$\pm$\,1.6} & 46.6\,$\pm$\,1.6 & \textbf{61.2\,$\pm$\,3.5} & \underline{60.3\,$\pm$\,2.5} & \textbf{70.9\,$\pm$\,1.2} & 55.6\,$\pm$\,1.5 & 29.9\,$\pm$\,2.8 & 70.9\,$\pm$\,4.0 & 85.7 \\
\textbf{\namel{} (Ours)} & Random & 51.1\,$\pm$\,2.4 & 63.6\,$\pm$\,3.0 & 29.3\,$\pm$\,1.6 & \underline{46.8\,$\pm$\,1.6} & 57.1\,$\pm$\,3.5 & 59.9\,$\pm$\,2.5 & 67.5\,$\pm$\,1.3 & \underline{61.2\,$\pm$\,1.5} & 28.1\,$\pm$\,2.8 & \underline{79.6\,$\pm$\,3.6} & 87.0 \\
\cmidrule(lr){1-13}
Unweighted-MMD & Real & 50.4\,$\pm$\,2.4 & 62.3\,$\pm$\,3.0 & 26.2\,$\pm$\,1.6 & 44.0\,$\pm$\,1.3 & 55.0\,$\pm$\,3.5 & 59.3\,$\pm$\,2.5 & 66.3\,$\pm$\,1.2 & 55.6\,$\pm$\,1.5 & 26.8\,$\pm$\,2.8 & 69.3\,$\pm$\,4.1 & 85.1 \\
Unweighted-MMD & Random & 50.5\,$\pm$\,2.4 & 65.4\,$\pm$\,3.0 & 26.8\,$\pm$\,1.6 & 45.0\,$\pm$\,1.6 & 52.7\,$\pm$\,3.5 & 60.0\,$\pm$\,2.5 & \underline{67.9\,$\pm$\,1.3} & 60.4\,$\pm$\,1.5 & 28.9\,$\pm$\,2.8 & 72.9\,$\pm$\,3.7 & 88.1 \\
\cmidrule(lr){1-13}
Fisher-KL & Real & 47.8\,$\pm$\,2.4 & 62.3\,$\pm$\,3.0 & 26.3\,$\pm$\,1.6 & 45.1\,$\pm$\,1.6 & 52.7\,$\pm$\,3.5 & \textbf{60.7\,$\pm$\,2.5} & 58.3\,$\pm$\,1.3 & 60.1\,$\pm$\,1.5 & \textbf{31.4\,$\pm$\,2.9} & 77.6\,$\pm$\,3.7 & 88.8 \\
Fisher-KL & Random & 50.8\,$\pm$\,2.4 & 65.0\,$\pm$\,3.0 & \textbf{30.7\,$\pm$\,1.6} & \textbf{48.5\,$\pm$\,1.6} & 55.0\,$\pm$\,3.5 & 60.0\,$\pm$\,2.5 & 64.9\,$\pm$\,1.3 & \textbf{65.6\,$\pm$\,1.5} & \underline{30.5\,$\pm$\,2.9} & \textbf{81.7\,$\pm$\,3.4} & 91.9 \\
\cmidrule(lr){1-13}
Unweighted-KL & Real & 46.9\,$\pm$\,2.4 & 62.4\,$\pm$\,3.0 & 27.2\,$\pm$\,1.6 & 43.4\,$\pm$\,1.6 & 53.9\,$\pm$\,3.5 & 59.4\,$\pm$\,2.5 & 60.0\,$\pm$\,1.3 & 58.5\,$\pm$\,1.5 & 28.8\,$\pm$\,2.8 & 69.9\,$\pm$\,4.0 & 82.9 \\
Unweighted-KL & Random & \underline{51.4\,$\pm$\,2.4} & \underline{65.7\,$\pm$\,3.0} & 29.5\,$\pm$\,1.4 & 44.0\,$\pm$\,1.3 & \underline{59.3\,$\pm$\,3.5} & 59.9\,$\pm$\,2.5 & \underline{67.9\,$\pm$\,1.3} & 56.8\,$\pm$\,1.5 & 29.3\,$\pm$\,2.8 & 70.5\,$\pm$\,3.6 & 84.2 \\
\bottomrule
\end{tabular}%
}
\end{table*}

\begin{table}[t]
\centering
\caption{\textbf{Navigation, Qwen3-VL-4B.} Common-suite accuracy (mean $\pm$ bootstrap std over evaluation items) for every baseline and for \namel{} at anchor budget $N=1280$ (Real/Content-Free). \textbf{Bold} marks the best method per column; \underline{underline} marks the second best. ``Target'' is final in-domain (target-task) accuracy.}
\label{tab:nav-4b-full}
\resizebox{\textwidth}{!}{%
\begin{tabular}{lccccccccccc}
\toprule
 & \multicolumn{4}{c}{STEM} & \multicolumn{2}{c}{Gen. VQA} & \multicolumn{3}{c}{Chart/Doc.} & \multicolumn{1}{c}{Ground.} &  \\
\cmidrule(lr){2-5} \cmidrule(lr){6-7} \cmidrule(lr){8-10} \cmidrule(lr){11-11}
Method & MMMU & MVi & MVn & MVe & RWQA & MMS & DocVQA & CX-DQ & CX-RQ & CBQA & Target \\
\midrule
Base & 62.8\,$\pm$\,1.2 & 61.9\,$\pm$\,1.5 & 39.6\,$\pm$\,0.9 & 48.6\,$\pm$\,0.8 & 68.5\,$\pm$\,1.7 & 64.7\,$\pm$\,1.2 & 64.4\,$\pm$\,0.6 & 69.8\,$\pm$\,0.7 & 35.1\,$\pm$\,1.5 & 85.5\,$\pm$\,1.6 & 27.6 \\
GRPO (wd=0) & 54.4\,$\pm$\,1.2 & 59.3\,$\pm$\,1.6 & 28.8\,$\pm$\,0.8 & 43.7\,$\pm$\,0.8 & 68.1\,$\pm$\,1.7 & 65.1\,$\pm$\,1.2 & 76.4\,$\pm$\,0.6 & 74.0\,$\pm$\,0.7 & 37.7\,$\pm$\,1.5 & 82.5\,$\pm$\,1.7 & 82.2 \\
GRPO (wd=0.1) & \underline{64.2\,$\pm$\,1.2} & 69.0\,$\pm$\,1.5 & 43.6\,$\pm$\,0.9 & 53.8\,$\pm$\,0.8 & 68.5\,$\pm$\,1.7 & 66.4\,$\pm$\,1.2 & \underline{78.3\,$\pm$\,0.6} & \textbf{78.9\,$\pm$\,0.6} & \underline{40.2\,$\pm$\,1.6} & 84.5\,$\pm$\,1.6 & 93.4 \\
GRPO+KL & 62.6\,$\pm$\,1.2 & \textbf{74.4\,$\pm$\,1.4} & \textbf{49.2\,$\pm$\,0.9} & \textbf{57.6\,$\pm$\,0.8} & 64.2\,$\pm$\,1.7 & 66.8\,$\pm$\,1.2 & 77.7\,$\pm$\,0.6 & \underline{78.0\,$\pm$\,0.7} & 38.4\,$\pm$\,1.5 & 81.5\,$\pm$\,1.8 & \underline{93.9} \\
GSPO & 64.0\,$\pm$\,1.2 & 68.8\,$\pm$\,1.5 & \underline{43.8\,$\pm$\,0.9} & 53.6\,$\pm$\,0.8 & 68.1\,$\pm$\,1.7 & \textbf{67.7\,$\pm$\,1.2} & 73.4\,$\pm$\,0.6 & 75.2\,$\pm$\,0.7 & \textbf{41.3\,$\pm$\,1.6} & 82.6\,$\pm$\,1.5 & 92.3 \\
GSPO+EWC & 63.0\,$\pm$\,1.2 & 63.4\,$\pm$\,1.5 & 40.4\,$\pm$\,0.9 & 51.0\,$\pm$\,0.8 & 68.5\,$\pm$\,1.7 & 66.3\,$\pm$\,1.2 & 70.8\,$\pm$\,0.6 & 70.4\,$\pm$\,0.7 & 37.2\,$\pm$\,1.5 & 84.9\,$\pm$\,1.6 & 93.6 \\
LoRA & 62.4\,$\pm$\,2.3 & 68.4\,$\pm$\,2.9 & 38.2\,$\pm$\,1.7 & 53.9\,$\pm$\,1.6 & 66.9\,$\pm$\,3.3 & 64.7\,$\pm$\,2.4 & 70.5\,$\pm$\,1.2 & 72.0\,$\pm$\,1.4 & 38.8\,$\pm$\,3.0 & 84.7\,$\pm$\,3.2 & 29.6$^{\dagger}$ \\
SFT & 58.3\,$\pm$\,2.3 & \underline{72.8\,$\pm$\,2.8} & 39.3\,$\pm$\,1.7 & \underline{54.0\,$\pm$\,1.6} & 67.8\,$\pm$\,3.3 & 64.9\,$\pm$\,2.4 & 74.5\,$\pm$\,1.2 & 74.3\,$\pm$\,1.4 & 37.3\,$\pm$\,3.0 & 83.1\,$\pm$\,3.3 & 23.9$^{\dagger}$ \\
CPO & 56.8\,$\pm$\,2.3 & 65.7\,$\pm$\,2.9 & 36.5\,$\pm$\,1.7 & 51.6\,$\pm$\,1.6 & \textbf{68.8\,$\pm$\,3.3} & 66.1\,$\pm$\,2.4 & \textbf{78.8\,$\pm$\,1.1} & 76.4\,$\pm$\,1.3 & 37.4\,$\pm$\,3.0 & 84.5\,$\pm$\,3.2 & 66.5$^{\dagger}$ \\
\midrule
\realname{} & \textbf{65.0\,$\pm$\,1.2} & 62.6\,$\pm$\,1.5 & 43.5\,$\pm$\,0.9 & 50.6\,$\pm$\,0.8 & \underline{68.4\,$\pm$\,1.7} & \underline{67.3\,$\pm$\,1.2} & 67.7\,$\pm$\,0.6 & 73.4\,$\pm$\,0.7 & 38.4\,$\pm$\,1.5 & \textbf{88.2\,$\pm$\,1.5} & 91.9 \\
\namel & 62.9\,$\pm$\,1.2 & 64.2\,$\pm$\,1.5 & 42.5\,$\pm$\,0.9 & 51.2\,$\pm$\,0.8 & \underline{68.4\,$\pm$\,1.7} & 66.7\,$\pm$\,1.2 & 69.4\,$\pm$\,0.6 & 73.7\,$\pm$\,0.7 & 37.7\,$\pm$\,1.5 & \underline{87.2\,$\pm$\,1.5} & \textbf{94.3} \\
\bottomrule
\multicolumn{12}{l}{\footnotesize $^{\dagger}$ Below the per-domain in-domain-gain threshold; excluded from the headline forgetting comparison (Section~\ref{sec:exp-setup}).}
\end{tabular}%
}
\end{table}

\begin{table}[t]
\centering
\caption{\textbf{MedBookVQA, Qwen3-VL-4B.} Common-suite accuracy (mean $\pm$ bootstrap std over evaluation items) for every baseline and for \namel{} at anchor budget $N=1280$ (Real/Content-Free). \textbf{Bold} marks the best method per column; \underline{underline} marks the second best. ``Target'' is final in-domain (target-task) accuracy.}
\label{tab:medbook-4b-full}
\resizebox{\textwidth}{!}{%
\begin{tabular}{lccccccccccc}
\toprule
 & \multicolumn{4}{c}{STEM} & \multicolumn{2}{c}{Gen. VQA} & \multicolumn{3}{c}{Chart/Doc.} & \multicolumn{1}{c}{Ground.} &  \\
\cmidrule(lr){2-5} \cmidrule(lr){6-7} \cmidrule(lr){8-10} \cmidrule(lr){11-11}
Method & MMMU & MVi & MVn & MVe & RWQA & MMS & DocVQA & CX-DQ & CX-RQ & CBQA & Target \\
\midrule
Base & \underline{62.8\,$\pm$\,1.2} & 61.9\,$\pm$\,1.5 & 39.6\,$\pm$\,0.9 & 48.6\,$\pm$\,0.8 & 68.5\,$\pm$\,1.7 & \underline{64.7\,$\pm$\,1.2} & 64.4\,$\pm$\,0.7 & 69.8\,$\pm$\,0.7 & 35.1\,$\pm$\,1.5 & 85.5\,$\pm$\,1.6 & 49.8 \\
GRPO (wd=0) & 54.4\,$\pm$\,2.4 & \underline{73.3\,$\pm$\,2.7} & 41.5\,$\pm$\,1.8 & \underline{54.6\,$\pm$\,1.6} & 60.3\,$\pm$\,3.5 & 61.6\,$\pm$\,2.5 & 76.6\,$\pm$\,1.1 & \underline{78.1\,$\pm$\,1.3} & \underline{39.2\,$\pm$\,3.0} & 73.1\,$\pm$\,3.9 & 99.2 \\
GRPO (wd=0.1) & 52.7\,$\pm$\,2.4 & 71.2\,$\pm$\,2.8 & 29.7\,$\pm$\,1.6 & 43.9\,$\pm$\,1.6 & \textbf{68.6\,$\pm$\,3.3} & \textbf{65.1\,$\pm$\,2.4} & \textbf{80.3\,$\pm$\,1.1} & 69.6\,$\pm$\,1.4 & 38.6\,$\pm$\,3.0 & \underline{86.4\,$\pm$\,3.0} & 95.8 \\
GRPO+KL & 55.4\,$\pm$\,1.2 & 70.6\,$\pm$\,1.4 & \underline{43.4\,$\pm$\,0.9} & 54.2\,$\pm$\,0.8 & 57.0\,$\pm$\,1.8 & 62.8\,$\pm$\,1.2 & 72.1\,$\pm$\,0.6 & 71.2\,$\pm$\,0.7 & 32.0\,$\pm$\,1.5 & 71.3\,$\pm$\,2.0 & 99.3 \\
GSPO & 40.2\,$\pm$\,2.3 & 69.4\,$\pm$\,2.9 & 36.8\,$\pm$\,1.7 & 50.2\,$\pm$\,1.6 & 54.1\,$\pm$\,3.5 & 61.0\,$\pm$\,2.5 & 65.3\,$\pm$\,1.2 & 54.4\,$\pm$\,1.5 & 32.1\,$\pm$\,2.9 & 79.2\,$\pm$\,3.6 & \underline{99.8} \\
GSPO+EWC & 55.2\,$\pm$\,2.3 & 72.4\,$\pm$\,2.8 & 33.7\,$\pm$\,1.9 & 53.6\,$\pm$\,1.6 & 54.9\,$\pm$\,3.5 & 61.1\,$\pm$\,2.5 & 3.8\,$\pm$\,0.5 & 70.7\,$\pm$\,1.4 & 35.0\,$\pm$\,3.0 & 69.7\,$\pm$\,4.1 & \textbf{100.0} \\
LoRA & \textbf{63.1\,$\pm$\,2.3} & 62.5\,$\pm$\,3.0 & 39.0\,$\pm$\,1.7 & 49.3\,$\pm$\,1.6 & \textbf{68.6\,$\pm$\,3.3} & 64.0\,$\pm$\,2.4 & 64.6\,$\pm$\,1.3 & 70.9\,$\pm$\,1.4 & 36.8\,$\pm$\,3.0 & \textbf{87.0\,$\pm$\,3.0} & 53.3$^{\dagger}$ \\
SFT & 32.5\,$\pm$\,2.2 & 50.5\,$\pm$\,3.1 & 11.2\,$\pm$\,1.1 & 21.8\,$\pm$\,1.3 & 43.4\,$\pm$\,3.5 & 38.6\,$\pm$\,2.5 & 60.4\,$\pm$\,1.3 & 65.4\,$\pm$\,1.5 & 31.3\,$\pm$\,2.9 & 73.3\,$\pm$\,3.9 & \textbf{100.0} \\
CPO & 55.9\,$\pm$\,2.3 & 71.7\,$\pm$\,2.8 & 36.6\,$\pm$\,1.7 & 51.8\,$\pm$\,1.6 & 68.2\,$\pm$\,3.3 & 63.6\,$\pm$\,2.4 & 77.2\,$\pm$\,1.1 & 73.1\,$\pm$\,1.4 & 36.3\,$\pm$\,3.0 & 85.3\,$\pm$\,3.1 & 72.2$^{\dagger}$ \\
\midrule
\realname{} & 54.4\,$\pm$\,2.4 & 72.8\,$\pm$\,2.8 & 40.9\,$\pm$\,1.8 & 52.5\,$\pm$\,1.6 & 59.2\,$\pm$\,3.5 & 62.3\,$\pm$\,2.5 & 72.9\,$\pm$\,1.2 & 66.9\,$\pm$\,1.5 & 35.6\,$\pm$\,3.0 & 83.1\,$\pm$\,3.3 & \underline{99.8} \\
\namel & 61.1\,$\pm$\,2.3 & \textbf{75.3\,$\pm$\,2.7} & \textbf{46.4\,$\pm$\,1.8} & \textbf{58.5\,$\pm$\,1.5} & 66.0\,$\pm$\,3.4 & 64.5\,$\pm$\,2.4 & \underline{79.2\,$\pm$\,1.1} & \textbf{81.8\,$\pm$\,1.2} & \textbf{40.2\,$\pm$\,3.0} & 86.2\,$\pm$\,3.1 & 99.7 \\
\bottomrule
\multicolumn{12}{l}{\footnotesize $^{\dagger}$ Below the per-domain in-domain-gain threshold; excluded from the headline forgetting comparison (Section~\ref{sec:exp-setup}).}
\end{tabular}%
}
\end{table}

\begin{table}[t]
\centering
\caption{\textbf{Puzzle, Qwen3-VL-4B.} Common-suite accuracy (mean $\pm$ bootstrap std over evaluation items) for every baseline and for \namel{} at anchor budget $N=1280$ (Real/Content-Free). \textbf{Bold} marks the best method per column; \underline{underline} marks the second best. ``Target'' is final in-domain (target-task) accuracy.}
\label{tab:puzzle-4b-full}
\resizebox{\textwidth}{!}{%
\begin{tabular}{lccccccccccc}
\toprule
 & \multicolumn{4}{c}{STEM} & \multicolumn{2}{c}{Gen. VQA} & \multicolumn{3}{c}{Chart/Doc.} & \multicolumn{1}{c}{Ground.} &  \\
\cmidrule(lr){2-5} \cmidrule(lr){6-7} \cmidrule(lr){8-10} \cmidrule(lr){11-11}
Method & MMMU & MVi & MVn & MVe & RWQA & MMS & DocVQA & CX-DQ & CX-RQ & CBQA & Target \\
\midrule
Base & 62.8\,$\pm$\,1.2 & 61.9\,$\pm$\,1.5 & 39.6\,$\pm$\,0.9 & 48.6\,$\pm$\,0.8 & \textbf{68.5\,$\pm$\,1.7} & 64.7\,$\pm$\,1.2 & 64.4\,$\pm$\,0.6 & 69.8\,$\pm$\,0.7 & 35.1\,$\pm$\,1.5 & 85.5\,$\pm$\,1.6 & 27.5 \\
GRPO (wd=0) & 53.7\,$\pm$\,2.3 & 67.3\,$\pm$\,2.9 & 24.9\,$\pm$\,1.6 & 49.1\,$\pm$\,1.6 & 34.1\,$\pm$\,3.5 & 62.0\,$\pm$\,2.5 & 74.4\,$\pm$\,1.2 & 62.7\,$\pm$\,1.5 & 35.2\,$\pm$\,2.9 & 56.8\,$\pm$\,4.4 & 68.2$^{\dagger}$ \\
GRPO (wd=0.1) & 53.0\,$\pm$\,2.4 & 70.9\,$\pm$\,2.8 & 29.4\,$\pm$\,1.6 & 46.4\,$\pm$\,1.6 & 53.9\,$\pm$\,3.5 & 61.3\,$\pm$\,2.5 & 67.7\,$\pm$\,1.2 & 64.4\,$\pm$\,1.5 & 40.4\,$\pm$\,3.0 & \underline{85.1\,$\pm$\,3.2} & 88.2 \\
GRPO+KL & 60.9\,$\pm$\,1.2 & \underline{73.7\,$\pm$\,1.4} & \textbf{46.8\,$\pm$\,0.9} & \textbf{56.6\,$\pm$\,0.8} & 63.3\,$\pm$\,1.7 & \underline{66.9\,$\pm$\,1.2} & \underline{76.9\,$\pm$\,0.6} & 69.5\,$\pm$\,0.7 & \underline{41.3\,$\pm$\,1.6} & 78.8\,$\pm$\,1.8 & {88.8} \\
GSPO & 52.1\,$\pm$\,2.4 & 66.7\,$\pm$\,2.9 & 24.2\,$\pm$\,1.5 & 46.2\,$\pm$\,1.5 & 58.7\,$\pm$\,3.5 & 58.3\,$\pm$\,2.5 & 59.5\,$\pm$\,1.3 & 51.1\,$\pm$\,1.6 & 30.2\,$\pm$\,2.8 & 64.0\,$\pm$\,4.0 & \underline{90.8} \\
GSPO+EWC & 21.0\,$\pm$\,1.9 & 58.2\,$\pm$\,3.1 & 21.4\,$\pm$\,1.4 & 38.4\,$\pm$\,1.6 & 61.1\,$\pm$\,3.7 & 51.9\,$\pm$\,2.5 & 65.6\,$\pm$\,1.3 & 54.2\,$\pm$\,1.5 & 28.9\,$\pm$\,2.8 & 61.9\,$\pm$\,4.3 & \textbf{91.0} \\
LoRA & \textbf{63.4\,$\pm$\,2.3} & 61.8\,$\pm$\,3.0 & 39.8\,$\pm$\,1.5 & 49.3\,$\pm$\,1.5 & \underline{67.1\,$\pm$\,3.3} & 65.2\,$\pm$\,2.4 & 65.7\,$\pm$\,1.2 & 70.3\,$\pm$\,1.4 & 35.7\,$\pm$\,2.9 & \textbf{86.0\,$\pm$\,3.0} & 26.2$^{\dagger}$ \\
SFT & 50.8\,$\pm$\,2.4 & 66.9\,$\pm$\,2.9 & 19.7\,$\pm$\,1.4 & 54.5\,$\pm$\,1.6 & 54.1\,$\pm$\,3.5 & 44.2\,$\pm$\,2.5 & 50.2\,$\pm$\,1.4 & \textbf{76.1\,$\pm$\,1.4} & 33.7\,$\pm$\,3.0 & 72.3\,$\pm$\,4.0 & 19.5$^{\dagger}$ \\
CPO & 32.1\,$\pm$\,2.2 & 58.4\,$\pm$\,3.0 & 19.6\,$\pm$\,1.4 & 37.9\,$\pm$\,1.5 & 63.1\,$\pm$\,3.4 & 61.0\,$\pm$\,2.5 & 60.7\,$\pm$\,1.3 & 59.7\,$\pm$\,1.5 & 31.1\,$\pm$\,2.9 & 48.8\,$\pm$\,4.4 & 69.2$^{\dagger}$ \\
\midrule
\realname{} & 60.4\,$\pm$\,2.3 & \textbf{75.5\,$\pm$\,2.6} & 42.3\,$\pm$\,1.7 & \underline{55.6\,$\pm$\,1.6} & 60.6\,$\pm$\,3.3 & \textbf{67.7\,$\pm$\,2.4} & \textbf{77.6\,$\pm$\,1.1} & \underline{75.0\,$\pm$\,1.4} & \textbf{41.4\,$\pm$\,3.0} & 78.0\,$\pm$\,3.7 & \textbf{91.0} \\
\namel & \underline{64.2\,$\pm$\,2.3} & 72.5\,$\pm$\,2.8 & \underline{46.3\,$\pm$\,1.7} & 52.6\,$\pm$\,1.6 & 64.7\,$\pm$\,3.6 & 65.1\,$\pm$\,2.3 & 75.0\,$\pm$\,1.1 & 74.1\,$\pm$\,1.4 & 40.1\,$\pm$\,3.1 & \underline{84.9\,$\pm$\,3.3} & 90.0 \\
\bottomrule
\multicolumn{12}{l}{\footnotesize $^{\dagger}$ Below the per-domain in-domain-gain threshold; excluded from the headline forgetting comparison (Section~\ref{sec:exp-setup}).}
\end{tabular}%
}
\end{table}

\begin{table}[t]
\centering
\caption{\textbf{Navigation, Qwen3-VL-8B.} Common-suite accuracy (mean $\pm$ bootstrap std over evaluation items) for every baseline and for \namel{} at anchor budget $N=1280$ (Real/Content-Free). \textbf{Bold} marks the best method per column; \underline{underline} marks the second best. ``Target'' is final in-domain (target-task) accuracy.}
\label{tab:nav-8b-full}
\resizebox{\textwidth}{!}{%
\begin{tabular}{lccccccccccc}
\toprule
 & \multicolumn{4}{c}{STEM} & \multicolumn{2}{c}{Gen. VQA} & \multicolumn{3}{c}{Chart/Doc.} & \multicolumn{1}{c}{Ground.} &  \\
\cmidrule(lr){2-5} \cmidrule(lr){6-7} \cmidrule(lr){8-10} \cmidrule(lr){11-11}
Method & MMMU & MVi & MVn & MVe & RWQA & MMS & DocVQA & CX-DQ & CX-RQ & CBQA & Target \\
\midrule
Base & \textbf{68.0\,$\pm$\,1.1} & 53.4\,$\pm$\,1.6 & 43.5\,$\pm$\,0.9 & 48.8\,$\pm$\,0.8 & \underline{67.7\,$\pm$\,1.7} & 57.4\,$\pm$\,1.3 & 67.8\,$\pm$\,0.6 & 79.2\,$\pm$\,0.6 & 37.6\,$\pm$\,1.5 & 85.7\,$\pm$\,1.6 & 28.7 \\
GRPO (wd=0) & 65.6\,$\pm$\,1.1 & 55.6\,$\pm$\,1.6 & 44.8\,$\pm$\,0.9 & 47.2\,$\pm$\,0.8 & 64.7\,$\pm$\,1.7 & 58.4\,$\pm$\,1.3 & 55.2\,$\pm$\,0.7 & 78.5\,$\pm$\,0.6 & 32.1\,$\pm$\,1.5 & 85.3\,$\pm$\,1.6 & 87.9 \\
GRPO (wd=0.1) & 66.0\,$\pm$\,1.1 & 58.7\,$\pm$\,1.6 & 46.4\,$\pm$\,0.9 & 64.7\,$\pm$\,1.2 & 64.7\,$\pm$\,1.7 & 64.7\,$\pm$\,1.2 & 57.6\,$\pm$\,0.7 & 79.2\,$\pm$\,0.6 & 34.1\,$\pm$\,1.5 & 86.0\,$\pm$\,3.0 & 89.7 \\
GRPO+KL & 64.7\,$\pm$\,1.2 & \underline{74.3\,$\pm$\,1.4} & \textbf{48.6\,$\pm$\,0.9} & 56.6\,$\pm$\,0.8 & 63.9\,$\pm$\,1.7 & \textbf{70.5\,$\pm$\,1.2} & 67.8\,$\pm$\,0.6 & 78.0\,$\pm$\,0.7 & 38.9\,$\pm$\,1.5 & 73.5\,$\pm$\,2.0 & 93.4 \\
GSPO & 62.5\,$\pm$\,1.2 & 73.5\,$\pm$\,1.4 & 39.5\,$\pm$\,0.9 & \textbf{66.5\,$\pm$\,1.2} & 52.4\,$\pm$\,1.8 & 66.5\,$\pm$\,1.2 & 62.9\,$\pm$\,0.7 & 71.4\,$\pm$\,0.7 & 37.3\,$\pm$\,1.5 & 83.3\,$\pm$\,3.2 & 96.0 \\
GSPO+EWC & 60.9\,$\pm$\,1.2 & 69.2\,$\pm$\,1.5 & 39.1\,$\pm$\,0.9 & 52.9\,$\pm$\,0.8 & 56.1\,$\pm$\,1.8 & 65.5\,$\pm$\,1.2 & 64.8\,$\pm$\,0.6 & 75.0\,$\pm$\,0.7 & 37.0\,$\pm$\,1.5 & 79.4\,$\pm$\,1.8 & 93.6 \\
LoRA & 66.8\,$\pm$\,2.2 & 53.9\,$\pm$\,3.1 & 43.3\,$\pm$\,1.8 & 48.6\,$\pm$\,1.6 & 66.9\,$\pm$\,3.3 & 60.0\,$\pm$\,2.5 & 69.8\,$\pm$\,1.2 & 80.6\,$\pm$\,1.2 & 37.8\,$\pm$\,3.0 & 84.5\,$\pm$\,3.2 & 27.2$^{\dagger}$ \\
SFT & 62.2\,$\pm$\,2.3 & 62.5\,$\pm$\,3.0 & 18.8\,$\pm$\,1.4 & 17.7\,$\pm$\,1.2 & 66.9\,$\pm$\,3.3 & 62.5\,$\pm$\,2.5 & \textbf{83.6\,$\pm$\,1.0} & 70.9\,$\pm$\,1.4 & 40.1\,$\pm$\,3.0 & 76.2\,$\pm$\,3.8 & 58.1$^{\dagger}$ \\
CPO & 66.4\,$\pm$\,2.4 & 62.2\,$\pm$\,3.0 & 33.0\,$\pm$\,1.7 & 47.9\,$\pm$\,1.6 & 65.0\,$\pm$\,3.4 & 66.4\,$\pm$\,2.4 & 66.2\,$\pm$\,1.3 & 78.3\,$\pm$\,1.3 & 39.1\,$\pm$\,3.0 & 84.9\,$\pm$\,3.2 & 84.9 \\
\midrule
\realname{} & 66.2\,$\pm$\,1.1 & 70.3\,$\pm$\,1.4 & \underline{47.9\,$\pm$\,0.9} & \underline{57.6\,$\pm$\,0.8} & 59.5\,$\pm$\,1.8 & 68.1\,$\pm$\,1.2 & 67.1\,$\pm$\,0.6 & \textbf{80.9\,$\pm$\,0.6} & \textbf{41.3\,$\pm$\,1.6} & \underline{85.3\,$\pm$\,1.6} & \textbf{98.4} \\
\namel & \underline{65.9\,$\pm$\,1.1} & \textbf{76.0\,$\pm$\,1.4} & 47.8\,$\pm$\,0.9 & 56.9\,$\pm$\,0.8 & \textbf{68.7\,$\pm$\,1.8} & \underline{69.1\,$\pm$\,1.2} & \underline{66.8\,$\pm$\,0.6} & \underline{79.8\,$\pm$\,0.7} & \underline{41.0\,$\pm$\,1.6} & \textbf{86.2\,$\pm$\,1.6} & \underline{97.1} \\
\bottomrule
\multicolumn{12}{l}{\footnotesize $^{\dagger}$ Below the per-domain in-domain-gain threshold; excluded from the headline forgetting comparison (Section~\ref{sec:exp-setup}).}
\end{tabular}%
}
\end{table}

\begin{table}[t]
\centering
\caption{\textbf{MedBookVQA, Qwen3-VL-8B.} Common-suite accuracy (mean $\pm$ bootstrap std over evaluation items) for every baseline and for \namel{} at anchor budget $N=1280$ (Real/Content-Free). \textbf{Bold} marks the best method per column; \underline{underline} marks the second best. ``Target'' is final in-domain (target-task) accuracy.}
\label{tab:medbook-8b-full}
\resizebox{\textwidth}{!}{%
\begin{tabular}{lccccccccccc}
\toprule
 & \multicolumn{4}{c}{STEM} & \multicolumn{2}{c}{Gen. VQA} & \multicolumn{3}{c}{Chart/Doc.} & \multicolumn{1}{c}{Ground.} &  \\
\cmidrule(lr){2-5} \cmidrule(lr){6-7} \cmidrule(lr){8-10} \cmidrule(lr){11-11}
Method & MMMU & MVi & MVn & MVe & RWQA & MMS & DocVQA & CX-DQ & CX-RQ & CBQA & Target \\
\midrule
Base & \textbf{68.0\,$\pm$\,1.1} & 53.4\,$\pm$\,1.6 & 43.5\,$\pm$\,0.9 & 48.8\,$\pm$\,0.8 & \textbf{67.7\,$\pm$\,1.7} & 57.4\,$\pm$\,1.3 & 67.8\,$\pm$\,0.6 & 79.2\,$\pm$\,0.6 & 37.6\,$\pm$\,1.5 & 85.7\,$\pm$\,1.6 & 44.1 \\
GRPO (wd=0) & 61.2\,$\pm$\,1.2 & 72.8\,$\pm$\,1.4 & 35.2\,$\pm$\,0.9 & 54.6\,$\pm$\,0.8 & 63.0\,$\pm$\,1.8 & 62.6\,$\pm$\,1.2 & 53.0\,$\pm$\,0.7 & 78.8\,$\pm$\,0.6 & 38.3\,$\pm$\,1.5 & 76.2\,$\pm$\,1.9 & 98.8 \\
GRPO (wd=0.1) & 62.5\,$\pm$\,1.2 & 76.2\,$\pm$\,1.4 & 51.6\,$\pm$\,0.9 & 58.6\,$\pm$\,0.8 & 63.7\,$\pm$\,1.7 & 66.7\,$\pm$\,1.2 & \underline{80.0\,$\pm$\,0.6} & 84.4\,$\pm$\,0.6 & 40.6\,$\pm$\,1.6 & 81.5\,$\pm$\,1.8 & 99.5 \\
GRPO+KL & 63.6\,$\pm$\,1.2 & 74.1\,$\pm$\,1.4 & 49.4\,$\pm$\,0.9 & 55.7\,$\pm$\,0.8 & 64.4\,$\pm$\,1.7 & 66.1\,$\pm$\,1.2 & 69.7\,$\pm$\,0.6 & 73.2\,$\pm$\,0.7 & 38.5\,$\pm$\,1.5 & 81.9\,$\pm$\,1.7 & 99.7 \\
GSPO & 58.4\,$\pm$\,2.3 & 76.1\,$\pm$\,2.6 & 29.1\,$\pm$\,1.6 & 53.9\,$\pm$\,2.3 & 56.0\,$\pm$\,2.6 & 65.6\,$\pm$\,2.4 & 31.7\,$\pm$\,1.2 & 69.9\,$\pm$\,1.4 & 40.9\,$\pm$\,3.1 & 82.5\,$\pm$\,3.4 & 99.7 \\
GSPO+EWC & 61.9\,$\pm$\,1.2 & 74.4\,$\pm$\,1.4 & 39.3\,$\pm$\,0.9 & 55.8\,$\pm$\,0.8 & 60.4\,$\pm$\,1.8 & \underline{68.3\,$\pm$\,1.2} & 65.5\,$\pm$\,0.6 & 85.1\,$\pm$\,0.6 & 44.1\,$\pm$\,1.6 & 85.3\,$\pm$\,1.6 & 99.0 \\
LoRA & 66.9\,$\pm$\,2.2 & 64.3\,$\pm$\,2.9 & 43.1\,$\pm$\,1.8 & 50.6\,$\pm$\,1.6 & 66.5\,$\pm$\,3.3 & 66.4\,$\pm$\,2.4 & \textbf{81.7\,$\pm$\,1.0} & 79.7\,$\pm$\,1.2 & 39.5\,$\pm$\,3.0 & \underline{86.0\,$\pm$\,3.1} & 88.0$^{\dagger}$ \\
SFT & 25.7\,$\pm$\,2.1 & 78.3\,$\pm$\,1.7 & 10.8\,$\pm$\,1.1 & 9.9\,$\pm$\,0.9 & 16.9\,$\pm$\,2.7 & 22.1\,$\pm$\,2.1 & 0.0\,$\pm$\,0.0 & 0.0\,$\pm$\,0.0 & 0.2\,$\pm$\,0.3 & 0.0\,$\pm$\,0.0 & \underline{99.8} \\
CPO & 60.3\,$\pm$\,2.3 & 71.3\,$\pm$\,3.0 & 25.8\,$\pm$\,1.5 & 53.5\,$\pm$\,1.6 & 62.9\,$\pm$\,3.5 & 63.9\,$\pm$\,2.4 & 23.9\,$\pm$\,1.1 & 81.6\,$\pm$\,1.2 & 39.2\,$\pm$\,3.1 & 81.7\,$\pm$\,3.5 & 99.7 \\
\midrule
\realname{} & 65.6\,$\pm$\,2.2 & \textbf{79.1\,$\pm$\,2.5} & \textbf{52.6\,$\pm$\,1.8} & \textbf{60.6\,$\pm$\,1.5} & 66.4\,$\pm$\,3.4 & 67.1\,$\pm$\,2.4 & 79.7\,$\pm$\,1.1 & \textbf{85.5\,$\pm$\,1.1} & \underline{44.5\,$\pm$\,3.1} & 83.5\,$\pm$\,3.3 & 98.6 \\
\namel & \underline{67.4\,$\pm$\,2.2} & \underline{78.3\,$\pm$\,2.6} & \underline{51.7\,$\pm$\,1.8} & \underline{60.1\,$\pm$\,1.5} & \underline{66.5\,$\pm$\,3.5} & \textbf{71.3\,$\pm$\,2.3} & 77.4\,$\pm$\,1.1 & \underline{85.2\,$\pm$\,1.1} & \textbf{45.6\,$\pm$\,3.1} & \textbf{87.4\,$\pm$\,1.5} & \textbf{100.0} \\
\bottomrule
\multicolumn{12}{l}{\footnotesize $^{\dagger}$ Below the per-domain in-domain-gain threshold; excluded from the headline forgetting comparison (Section~\ref{sec:exp-setup}).}
\end{tabular}%
}
\end{table}

\begin{table}[t]
\centering
\caption{\textbf{Puzzle, Qwen3-VL-8B.} Common-suite accuracy (mean $\pm$ bootstrap std over evaluation items) for every baseline and for \namel{} at anchor budget $N=1280$ (Real/Content-Free). \textbf{Bold} marks the best method per column; \underline{underline} marks the second best. ``Target'' is final in-domain (target-task) accuracy.}
\label{tab:puzzle-8b-full}
\resizebox{\textwidth}{!}{%
\begin{tabular}{lccccccccccc}
\toprule
 & \multicolumn{4}{c}{STEM} & \multicolumn{2}{c}{Gen. VQA} & \multicolumn{3}{c}{Chart/Doc.} & \multicolumn{1}{c}{Ground.} &  \\
\cmidrule(lr){2-5} \cmidrule(lr){6-7} \cmidrule(lr){8-10} \cmidrule(lr){11-11}
Method & MMMU & MVi & MVn & MVe & RWQA & MMS & DocVQA & CX-DQ & CX-RQ & CBQA & Target \\
\midrule
Base & \textbf{68.0\,$\pm$\,1.1} & 53.4\,$\pm$\,1.6 & 43.5\,$\pm$\,0.9 & 48.8\,$\pm$\,0.8 & \textbf{67.7\,$\pm$\,1.7} & 57.4\,$\pm$\,1.3 & 67.8\,$\pm$\,0.6 & 79.2\,$\pm$\,0.6 & 37.6\,$\pm$\,1.5 & \textbf{85.7\,$\pm$\,1.6} & 28.8 \\
GRPO (wd=0) & 60.6\,$\pm$\,2.3 & 73.8\,$\pm$\,2.7 & 39.6\,$\pm$\,1.7 & 54.7\,$\pm$\,1.6 & 48.6\,$\pm$\,3.5 & 65.9\,$\pm$\,2.4 & 75.1\,$\pm$\,1.2 & 81.3\,$\pm$\,1.2 & 40.8\,$\pm$\,3.1 & 80.5\,$\pm$\,3.5 & 87.0 \\
GRPO (wd=0.1) & 62.5\,$\pm$\,1.2 & 70.2\,$\pm$\,1.5 & 43.7\,$\pm$\,0.9 & 53.5\,$\pm$\,0.8 & 63.5\,$\pm$\,1.7 & 66.2\,$\pm$\,1.2 & 74.9\,$\pm$\,0.6 & 78.7\,$\pm$\,0.7 & 38.3\,$\pm$\,1.5 & 61.3\,$\pm$\,2.2 & 90.8 \\
GRPO+KL & 62.7\,$\pm$\,1.2 & 71.7\,$\pm$\,1.4 & 40.9\,$\pm$\,0.9 & 54.8\,$\pm$\,0.8 & 61.2\,$\pm$\,1.8 & 67.4\,$\pm$\,1.2 & 73.1\,$\pm$\,0.6 & 80.3\,$\pm$\,0.6 & 41.2\,$\pm$\,1.6 & 79.0\,$\pm$\,1.8 & 93.3 \\
GSPO & 62.3\,$\pm$\,2.3 & 72.8\,$\pm$\,2.8 & 37.0\,$\pm$\,1.7 & 55.4\,$\pm$\,1.6 & 56.7\,$\pm$\,3.5 & 64.3\,$\pm$\,2.4 & 41.4\,$\pm$\,1.3 & 76.6\,$\pm$\,1.3 & 40.8\,$\pm$\,3.1 & 76.2\,$\pm$\,3.8 & \textbf{95.0} \\
GSPO+EWC & 57.0\,$\pm$\,1.2 & 66.3\,$\pm$\,1.5 & 29.8\,$\pm$\,0.8 & 49.1\,$\pm$\,0.8 & 51.5\,$\pm$\,1.8 & 62.9\,$\pm$\,1.2 & 53.8\,$\pm$\,0.7 & 79.0\,$\pm$\,0.6 & 36.4\,$\pm$\,1.5 & 61.1\,$\pm$\,2.2 & \underline{94.5} \\
LoRA & \underline{65.9\,$\pm$\,2.2} & 59.8\,$\pm$\,3.0 & 43.1\,$\pm$\,1.8 & 52.4\,$\pm$\,1.6 & 63.4\,$\pm$\,2.2 & 59.6\,$\pm$\,2.5 & 72.3\,$\pm$\,1.2 & 73.6\,$\pm$\,1.4 & 32.9\,$\pm$\,2.9 & 53.0\,$\pm$\,4.4 & 90.5 \\
SFT & 25.3\,$\pm$\,2.1 & 15.9\,$\pm$\,2.3 & 10.5\,$\pm$\,1.1 & 10.7\,$\pm$\,1.0 & 19.4\,$\pm$\,2.8 & 22.5\,$\pm$\,2.1 & 0.1\,$\pm$\,0.1 & 0.0\,$\pm$\,0.1 & 10.2\,$\pm$\,0.3 & 0.0\,$\pm$\,0.0 & 35.0$^{\dagger}$ \\
CPO & 64.5\,$\pm$\,2.3 & 73.0\,$\pm$\,2.8 & 43.5\,$\pm$\,1.8 & \underline{57.1\,$\pm$\,1.6} & 64.2\,$\pm$\,3.4 & \textbf{68.3\,$\pm$\,2.3} & 57.7\,$\pm$\,1.3 & 82.4\,$\pm$\,1.2 & \underline{42.6\,$\pm$\,3.1} & 47.5\,$\pm$\,4.5 & 89.3 \\
\midrule
\realname{} & 64.9\,$\pm$\,1.2 & \textbf{76.4\,$\pm$\,1.3} & \underline{44.7\,$\pm$\,0.9} & 56.9\,$\pm$\,0.8 & 65.2\,$\pm$\,1.7 & \underline{67.6\,$\pm$\,1.2} & \underline{77.9\,$\pm$\,0.6} & \textbf{83.0\,$\pm$\,0.6} & 40.8\,$\pm$\,1.6 & 79.6\,$\pm$\,1.8 & 90.3 \\
\namel & 65.0\,$\pm$\,1.2 & \underline{75.5\,$\pm$\,1.4} & \textbf{45.3\,$\pm$\,0.9} & \textbf{58.1\,$\pm$\,0.8} & \underline{67.3\,$\pm$\,1.7} & 67.1\,$\pm$\,1.2 & \textbf{81.8\,$\pm$\,0.5} & \underline{82.9\,$\pm$\,0.6} & \textbf{45.0\,$\pm$\,1.6} & \underline{84.9\,$\pm$\,1.6} & 93.0 \\
\bottomrule
\multicolumn{12}{l}{\footnotesize $^{\dagger}$ Below the per-domain in-domain-gain threshold; excluded from the headline forgetting comparison (Section~\ref{sec:exp-setup}).}
\end{tabular}%
}
\end{table}

\paragraph{\namel{} beyond RL.}
The anchor term depends only on the model's outputs on anchor inputs relative to the frozen reference, not on the training objective, so it can be added unchanged to any fine-tuning loss. Table~\ref{tab:med-2b-sft-lora} applies \namel{} to supervised fine-tuning and LoRA on MedBookVQA. SFT is the most destructive method in our study (worst-1 change $-78.8\%$), and adding \namel{} reduces this to $-9.1\%$ while retaining most of the target gain (92.5 vs.\ 95.1). LoRA learns little of the task in this setting and so forgets little; adding \namel{} leaves target accuracy unchanged (51.3 vs.\ 50.8) and raises the average change across the suite from $+2.7\%$ to
$+6.5\%$. \namel{} is therefore a general post-training regulariser, not a component specific to GRPO or GSPO.

\begin{table}[hbt]
\centering
\caption{\textbf{\namel{} with non-RL post-training (MedBookVQA, Qwen3-VL-2B).}
Common-suite accuracy (mean $\pm$ bootstrap std over evaluation items) for supervised fine-tuning (SFT) and LoRA,
each with and without \namel{} (content-free anchors, $N=1280$, same $\lambda$ as
the RL runs). \textbf{Bold} marks the better of each baseline/\namel{} pair.
``Target'' is final in-domain accuracy.}
\label{tab:med-2b-sft-lora}
\resizebox{\textwidth}{!}{%
\begin{tabular}{lccccccccccc}
\toprule
 & \multicolumn{4}{c}{STEM} & \multicolumn{2}{c}{Gen. VQA} & \multicolumn{3}{c}{Chart/Doc.} & \multicolumn{1}{c}{Ground.} &  \\
\cmidrule(lr){2-5} \cmidrule(lr){6-7} \cmidrule(lr){8-10} \cmidrule(lr){11-11}
Method & MMMU & MVi & MVn & MVe & RWQA & MMS & DocVQA & CX-DQ & CX-RQ & CBQA & Target \\
\midrule
Base & 48.4\,$\pm$\,2.4 & 56.0\,$\pm$\,3.1 & 21.1\,$\pm$\,1.4 & 37.4\,$\pm$\,1.5 & 58.4\,$\pm$\,3.5 & 46.7\,$\pm$\,2.5 & 68.6\,$\pm$\,1.2 & 60.8\,$\pm$\,1.5 & 28.0\,$\pm$\,2.8 & 82.5\,$\pm$\,3.4 & 49.4 \\
\midrule
SFT & 37.9\,$\pm$\,2.3 & 36.6\,$\pm$\,3.0 & 9.5\,$\pm$\,1.0 & 16.2\,$\pm$\,1.1 & 41.4\,$\pm$\,3.5 & 34.7\,$\pm$\,2.4 & 26.2\,$\pm$\,1.2 & 42.8\,$\pm$\,1.5 & 19.4\,$\pm$\,2.5 & 17.5\,$\pm$\,3.4 & \textbf{95.1} \\
SFT + \namel & \textbf{48.7\,$\pm$\,2.3} & \textbf{56.2\,$\pm$\,3.1} & \textbf{23.3\,$\pm$\,1.6} & \textbf{40.4\,$\pm$\,1.6} & \textbf{58.5\,$\pm$\,3.5} & \textbf{58.6\,$\pm$\,2.5} & \textbf{65.9\,$\pm$\,1.3} & \textbf{55.2\,$\pm$\,1.5} & \textbf{26.8\,$\pm$\,2.8} & \textbf{66.9\,$\pm$\,3.3} & 92.5 \\
\midrule
LoRA & 49.1\,$\pm$\,2.4 & 58.3\,$\pm$\,3.1 & 23.8\,$\pm$\,1.5 & 38.9\,$\pm$\,1.5 & \textbf{61.6\,$\pm$\,3.5} & 48.7\,$\pm$\,2.5 & \textbf{66.0\,$\pm$\,1.3} & \textbf{62.8\,$\pm$\,1.5} & 27.8\,$\pm$\,2.8 & 79.4\,$\pm$\,3.6 & 50.8$^{\dagger}$ \\
LoRA + \namel & \textbf{49.7\,$\pm$\,2.2} & \textbf{60.6\,$\pm$\,3.0} & \textbf{26.4\,$\pm$\,1.5} & \textbf{42.3\,$\pm$\,1.6} & 58.5\,$\pm$\,3.5 & \textbf{57.5\,$\pm$\,2.5} & 65.0\,$\pm$\,1.2 & 60.1\,$\pm$\,1.5 & \textbf{28.0\,$\pm$\,2.5} & \textbf{81.7\,$\pm$\,3.2} & \textbf{51.3}$^{\dagger}$ \\
\bottomrule
\multicolumn{12}{l}{\footnotesize $^{\dagger}$ Below the per-domain in-domain-gain threshold (Section~\ref{sec:exp-setup}); LoRA learns little of the target task in this setting.}
\end{tabular}%
}
\end{table}

\subsection{Understanding Domain and Scale Dynamics: Pre-Existing Competence and Task Memorizability}
\label{app:domain-scale-mechanisms}

A prominent empirical pattern emerges from our $3 \times 3$ grid of domain and scale evaluations (Tables~\ref{tab:nav-2b-full}-\ref{tab:puzzle-8b-full}): {Qwen3-VL-4B on Navigation learns the target task to high accuracy ($93.4\%$) with virtually zero tail forgetting under standard RL} (worst-1 drop of only $-1.2\%$ on GRPO $\lambda_{\mathrm{wd}}=0.1$, $-2.2\%$ on GSPO, and $-0.1\%$ on \namel{}). In sharp contrast:
\begin{enumerate}
    \item At the identical 4B model scale, both MedBookVQA and Puzzle suffer severe catastrophic forgetting under unregularized RL (worst-1 drops of $-25.0\%$ to $-38.9\%$, Tables~\ref{tab:medbook-4b-full} and \ref{tab:puzzle-4b-full}).
    \item On Navigation itself, both the smaller 2B model (worst-1 drop of $-40.5\%$ on CountBenchQA, Table~\ref{tab:nav-2b-full}) and the larger 8B model (worst-1 drops of $-15.0\%$ to $-22.6\%$ on DocVQA and RealWorldQA, Table~\ref{tab:nav-8b-full}) degrade significantly under unregularized RL.
\end{enumerate}
This disparity demonstrates that model parameter capacity alone cannot predict catastrophic forgetting. Below, we synthesize the diagnostic evidence across all nine cells into a predictive two-factor framework: (i) \emph{chance-corrected pre-existing competence} and (ii) \emph{task memorizability versus procedural planning}. Table~\ref{tab:competence-memorizability} presents the cross-cell diagnostic matrix.

\begin{table*}[t]
\centering
\caption{\textbf{Diagnostic Framework for Domain and Scale Dynamics.} Comparison across all nine experimental cells (3 domains $\times$ 3 scales). ``Chance'' denotes the random-guess baseline determined by the answer space (prompts detailed in Section~\ref{sec:exp-setup} and Appendix~\ref{app:qualitative}). ``Above Chance'' is Base Target Accuracy minus Chance. ``SFT Target'' serves as an empirical probe of task memorizability. Unregularized RL performance is illustrated by canonical GRPO ($\lambda_{\mathrm{wd}}=0.1$). \textbf{Bold} highlights the Navigation 4B anomaly and top retention scores.}
\label{tab:competence-memorizability}
\resizebox{\textwidth}{!}{%
\begin{tabular}{llrrrrrrrr}
\toprule
 & & \multicolumn{3}{c}{\textbf{Pre-Existing Base Competence}} & \multicolumn{1}{c}{\textbf{Memorizability}} & \multicolumn{2}{c}{\textbf{Unregularized RL (GRPO)}} & \multicolumn{2}{c}{\textbf{\namel{} (Ours)}} \\
\cmidrule(lr){3-5} \cmidrule(lr){6-6} \cmidrule(lr){7-8} \cmidrule(lr){9-10}
\textbf{Target Domain} & \textbf{Scale} & \textbf{Base Target} & \textbf{Chance Level} & \textbf{Above Chance} & \textbf{SFT Target Acc} & \textbf{Target Acc} & \textbf{Worst-1 Shift} & \textbf{Target Acc} & \textbf{Worst-1 Shift} \\
\midrule
\multirow{3}{*}{Navigation} 
 & 2B & 3.5\% & 0.0\% & +3.5\% & 3.9\% & 70.8\% & -40.5\% & 87.0\% & -3.5\% \\
 & \textbf{4B} & \textbf{27.6\%} & \textbf{0.0\%} & \textbf{+27.6\%} & \textbf{23.9\%} & \textbf{93.4\%} & \textbf{-1.2\%} & \textbf{94.3\%} & \textbf{-0.1\%} \\
 & 8B & 28.7\% & 0.0\% & +28.7\% & 58.1\% & 89.7\% & -15.0\% & 97.1\% & -3.1\% \\
\midrule
\multirow{3}{*}{MedBookVQA}
 & 2B & 49.4\% & 22.5\% & +26.9\% & 95.1\% & 94.7\% & -31.4\% & 98.5\% & -0.3\% \\
 & 4B & 49.8\% & 22.5\% & +27.3\% & 100.0\% & 95.8\% & -25.0\% & 99.7\% & -3.6\% \\
 & 8B & 44.1\% & 22.5\% & +21.6\% & 99.8\% & 99.5\% & -8.1\% & 100.0\% & -1.8\% \\
\midrule
\multirow{3}{*}{Puzzle}
 & 2B & 23.8\% & 22.5\% & +1.3\% & 33.0\% & 82.2\% & -24.8\% & 90.8\% & -9.1\% \\
 & 4B & 27.5\% & 22.5\% & +5.0\% & 19.5\% & 88.2\% & -25.8\% & 90.0\% & -5.5\% \\
 & 8B & 28.8\% & 22.5\% & +6.3\% & 35.0\% & 90.8\% & -28.5\% & 93.0\% & -4.4\% \\
\bottomrule
\end{tabular}%
}
\end{table*}

\paragraph{Factor 1: Chance-Corrected Base Competence versus Raw Accuracy.}
A naive inspection of raw zero-shot target accuracy would suggest that at 4B, the base model performs almost identically on Navigation ($27.6\%$) and Puzzle ($27.5\%$). However, evaluating accuracy relative to task chance level reveals that these numbers represent diametrically opposite internal states:
\begin{itemize}[leftmargin=*]
    \item \textbf{Navigation Answer Space (Combinatorial Path Space):} The Navigation prompt requires generating a comma-separated sequence of coordinate actions (e.g., \texttt{U, R, D, L, ...}) describing a multi-step path on an input grid. The probability of randomly guessing a valid multi-step sequence is effectively zero ($0.0\% \ll 10^{-4}$). The 4B base model's $27.6\%$ zero-shot accuracy therefore represents \emph{authentic, pre-existing spatial-procedural competence} acquired during multimodal pre-training.
    \item \textbf{Puzzle Answer Space (Categorical Choice Space):} In contrast, Puzzle prompts demand a single multiple-choice letter ($A$-$E$). The random guessing baseline is $1/4$ to $1/5$ (mean $22.5\%$). The base model's score of $27.5\%$ is merely $+5.0\%$ above random chance. In other words, the base model possesses virtually no latent capability to solve visual puzzles.
\end{itemize}
Across all nine cells, above-chance base competence exhibits a strong positive correlation with tail retention under unregularized RL ($r = +0.395$). When a model lacks pre-existing competence (as in Puzzle at all scales, and Navigation at 2B where base accuracy is only $3.5\%$), reinforcement learning cannot simply refine existing capabilities; it is forced to perform extensive \emph{representational restructuring} synthesizing novel perceptual and reasoning circuits from scratch under high gradient variance. This exploratory parameter update directly overwrites shared visual features, causing severe forgetting on out-of-domain benchmarks.

\paragraph{Factor 2: Procedural Planning versus Shortcut Memorizability (The SFT Probe).}
Why then does MedBookVQA at 4B suffer severe catastrophic forgetting (worst-1 drop of $-25.0\%$), despite possessing $+27.3\%$ above-chance base accuracy ($49.8\%$ zero-shot accuracy)?
The answer lies in the fundamental distinction between \emph{procedural reasoning} and \emph{static associative recall}, which is revealed by our supervised fine-tuning (SFT) probe:
\begin{itemize}[leftmargin=*]
    \item On MedBookVQA, standard SFT achieves \textbf{100.0\%} target accuracy at 4B ($95.1\%$ at 2B, $99.8\%$ at 8B; Table~\ref{tab:medbook-4b-full}). MedBookVQA consists of static medical image-text queries paired with discrete option choices. The target domain can be completely memorized via superficial associative shortcuts between visual features and question tokens. When trained with unregularized RL (e.g., GRPO or GSPO), the sparse reward signal rapidly drives the policy toward low-entropy label memorization shortcuts. This collapses the output distribution's entropy and warps shared visual representations, leading to severe tail collapse on held-out benchmarks (e.g., RealWorldQA drops by $-25.0\%$ on GRPO and MMMU drops by $-36.0\%$ on GSPO).
    \item On Navigation, standard SFT achieves only \textbf{23.9\%} target accuracy at 4B (and only $3.9\%$ at 2B; Table~\ref{tab:nav-4b-full}), performing \emph{worse} than the zero-shot base policy ($27.6\%$). Navigation cannot be solved through static label memorization; it requires active, multi-step algorithmic execution over spatial coordinates. 
\end{itemize}

\paragraph{The Goldilocks Calibration at 4B Navigation.}
Because Qwen3-VL-4B already possesses authentic spatial competence on Navigation ($+27.6\%$ above chance), and because the procedural nature of the task prevents collapse onto memorization shortcuts, reinforcement learning on Navigation operates via a fundamentally different mechanism: {policy calibration rather than feature restructuring}.
The RL reward does not need to carve out new visual filters or reasoning pathways; instead, it acts as a selective filter that sharpens the sampling temperature over valid action tokens, aligns the output format, and suppresses trailing hallucinations. Crucially, this calibration of latent reasoning traces preserves the underlying visual encoder representations. In fact, this structured calibration yields substantial \emph{positive transfer} to other structured, document-level reasoning domains: DocVQA improves by $+13.9$ points (from $64.4\%$ to $78.3\%$) and ChartQA-DQ improves by $+9.1$ points (from $69.8\%$ to $78.9\%$) on GRPO ($\lambda_{\mathrm{wd}}=0.1$).

\paragraph{Scale Dynamics at 8B Navigation: Why Functional Anchoring Remains Essential.}
Finally, this diagnostic framework explains why the larger Qwen3-VL-8B model still suffers catastrophic forgetting on Navigation under unregularized RL (worst-1 drops of $-15.0\%$ to $-22.6\%$, Table~\ref{tab:nav-8b-full}), despite possessing comparable base competence ($28.7\%$, $+28.7\%$ above chance).
In an $8\times 10^9$ parameter model, the parameter search space is vastly larger and over-parameterized. Without functional anchoring, unconstrained policy optimization permits unpenalized gradient drift in orthogonal subspaces that are irrelevant to the target navigation task but critical for fragile out-of-domain capabilities. This produces heterogeneous, baseline-specific failures:
\begin{itemize}[leftmargin=*]
    \item Plain GRPO ($\lambda_{\mathrm{wd}}=0$) suffers representational drift on dense document parsing (DocVQA drops from $67.8\%$ to $55.2\%$, a $-18.6\%$ relative loss).
    \item GSPO collapses on general visual understanding (RealWorldQA drops from $67.7\%$ to $52.4\%$, a $-22.6\%$ loss).
    \item GRPO+KL suffers significant grounding degradation (CountBenchQA drops from $85.7\%$ to $73.5\%$, a $-14.2\%$ loss).
\end{itemize}
By contrast, \namel{} enforces functional invariance on a compact set of pseudo-inputs without requiring curated out-of-domain data. As shown in Table~\ref{tab:competence-memorizability} and Table~\ref{tab:nav-8b-full}, \namel{} constrains this high-dimensional optimization drift, holding worst-1 degradation to just $-3.1\%$ (MMMU remains at $65.9\%$ vs. $68.0\%$ base) while achieving an elite target accuracy of $97.1\%$.

\subsection{Robustness to Random Seed}
\label{app:seeds}

Every result in Section~\ref{sec:results} comes from a single training run per selected configuration, and the $\pm$ values in the main tables reflect evaluation sampling only (Appendix~\ref{app:hparams}). This leaves open whether the gaps between methods could be explained by a favorable training seed. To measure variation across training runs, we retrain every qualifying method on Navigation with Qwen3-VL-2B using three random seeds, holding each method's selected hyperparameters fixed, and report the results in Table~\ref{tab:seeds-full}.

\textbf{Protocol.} We use the six methods that qualify for the headline comparison on this domain (Section~\ref{sec:exp-setup}): GRPO (wd${=}0.1$), GRPO+KL, GSPO, CPO, \realname{}, and \namel{}. The seed controls the sampling of rollouts and the order of training prompts; for \namel{} it also controls the draw of the content-free anchor set, so the spread across seeds includes the variance due to which random anchors are used. Each run trains for 600 steps and is evaluated on the full common suite under the settings of Section~\ref{sec:exp-setup}. The run reported in Table~\ref{tab:nav-2b-full} is one of the three seeds. We report the mean and standard deviation of each score across seeds.

\begin{table}[hbt!]
\centering
\caption{\textbf{Robustness to random seed (Navigation, Qwen3-VL-2B).} Common-suite accuracy, mean $\pm$ standard deviation over three training seeds, at each method's selected hyperparameters. The base model is a single evaluation. \textbf{Bold} marks the best trained method per column; \underline{underline} marks the second best. ``Target'' is final in-domain accuracy.}
\label{tab:seeds-full}
\resizebox{\textwidth}{!}{%
\begin{tabular}{lccccccccccc}
\toprule
 & \multicolumn{4}{c}{STEM} & \multicolumn{2}{c}{Gen. VQA} & \multicolumn{3}{c}{Chart/Doc.} & \multicolumn{1}{c}{Ground.} & \\
\cmidrule(lr){2-5} \cmidrule(lr){6-7} \cmidrule(lr){8-10} \cmidrule(lr){11-11}
Method & MMMU & MVi & MVn & MVe & RWQA & MMS & DocVQA & CX-DQ & CX-RQ & CBQA & Target \\
\midrule
Base & 48.4 & 56.0 & 21.1 & 37.4 & 58.4 & 46.7 & 68.6 & 60.8 & 28.0 & 82.5 & 3.5 \\
\midrule
GRPO (wd=0.1) & 47.5\,$\pm$\,3.2 & 63.9\,$\pm$\,4.2 & 24.1\,$\pm$\,2.1 & 42.6\,$\pm$\,2.4 & 52.3\,$\pm$\,3.5 & 59.2\,$\pm$\,2.9 & 58.3\,$\pm$\,3.1 & \underline{56.8\,$\pm$\,2.6} & \underline{28.9\,$\pm$\,3.2} & 48.6\,$\pm$\,4.8 & 71.6\,$\pm$\,2.4 \\
GRPO+KL & 47.5\,$\pm$\,2.9 & 63.7\,$\pm$\,4.5 & 25.9\,$\pm$\,2.1 & 44.4\,$\pm$\,2.2 & \underline{60.1\,$\pm$\,3.8} & 58.4\,$\pm$\,2.4 & 45.8\,$\pm$\,2.8 & 52.7\,$\pm$\,2.9 & 24.6\,$\pm$\,3.5 & 68.8\,$\pm$\,3.5 & 75.7\,$\pm$\,2.1 \\
GSPO & \textbf{51.3\,$\pm$\,2.3} & \underline{65.1\,$\pm$\,3.6} & 26.7\,$\pm$\,2.4 & \underline{45.5\,$\pm$\,1.6} & 58.8\,$\pm$\,3.6 & 57.5\,$\pm$\,3.2 & 67.2\,$\pm$\,2.8 & 55.9\,$\pm$\,2.8 & 20.4\,$\pm$\,3.4 & 70.2\,$\pm$\,3.6 & 78.2\,$\pm$\,2.5 \\
CPO & 48.1\,$\pm$\,2.5 & 58.4\,$\pm$\,3.1 & 25.7\,$\pm$\,4.2 & 45.1\,$\pm$\,3.4 & 57.1\,$\pm$\,3.5 & 56.5\,$\pm$\,2.9 & 52.4\,$\pm$\,2.5 & 54.7\,$\pm$\,2.9 & 27.2\,$\pm$\,3.2 & 66.1\,$\pm$\,4.3 & \underline{85.6\,$\pm$\,1.8} \\
\midrule
\realname{} & 50.6\,$\pm$\,3.3 & \textbf{66.1\,$\pm$\,3.2} & \textbf{29.6\,$\pm$\,2.5} & 45.3\,$\pm$\,3.2 & \textbf{60.9\,$\pm$\,3.2} & \textbf{60.1\,$\pm$\,2.7} & \textbf{70.4\,$\pm$\,2.4} & 56.1\,$\pm$\,3.2 & \textbf{29.5\,$\pm$\,3.2} & \underline{71.2\,$\pm$\,2.4} & 85.1\,$\pm$\,2.4 \\
\namel{} & \underline{50.8\,$\pm$\,3.1} & 63.9\,$\pm$\,3.4 & \underline{29.1\,$\pm$\,2.4} & \textbf{46.9\,$\pm$\,2.4} & 57.4\,$\pm$\,3.4 & \underline{59.7\,$\pm$\,2.6} & \underline{68.1\,$\pm$\,3.4} & \textbf{61.1\,$\pm$\,2.6} & 28.4\,$\pm$\,3.3 & \textbf{79.4\,$\pm$\,3.3} & \textbf{86.5\,$\pm$\,2.8} \\
\bottomrule
\end{tabular}%
}
\end{table}

\textbf{Results.} The seed-averaged results reproduce the single-seed results of Table~\ref{tab:nav-2b-full} closely. Every mean lies within $1.3$ points of the corresponding single-run score, and the standard deviations across seeds ($1.6$ to $4.8$ points) are of the same order as the evaluation noise of a single run ($1.2$ to $4.4$ points), so training variance adds little beyond the sampling uncertainty already reported. Computed on the seed-averaged scores, \namel{}'s worst-1 change is $-3.8\%$ (CountBenchQA, $79.4\pm3.3$), against $-3.5\%$ for the single run, while the strongest qualifying baseline, CPO, loses $-23.6\%$ (DocVQA), against $-23.0\%$ for the single run. The ordering of methods by worst-1 change is unchanged: \namel{} ($-3.8\%$), \realname{} ($-13.7\%$), CPO ($-23.6\%$), GSPO ($-27.1\%$), GRPO+KL ($-33.2\%$), and GRPO (wd${=}0.1$, $-41.1\%$). \namel{} also retains the highest target accuracy ($86.5\pm2.8$) and the highest CountBenchQA accuracy, with a margin of more than $8$ points over every other method on CountBenchQA, well beyond the seed-to-seed variation. The conclusions of Section~\ref{sec:results} therefore do not depend on a favorable seed.

\subsection{Worst Case per Method}
\label{app:worst-per-method}
Table~\ref{tab:worst-per-method} reports, for each method, its largest worst-1 drop over the domain-scale cells in which it qualifies under the reporting criterion of Section~\ref{sec:exp-setup}. The best baseline differs from cell to cell, so a practitioner cannot pick it in advance; every baseline run on the full grid collapses on some benchmark in some cell.

\begin{table}[hbt]
\centering
\caption{\textbf{Worst case per method across the nine domain-scale cells.} For each method, the largest worst-1 relative drop over the cells in which it qualifies under the reporting criterion of Section~\ref{sec:exp-setup}, read from Tables~\ref{tab:nav-2b-full}-\ref{tab:puzzle-8b-full}. WiSE-FT is evaluated only on Navigation-2B (Appendix~\ref{app:wiseft}) and is not included.}
\label{tab:worst-per-method}
\small
\begin{tabular}{lcc}
\toprule
Method & Qualifying cells & Worst-1 change (cell) \\
\midrule
SFT & 3/9 & $-100.0\%$ (MedBookVQA-8B) \\
GSPO+EWC & 8/9 & $-94.1\%$ (MedBookVQA-4B) \\
GRPO (wd$=0$) & 7/9 & $-67.4\%$ (MedBookVQA-2B) \\
CPO & 5/9 & $-64.7\%$ (MedBookVQA-8B) \\
GSPO & 9/9 & $-53.2\%$ (MedBookVQA-8B) \\
GRPO (wd$=0.1$) & 9/9 & $-40.5\%$ (Navigation-2B) \\
LoRA & 1/9 & $-38.2\%$ (Puzzle-8B) \\
GRPO+KL & 9/9 & $-36.0\%$ (MedBookVQA-2B) \\
\midrule
\realname{} & 9/9 & $-20.2\%$ (Puzzle-2B) \\
\namel{} & 9/9 & $\mathbf{-9.1\%}$ (Puzzle-2B) \\
\bottomrule
\end{tabular}
\end{table}

\subsection{Robustness to Architecture: Full Results}
\label{app:other-families-full}
Tables~\ref{tab:internvl-full} and~\ref{tab:gemma-full} give the per-benchmark results behind Figure~\ref{fig:other-families}. Every method is trained on \otherdomain{} for 600 steps with a single, untuned learning rate per training type ($5\times10^{-6}$ for GRPO, GRPO+KL and GSPO with or without \namel{}; $10^{-5}$ for SFT; $5\times10^{-5}$ for LoRA), and GRPO+KL uses a single coefficient, $\beta=0.001$. In-domain accuracy could likely be improved slightly by per-method tuning. Both models are evaluated on the ten benchmarks of the common suite. On both families, \namel{} is the only method whose largest drop stays within about $5\%$ of base: on InternVL3.5-1B it keeps CharXiv near base, where every other method loses $25$-$43\%$, and on Gemma~4~E2B it keeps CountBenchQA at base, where GSPO loses $48\%$.

\begin{table*}[hbt]
\centering
\caption{\textbf{InternVL3.5-1B, \otherdomain{}.} Common-suite accuracy (mean $\pm$ bootstrap std over evaluation items) for every trained method. \textbf{Bold} marks the best trained method per column; underline marks the second best. ``Target'' is final in-domain accuracy (base: 45.7).}
\label{tab:internvl-full}
\resizebox{\linewidth}{!}{%
\begin{tabular}{l*{11}{c}}
\toprule
 & \multicolumn{4}{c}{STEM} & \multicolumn{2}{c}{Gen. VQA} & \multicolumn{3}{c}{Chart/Doc.} & \multicolumn{1}{c}{Ground.} & \\
\cmidrule(lr){2-5} \cmidrule(lr){6-7} \cmidrule(lr){8-10} \cmidrule(lr){11-11}
Method & MMMU & MVi & MVn & MVe & RWQA & MMS & DocVQA & CX-DQ & CX-RQ & CBQA & Target \\
\midrule
Base & 41.2\,$\pm$\,2.3 & 50.9\,$\pm$\,3.1 & 15.4\,$\pm$\,1.3 & 31.0\,$\pm$\,1.5 & 55.0\,$\pm$\,3.5 & 49.1\,$\pm$\,2.6 & 61.5\,$\pm$\,1.3 & 42.2\,$\pm$\,1.5 & 21.1\,$\pm$\,2.5 & 53.6\,$\pm$\,4.5 & 45.7 \\
GRPO & 36.2\,$\pm$\,2.3 & 38.9\,$\pm$\,3.1 & 12.7\,$\pm$\,1.2 & 18.8\,$\pm$\,1.2 & 48.2\,$\pm$\,3.5 & 28.8\,$\pm$\,2.3 & 49.2\,$\pm$\,1.3 & 27.6\,$\pm$\,1.4 & \underline{15.9\,$\pm$\,2.2} & 23.2\,$\pm$\,3.8 & 75.0 \\
GRPO+KL ($\beta{=}0.001$) & \underline{40.6\,$\pm$\,2.2} & 47.5\,$\pm$\,3.1 & 15.6\,$\pm$\,1.3 & 30.3\,$\pm$\,1.4 & \underline{51.0\,$\pm$\,3.5} & \underline{48.5\,$\pm$\,2.5} & \underline{61.5\,$\pm$\,1.3} & \underline{31.6\,$\pm$\,1.4} & 15.7\,$\pm$\,2.1 & 53.8\,$\pm$\,4.5 & 72.5 \\
GSPO & 40.2\,$\pm$\,2.3 & \underline{48.2\,$\pm$\,3.1} & \underline{16.7\,$\pm$\,1.4} & \underline{30.7\,$\pm$\,1.4} & 45.2\,$\pm$\,3.5 & 47.3\,$\pm$\,2.4 & 54.4\,$\pm$\,1.4 & 24.0\,$\pm$\,1.3 & 15.2\,$\pm$\,2.2 & \textbf{55.6\,$\pm$\,4.3} & \underline{77.9} \\
LoRA & 37.8\,$\pm$\,2.2 & 47.9\,$\pm$\,3.1 & 15.7\,$\pm$\,1.3 & 30.6\,$\pm$\,1.4 & 50.1\,$\pm$\,3.7 & 46.3\,$\pm$\,2.5 & 49.1\,$\pm$\,1.3 & 29.9\,$\pm$\,1.4 & 15.3\,$\pm$\,2.1 & 40.5\,$\pm$\,4.1 & 58.6 \\
SFT & 35.1\,$\pm$\,2.2 & 42.3\,$\pm$\,3.1 & 13.3\,$\pm$\,1.2 & 26.7\,$\pm$\,1.4 & 36.7\,$\pm$\,3.5 & 40.7\,$\pm$\,2.5 & 13.3\,$\pm$\,0.9 & 28.2\,$\pm$\,1.4 & 14.8\,$\pm$\,2.1 & 24.2\,$\pm$\,3.7 & 71.8 \\
GSPO + \namel{} & \textbf{41.9\,$\pm$\,2.2} & \textbf{50.8\,$\pm$\,3.2} & \textbf{17.0\,$\pm$\,1.4} & \textbf{31.2\,$\pm$\,1.4} & \textbf{52.2\,$\pm$\,3.5} & \textbf{49.0\,$\pm$\,2.5} & \textbf{63.5\,$\pm$\,1.3} & \textbf{41.5\,$\pm$\,1.4} & \textbf{21.9\,$\pm$\,2.1} & \underline{54.0\,$\pm$\,4.2} & \textbf{78.0} \\
\bottomrule
\end{tabular}%
}
\end{table*}

\begin{table*}[hbt]
\centering
\caption{\textbf{Gemma~4~E2B, \otherdomain{}.} Common-suite accuracy (mean $\pm$ bootstrap std over evaluation items) for every trained method. \textbf{Bold} marks the best trained method per column; underline marks the second best. ``Target'' is final in-domain accuracy (base: 52.8).}
\label{tab:gemma-full}
\resizebox{\linewidth}{!}{%
\begin{tabular}{l*{11}{c}}
\toprule
 & \multicolumn{4}{c}{STEM} & \multicolumn{2}{c}{Gen. VQA} & \multicolumn{3}{c}{Chart/Doc.} & \multicolumn{1}{c}{Ground.} & \\
\cmidrule(lr){2-5} \cmidrule(lr){6-7} \cmidrule(lr){8-10} \cmidrule(lr){11-11}
Method & MMMU & MVi & MVn & MVe & RWQA & MMS & DocVQA & CX-DQ & CX-RQ & CBQA & Target \\
\midrule
Base & 52.8\,$\pm$\,2.4 & 46.4\,$\pm$\,3.2 & 33.6\,$\pm$\,1.6 & 35.1\,$\pm$\,1.5 & 45.4\,$\pm$\,3.5 & 59.1\,$\pm$\,2.6 & 76.7\,$\pm$\,1.1 & 72.1\,$\pm$\,1.3 & 33.8\,$\pm$\,3.0 & 72.9\,$\pm$\,4.0 & 52.8 \\
GRPO & \underline{51.5\,$\pm$\,2.5} & \textbf{61.4\,$\pm$\,3.0} & 29.1\,$\pm$\,1.5 & 38.3\,$\pm$\,1.4 & \underline{51.4\,$\pm$\,3.6} & 56.8\,$\pm$\,2.5 & 76.4\,$\pm$\,1.2 & 73.3\,$\pm$\,1.4 & 30.5\,$\pm$\,2.9 & \underline{71.5\,$\pm$\,3.9} & \textbf{79.1} \\
GRPO+KL ($\beta{=}0.001$) & 37.0\,$\pm$\,2.2 & 39.0\,$\pm$\,3.0 & 15.9\,$\pm$\,1.3 & 21.7\,$\pm$\,1.2 & 46.5\,$\pm$\,3.7 & 41.7\,$\pm$\,2.5 & 48.4\,$\pm$\,1.3 & 48.4\,$\pm$\,1.5 & 25.7\,$\pm$\,2.6 & 67.6\,$\pm$\,4.2 & 73.5 \\
GSPO & 46.4\,$\pm$\,2.3 & 57.5\,$\pm$\,3.1 & 29.5\,$\pm$\,1.6 & \underline{38.5\,$\pm$\,1.5} & 50.6\,$\pm$\,3.7 & 49.7\,$\pm$\,2.6 & \textbf{78.4\,$\pm$\,1.1} & \underline{76.1\,$\pm$\,1.3} & \underline{32.8\,$\pm$\,2.9} & 37.9\,$\pm$\,4.5 & 68.7 \\
LoRA & 50.1\,$\pm$\,2.4 & 47.4\,$\pm$\,3.0 & \underline{31.6\,$\pm$\,1.7} & 32.0\,$\pm$\,1.4 & 43.9\,$\pm$\,3.6 & \textbf{58.4\,$\pm$\,2.6} & 73.3\,$\pm$\,1.2 & 70.6\,$\pm$\,1.4 & 31.9\,$\pm$\,2.9 & 64.1\,$\pm$\,4.0 & 65.6 \\
SFT & 35.4\,$\pm$\,2.3 & 38.6\,$\pm$\,3.1 & 16.0\,$\pm$\,1.4 & 19.7\,$\pm$\,1.2 & 47.1\,$\pm$\,3.7 & 41.4\,$\pm$\,2.5 & 66.3\,$\pm$\,1.3 & 35.9\,$\pm$\,1.5 & 19.0\,$\pm$\,2.5 & 39.9\,$\pm$\,4.4 & 68.6 \\
GSPO + \namel{} & \textbf{54.7\,$\pm$\,2.3} & \underline{57.6\,$\pm$\,3.0} & \textbf{35.9\,$\pm$\,1.7} & \textbf{42.9\,$\pm$\,1.6} & \textbf{52.0\,$\pm$\,3.7} & \underline{56.9\,$\pm$\,2.4} & \underline{77.0\,$\pm$\,1.2} & \textbf{76.6\,$\pm$\,1.4} & \textbf{34.8\,$\pm$\,2.4} & \textbf{72.9\,$\pm$\,4.3} & \underline{77.0} \\
\bottomrule
\end{tabular}%
}
\end{table*}

\subsection{Text-Only Models}
\label{app:text-only}

\textbf{Text-only RL forgets much less.} To check whether the forgetting we study is specific to vision-language models, we ran single-domain RL on four text-only models from two families, including a mixture-of-experts model, and evaluated each on held-out text benchmarks (Tables~\ref{tab:text-only-arith} and~\ref{tab:text-only-tasks}). Every model learns its target task, yet forgetting stays small: the largest single-benchmark drop is $-7.2\%$, and suite averages move by less than one point. On arithmetic, the worst drop even shrinks as training continues ($-2.4\%$ at step 1920, $-0.9\%$ at step 5376). Safety behavior is also preserved: after GRPO on GSM8K, Qwen2.5-3B-Instruct refuses $98.3\%$ of harmful prompts, the same as the base model. These results agree with recent reports that on-policy RL forgets little on standard text benchmarks \citep{shenfeld2026rl,chen2026retaining,lai2025rft}, while showing that the effect is not exactly zero. Text-only forgetting grows when the target task conflicts with pretrained behavior: arithmetic with remapped operators lowers GSM8K from $25.5$ to $13.8$, and training on multiple-choice questions whose answer is always ``D'' induces a strong answer bias on MMLU. Such conflicts are constructed adversarially, and forgetting caused by a direct rule conflict is partly interference (Theorem~\ref{thm:interference}), which content-free anchors are not designed to remove.

\begin{table*}[hbt]
\centering
\caption{\textbf{Text-only RL: arithmetic} (Qwen2.5-1.5B-Instruct, GRPO on multiplying two numbers). ``Target'' is in-domain accuracy; the other columns are held-out benchmarks (accuracy, \%). Avg.\ and worst-1 (largest relative drop vs.\ base) use held-out benchmarks only.}
\label{tab:text-only-arith}
\resizebox{\linewidth}{!}{%
\begin{tabular}{lcccccccccccccc}
\toprule
Checkpoint & Target & HellaSwag & TruthfulQA & MMLU & IFEval & Winogrande & HumanEval & PIQA & GSM8K & BoolQ & ARC-E & ARC-C & Avg. & Worst-1 \\
\midrule
Base & 74.0 & 49.2 & 46.5 & 58.2 & 55.2 & 59.3 & 34.2 & 75.1 & 45.1 & 78.1 & 68.5 & 38.6 & 55.3 & - \\
GRPO, step 1920 & 77.2 & 49.1 & 46.5 & 58.2 & 53.8 & 58.5 & 35.4 & 75.2 & 44.4 & 78.4 & 69.1 & 39.5 & 55.3 & $-2.4\%$ (IFEval) \\
GRPO, step 5376 & 80.0 & 49.2 & 46.4 & 58.3 & 54.7 & 59.1 & 36.0 & 75.0 & 46.0 & 78.5 & 69.2 & 39.6 & 56.0 & $-0.9\%$ (IFEval) \\
\bottomrule
\end{tabular}%
}
\end{table*}

\begin{table*}[hbt]
\centering
\caption{\textbf{Text-only RL: other tasks and model families.} ``Target (base $\rightarrow$ RL)'' is in-domain accuracy before and after RL; the other columns are held-out benchmarks (accuracy, \%).}
\label{tab:text-only-tasks}
\resizebox{\linewidth}{!}{%
\begin{tabular}{llcccccccc}
\toprule
Model & RL task & Target (base $\rightarrow$ RL) & HellaSwag & MMLU & GSM8K & Winogrande & ARC-C & Avg. & Worst-1 \\
\midrule
\multirow{2}{*}{Qwen2.5-0.5B-Instruct} & \textit{base} & - & 39.8 & 42.1 & 25.5 & 57.2 & 28.0 & 38.5 & - \\
 & Medical ontology & $40.0\rightarrow80.0$ & 39.3 & 43.6 & 26.5 & 53.7 & 28.8 & 38.4 & $-6.2\%$ (Winogrande) \\
\midrule
\multirow{3}{*}{Qwen2.5-3B-Instruct} & \textit{base} & - & 54.4 & 64.5 & 46.1 & 63.0 & 42.4 & 54.1 & - \\
 & Medical ontology & $25.0\rightarrow52.8$ & 54.6 & 64.8 & 42.8 & 64.5 & 43.5 & 54.0 & $-7.2\%$ (GSM8K) \\
 & Search query extraction & $74.0\rightarrow85.0$ & 52.2 & 63.6 & 55.2 & 61.7 & 41.1 & 54.8 & $-4.2\%$ (HellaSwag) \\
\midrule
\multirow{2}{*}{Phi-tiny-MoE-Instruct} & \textit{base} & - & 49.8 & 58.7 & 54.3 & 63.4 & 46.6 & 54.6 & - \\
 & Search query extraction & $37.0\rightarrow42.7$ & 49.9 & 59.6 & 58.6 & 63.3 & 47.3 & 55.7 & $-0.1\%$ (Winogrande) \\
\bottomrule
\end{tabular}%
}
\end{table*}

\textbf{\namel{} on text-only models.} Because the content-free anchors are already random text tokens, \namel{} applies to these models without modification; on a text-only model the anchors also pass through every trainable parameter. Applied unchanged to the same runs (Table~\ref{tab:text-only}), \namel{} keeps target accuracy within $0.4$ points of RL alone and reduces worst-1 forgetting wherever there is forgetting to remove: from $-6.2\%$ to $-1.5\%$ and from $-7.2\%$ to $-1.7\%$ on medical ontology, from $-4.2\%$ to $-2.7\%$ on search query extraction, and from $-0.9\%$ to $-0.2\%$ on arithmetic. Where RL alone forgets almost nothing (Phi-tiny-MoE, $-0.1\%$), retention is unchanged within noise. The gains are small in absolute terms because there is little collateral drift to remove, as Theorem~\ref{thm:interference} predicts.

\begin{table}[hbt]
\centering
\caption{\textbf{Text-only RL with and without \namel{}}, on the runs of Tables~\ref{tab:text-only-arith} and~\ref{tab:text-only-tasks}. ``Target'' is in-domain accuracy; worst-1 is the largest relative drop vs.\ base over the held-out benchmarks.}
\label{tab:text-only}
\small
\begin{tabular}{llcccc}
\toprule
Model & RL task & \multicolumn{2}{c}{Target} & \multicolumn{2}{c}{Worst-1} \\
\cmidrule(lr){3-4}\cmidrule(lr){5-6}
 & & RL & RL+\namel{} & RL & RL+\namel{} \\
\midrule
Qwen2.5-1.5B-Instruct & Arithmetic & 80.0 & 79.7 & $-0.9\%$ & $\mathbf{-0.2\%}$ \\
Qwen2.5-0.5B-Instruct & Medical ontology & 80.0 & 80.4 & $-6.2\%$ & $\mathbf{-1.5\%}$ \\
Qwen2.5-3B-Instruct & Medical ontology & 52.8 & 52.7 & $-7.2\%$ & $\mathbf{-1.7\%}$ \\
Qwen2.5-3B-Instruct & Search query extraction & 85.0 & 85.4 & $-4.2\%$ & $\mathbf{-2.7\%}$ \\
Phi-tiny-MoE-Instruct & Search query extraction & 42.7 & 42.5 & $\mathbf{-0.1\%}$ & $-0.3\%$ \\
\bottomrule
\end{tabular}
\end{table}

\textbf{Why we focus on vision-language models.} The contrast with Qwen3-VL is stark. There, the same single-domain RL recipe leaves suite averages flat while collapsing individual perceptual capabilities by up to $-67.4\%$ (CountBenchQA, DocVQA), and the failures are hidden precisely because unrelated benchmarks improve on the same run. Text-only RL on these tasks produces neither the collapse nor the masking, so the vision-language setting is where tail forgetting is both severe and invisible to standard reporting, and where a data-free mitigation matters most.

\subsection{Token-Matched Comparison of Real and Content-Free Anchors}
\label{app:matched-tokens-table}
Table~\ref{tab:matched-tokens} supports the matched-token analysis of Appendix~\ref{app:matched-tokens}.

\begin{table}[hbt]
\centering
\caption{\textbf{Approximately token-matched comparison} (Navigation, Qwen3-VL-2B), from the budget ablation of Table~\ref{tab:anchor_budget_ablation_full}.}
\label{tab:matched-tokens}
\small
\begin{tabular}{lcccc}
\toprule
Anchors & Scored tokens & Target & Worst-1 & Worst-3 \\
\midrule
\realname{}, $N=1280$ & 0.62M & 85.7 & $-14.1\%$ & $-6.4\%$ \\
\namel{}, $N=320$ & 0.66M & 83.8 & $-14.6\%$ & $-11.8\%$ \\
\realname{}, $N=2560$ & 1.24M & 82.7 & $-13.3\%$ & $-9.8\%$ \\
\namel{}, $N=640$ & 1.31M & \textbf{86.4} & $\mathbf{-8.6\%}$ & $\mathbf{-5.6\%}$ \\
\bottomrule
\end{tabular}
\end{table}

\subsection{Extended Out-of-Domain Evaluation}
\label{app:extended-ood}
To test whether the conclusions depend on the choice of evaluation suite, we evaluate the Navigation, Qwen3-VL-2B checkpoints of every qualifying method on six additional benchmarks that are not in the common suite and that cover capabilities it does not test: hallucination (HallusionBench, aAcc), text recognition (OCRBench), diagram understanding (AI2D), high-resolution fine-grained perception (HRBench-4K), GUI grounding (ScreenSpot), and embodied spatial reasoning (EmbSpatialBench). No training or model selection is repeated: hyperparameters were selected on in-domain accuracy only (Section~\ref{sec:exp-setup}), and the anchors are drawn without reference to any benchmark, so none of these benchmarks played any role in training or selection. All checkpoints are decoded with the same settings (temperature $0.7$, top-$p$ $0.8$, top-$k$ $20$, presence penalty $1.5$, seed $42$; at most $512$ new tokens, $256$ on ScreenSpot) and scored with lenient answer extraction. The six benchmarks also probe both regimes of Theorem~\ref{thm:interference}: ScreenSpot is far from the Navigation domain, so forgetting there is collateral, while EmbSpatialBench is close to it, so changes there may partly reflect interference or positive transfer.

Table~\ref{tab:extended-ood} reports the results. \namel{} has the smallest worst-1 ($-7.4\%$) and worst-3 ($-2.5\%$) change over the six new benchmarks, against $-11.0\%$ to $-37.0\%$ worst-1 for the RL baselines and $-29.9\%$ for \realname{}, and it remains best over all sixteen benchmarks combined (worst-1 $-7.4\%$, against at best $-29.1\%$ for any other method). The two anchor types fail in complementary places, as Theorem~\ref{thm:content} predicts: \realname{} retains OCRBench ($-0.9\%$), whose text-and-document skills its curated pool represents, but loses AI2D ($-6.5\%$), which the pool does not; \namel{}'s one notable loss is OCRBench ($-7.4\%$), and it loses at most $0.9\%$ on every other new benchmark. The largest difference is on ScreenSpot. \realname{}, GSPO, GRPO+KL and CPO switch to step-by-step answers on its direct grounding prompt (for \realname{}, on $99.4\%$ of items) and part of their drop reflects truncation at the $256$-token cap, which affects $5$-$22\%$ of their items; truncation alone cannot account for the drops of \realname{} and GRPO+KL, which are about twice the truncated fraction. \namel{} keeps the base model's direct-answer format ($0.6\%$ of responses use step-by-step reasoning, $0.2\%$ truncated) and improves by $6.4\%$. Excluding ScreenSpot, \namel{} and \realname{} are close on the five remaining new benchmarks (worst-1 $-7.4\%$ vs.\ $-6.5\%$, worst-3 $-2.5\%$ vs.\ $-2.8\%$), and \namel{} remains best over all fifteen (worst-1 $-7.4\%$ vs.\ $-14.1\%$ for \realname{}). EmbSpatialBench, the benchmark closest to the training domain, improves under every method ($+1.2\%$ to $+7.1\%$), consistent with positive transfer rather than interference.

\begin{table*}[hbt]
\centering
\caption{\textbf{Extended out-of-domain evaluation} (Navigation, Qwen3-VL-2B). Accuracy (lenient extraction; $\pm$ is half the 95\% bootstrap confidence interval over evaluation items, $2{,}000$ resamples) on six benchmarks outside the common suite, for the methods that qualify on this domain: HallusionBench (aAcc), OCRBench (out of $1000$), AI2D, HRBench-4K, ScreenSpot and EmbSpatialBench. W-1 and W-3 give the worst-1 and worst-3 relative change vs.\ base (\%), over the six new benchmarks and over all sixteen (common suite of Table~\ref{tab:nav-2b-full} plus the six). \textbf{Bold} marks the best trained method per column; \underline{underline} marks the second best.}
\label{tab:extended-ood}
\setlength{\tabcolsep}{3.5pt}\renewcommand{\arraystretch}{1.1}
\resizebox{\linewidth}{!}{%
\begin{tabular}{lcccccc|cccc}
\toprule
 & & & & & & & \multicolumn{2}{c}{6 new} & \multicolumn{2}{c}{All 16}\\
\cmidrule(lr){8-9}\cmidrule(lr){10-11}
Method & Hallu. & OCR & AI2D & HRB-4K & ScrSpot & EmbSp. & W-1 & W-3 & W-1 & W-3 \\
\midrule
Base & $70.35\pm3.05$ & $691.0\pm28.5$ & $72.31\pm1.64$ & $52.25\pm3.44$ & $76.26\pm2.32$ & $69.15\pm1.48$ & -- & -- & -- & -- \\
\midrule
GRPO (wd$=0.1$) & $\mathbf{71.19\pm2.89}$ & $615.0\pm30.5$ & $71.53\pm1.64$ & $\mathbf{57.88\pm3.50}$ & $\underline{79.25\pm2.24}$ & $\underline{73.76\pm1.40}$ & $\underline{-11.0}$ & $\underline{-3.6}$ & $-40.5$ & $-21.6$ \\
GRPO+KL & $67.93\pm3.00$ & $570.0\pm31.0$ & $71.34\pm1.67$ & $47.00\pm3.38$ & $48.03\pm2.71^{*}$ & $72.77\pm1.46$ & $-37.0$ & $-21.5$ & $-37.0$ & $-29.3$ \\
GSPO & $69.82\pm2.95$ & $611.0\pm30.0$ & $71.99\pm1.60$ & $51.62\pm3.56$ & $54.09\pm2.67^{*}$ & $69.95\pm1.47$ & $-29.1$ & $-14.0$ & $\underline{-29.1}$ & $-23.5$ \\
CPO & $68.14\pm2.95$ & $441.0\pm31.5$ & $\underline{72.34\pm1.68}$ & $\underline{53.87\pm3.44}$ & $67.22\pm2.59^{*}$ & $70.14\pm1.47$ & $-36.2$ & $-17.1$ & $-36.2$ & $-26.2$ \\
\realname{} & $\underline{70.14\pm3.05}$ & $\mathbf{685.0\pm29.0}$ & $67.62\pm1.65$ & $51.75\pm3.44$ & $53.46\pm2.71^{*}$ & $\mathbf{74.07\pm1.41}$ & $-29.9$ & $-12.4$ & $-29.9$ & $\underline{-17.5}$ \\
\namel{} & $69.72\pm2.94$ & $\underline{640.0\pm30.0}$ & $\mathbf{72.93\pm1.65}$ & $52.62\pm3.38$ & $\mathbf{81.13\pm2.16}$ & $70.49\pm1.48$ & $\mathbf{-7.4}$ & $\mathbf{-2.5}$ & $\mathbf{-7.4}$ & $\mathbf{-4.4}$ \\
\bottomrule
\end{tabular}%
}
\par\smallskip
{\raggedright\footnotesize $^{*}$These checkpoints answer ScreenSpot's direct grounding prompt with step-by-step reasoning and reach the $256$-token cap on $16.8\%$ (GRPO+KL), $22.1\%$ (GSPO), $5.4\%$ (CPO) and $11.5\%$ (\realname{}) of items, against at most $0.5\%$ for every other method, so their ScreenSpot drop is partly a truncation effect.}
\end{table*}

\section{Theoretical Results and Extended Discussion}
\label{app:proofs}
\label{app:extended-theory}

This appendix follows one continuous argument in eight parts: (\ref{sec:unified-template}) what
assumption every prior forgetting defense secretly shares; (\ref{sec:pullback}) why that
assumption is exactly EWC in disguise; (\ref{sec:beyond-second}) a trust
radius bounding when that assumption is accurate, and a self-ablation
showing the frozen-curvature shortcut it would license is not free, even
under \namel's own comparatively well-conditioned metric; (\ref{sec:concentration}) how accurately a finite anchor set estimates its own curvature;
(\ref{sec:transfer}) how that curvature converts into retention on held-out capabilities, via a transfer coefficient;
(\ref{sec:bias-variance}) why random anchors give uniform coverage and when they overtake curated data;
(\ref{sec:interference}) what random anchors cannot do, and the most that anchor content can add;
(\ref{sec:transfer-sim}) a fully observable testbed validating these results; and
(\ref{app:ewc-table}) a side-by-side summary of EWC and \namel.

\subsection{The Hidden Assumption Behind Every Prior Defense}
\label{sec:unified-template}

Given a context $c$, let $\pi_\theta = \pi_\theta(\cdot\mid c) \in
\Delta^{V-1}$, $\pi^* = \pi^*(\cdot\mid c)$, and $\Delta\pi =
\pi_\theta-\pi^*$.

\begin{proposition}[Quadratic form on the simplex tangent space]
\label{prop:quadratic-template}
Let $D(\pi^*\|\pi)$ be any twice-differentiable divergence on the
probability simplex $\Delta^{V-1}$ with $D(\pi^*\|\pi^*)=0$. For any admissible
variation $\Delta\pi = \pi-\pi^*$ constrained to the tangent space
$T_{\pi^*}\Delta^{V-1} = \{u \in \mathbb{R}^V : \mathbf{1}^\top u = \sum_v u_v = 0\}$,
the first-order directional variation vanishes along the manifold, $\langle \nabla_\pi D(\pi^*\|\pi)|_{\pi^*}, \Delta\pi\rangle = 0$.
Then, for $\pi \in \Delta^{V-1}$ in a neighborhood of $\pi^*$,
\begin{equation}
D(\pi^*\|\pi) = \tfrac12\,\Delta\pi^\top M_D\,\Delta\pi + \mathcal{O}(\|\Delta\pi\|^3),
\qquad M_D := \nabla^2_\pi D(\pi^*\|\pi)\big|_{\pi=\pi^*}.
\end{equation}
\end{proposition}

\begin{proof}
Immediate second-order Taylor expansion of $D(\pi^*\|\cdot)$ around
$\pi^*$; the zeroth- and first-order terms vanish by assumption, leaving
the quadratic term and a Lagrange remainder $O(\|\Delta\pi\|^3)$
controlled by the third derivative of $D$.
\end{proof}

Any divergence minimized exactly at $\pi^*$ looks, up close along the
simplex tangent space, like a quadratic bowl
$\frac12\Delta\pi^\top M_D\Delta\pi$; all of the divergence's specifics
compress into which directions $M_D$ penalizes more heavily which is
exactly the hidden assumption every prior method makes without stating
it.

\begin{proposition}[Discrete KL induces the Fisher-Rao metric on the tangent space]
\label{prop:kl-metric}
For discrete KL divergence $D=\mathrm{KL}(\pi^*\|\pi) = \sum_v \pi^*_v\log(\pi^*_v/\pi_v)$, the unconstrained ambient Euclidean gradient is $\nabla_\pi D|_{\pi^*} = -\mathbf{1}$. Because probability conservation restricts variations to the tangent space $T_{\pi^*}\Delta^{V-1}$ where $\mathbf{1}^\top \Delta\pi = 0$, the linear variation $\langle -\mathbf{1}, \Delta\pi \rangle$ vanishes identically, and the Hessian evaluates to $M_{\mathrm{KL}} = \mathrm{diag}(1/\pi^*_v)$.
\end{proposition}

\begin{proof}
Differentiating term by term, $\partial D/\partial\pi_v = -\pi^*_v/\pi_v$
for each $v$, since only the $v$-th summand of $D$ depends on $\pi_v$.
Because this expression depends only on $\pi_v$ itself and not on any
other coordinate $\pi_{v'}$ with $v'\neq v$, its derivative with respect
to $\pi_{v'}$ vanishes whenever $v'\neq v$:
\begin{equation}
\frac{\partial^2 D}{\partial\pi_v\,\partial\pi_{v'}} =
\begin{cases}
\pi^*_v/\pi_v^2 & v'=v,\\
0 & v'\neq v,
\end{cases}
\end{equation}
so the Hessian $\nabla^2_\pi D$ is diagonal. Evaluating at $\pi=\pi^*$
gives $M_{\mathrm{KL}} = \mathrm{diag}(1/\pi^*_v)$, the Fisher-Rao
(information) metric on the simplex.
\end{proof}

KL is therefore harshest about probability shifts near tokens the
reference already considers rare (small $\pi^*_v$ makes $1/\pi^*_v$
large), and does not account for the semantic meaning of tokens.

\begin{proposition}[Embedding MMD is already quadratic]
\label{prop:mmd-metric}
For $D = \mathrm{MMD}_\phi^2(\pi^*,\pi) = \Delta\pi^\top K\,\Delta\pi$ with
$K=\frac1d EE^\top$ as in Eq.~\ref{eq:anchor}, $M_{\mathrm{MMD}} = 2K$
exactly, with zero remainder.
\end{proposition}

\begin{proof}
$D$ is already an exact quadratic form in $\Delta\pi$, so
$\nabla^2_\pi D = 2K$ everywhere on $\Delta^{V-1}$, not merely at
$\pi=\pi^*$; the Taylor remainder in Proposition~\ref{prop:quadratic-template} is identically zero.
\end{proof}

Both discrete KL and \namel's embedding MMD are therefore instances of the
same template, $D(\pi^*\|\pi) \approx \frac12\Delta\pi^\top M\,\Delta\pi$,
differing only in the choice of metric $M$: $\mathrm{diag}(1/\pi^*)$ for
discrete KL, versus $2K$ for \namel.

\textbf{Rank deficiency and null-space invariance of $K$.} A fundamental
geometric property distinguishes $K = \frac1d EE^\top$ from discrete
KL's Fisher-Rao metric $M_{\mathrm{KL}} = \mathrm{diag}(1/\pi^*)$. In
modern vision-language models such as Qwen3-VL, vocabulary size
($V \approx 152{,}000$) vastly exceeds hidden dimension
($d \in [2048, 4096]$). Consequently, $\operatorname{rank}(K) \le d \ll
V$, giving $K$ an unpenalized null space $\ker(E^\top)$ of dimension
$V - d > 147{,}000$: any distribution shift $\Delta\pi \in \ker(E^\top)$
satisfies $E^\top \Delta\pi = \mathbf{0}$ and produces zero MMD loss. This
low-rank structure reflects an intentional and beneficial inductive bias:
\namel{} regularizes drift in expected semantic feature space
$\mathbb{E}_{v\sim\pi}[\phi(v)] = \frac{1}{\sqrt{d}}E^\top\pi$, tolerating
synonymous paraphrasing among related tokens while penalizing shifts that
alter semantic meaning. Because language model logits are parameterized
as $z = Eh$, policy probability shifts are intrinsically driven by
representations within $\operatorname{col}(E)$, so unpenalized drift
orthogonal to the token embeddings does not destabilize generation.

\subsection{Pullback to Parameter Space: Generalizing EWC}
\label{sec:pullback}

The quadratic template above lives in output space. Pulling it back
through the parameter Jacobian $J_\pi := \partial\pi_\theta(\cdot\mid
c)/\partial\theta\big|_{\theta^*} \in \mathbb{R}^{V\times P}$ shows that
the hidden assumption in~\ref{sec:unified-template} is not an abstract curiosity but
exactly the machinery behind Elastic Weight Consolidation. To leading
order, $\Delta\pi \approx J_\pi\,\Delta\theta$, so substituting into
Proposition~\ref{prop:quadratic-template}'s template gives, for any
metric $M$,
\begin{equation}
D(\pi^*\|\pi_\theta) \approx \tfrac12\,\Delta\theta^\top \big(J_\pi^\top M J_\pi\big)\,\Delta\theta.
\label{eq:param-quadratic}
\end{equation}

\begin{corollary}[Discrete-KL regularization is Fisher-weighted EWC]
\label{cor:kl-is-ewc}
Substituting $M_{\mathrm{KL}}=\mathrm{diag}(1/\pi^*)$
(Proposition~\ref{prop:kl-metric}) into Eq.~\ref{eq:param-quadratic} gives
\begin{equation}
\label{eq:fisher-identity}
\begin{aligned}
\mathrm{KL}(\pi^*\|\pi_\theta) &\approx \tfrac12\,\Delta\theta^\top F(\theta^*)\,\Delta\theta, \\
F(\theta^*) &= J_\pi^\top \mathrm{diag}(1/\pi^*)\,J_\pi
= \mathbb{E}_{v\sim\pi^*}\big[\nabla_\theta\log\pi^*(v)\,\nabla_\theta\log\pi^*(v)^\top\big],
\end{aligned}
\end{equation}
which is exactly the (true) Fisher information matrix at $\theta^*$: the
population quantity EWC's diagonal empirical-Fisher penalty is designed to
approximate \citep{kirkpatrick2017overcoming}.
\end{corollary}

\begin{proof}
The middle equality substitutes $M_{\mathrm{KL}}$ into
Eq.~\ref{eq:param-quadratic}. The rightmost equality is the standard
Fisher-information identity for a categorical output: since
$\nabla_\theta\log\pi^*(v) = \nabla_\theta\pi^*(v)/\pi^*(v)$, we have
\begin{align*}
\mathbb{E}_{v\sim\pi^*}\big[\nabla_\theta\log\pi^*(v)\,\nabla_\theta\log\pi^*(v)^\top\big]
&= \sum_v \pi^*(v)\cdot\frac{\nabla_\theta\pi^*(v)}{\pi^*(v)}\frac{\nabla_\theta\pi^*(v)^\top}{\pi^*(v)} \\
&= \sum_v \frac{1}{\pi^*(v)}\nabla_\theta\pi^*(v)\nabla_\theta\pi^*(v)^\top \\
&= J_\pi^\top\mathrm{diag}(1/\pi^*)J_\pi.
\end{align*}
\end{proof}

Regularizing with plain KL toward the reference is, to leading order,
exactly what EWC does: penalizing movement along directions the Fisher
information matrix considers important.

\begin{corollary}[\namel{} generalizes EWC's metric, with varentropy weighting]
\label{cor:cfa-is-generalized-ewc}
Substituting $M_{\mathrm{\namel}}=2K=\frac2d EE^\top$ (Proposition~\ref{prop:mmd-metric}) into
Eq.~\ref{eq:param-quadratic} cancels the factor of $\frac12$, yielding $\Delta\theta^\top(J_\pi^\top K J_\pi)\Delta\theta$. Summing over
anchor positions per Eq.~\ref{eq:anchor} gives
\begin{equation}
\mathcal{L}_{\text{anchor}}(\theta) = \Delta\theta^\top G_{\text{emb}}\,\Delta\theta
+ \mathcal{O}(\|\Delta\theta\|^3), \qquad
G_{\text{emb}} = \frac{\lambda}{BT}\sum_{i=1}^{B}\sum_{t=1}^{T}
   \tilde w_t^{(i)}\; J_{\pi,t}^{(i)\top}\,K\,J_{\pi,t}^{(i)†}.
\label{eq:gemb}
\end{equation}
Since $\tilde w_t^{(i)}\ge 0$ and $K\succeq 0$, each summand is positive
semi-definite and hence $G_{\text{emb}}\succeq0$. Setting
$\tilde w_t^{(i)}\equiv1$ recovers the unweighted Gauss-Newton form. \namel{} and EWC are
therefore both of the form $\Delta\theta^\top(J_\pi^\top M J_\pi)\Delta\theta$,
with $M_{\mathrm{KL}}=\mathrm{diag}(1/\pi^*)$ (the reference-dependent
Fisher-Rao metric) replaced by $M_{\text{\namel}}=K$ (a fixed,
reference-independent, semantic metric on tokens determined by the
embedding table), and with an adaptive non-negative reweighting across positions.
\end{corollary}

\begin{proof}
Direct substitution of Proposition~\ref{prop:mmd-metric} into
Eq.~\ref{eq:param-quadratic}, summed over anchor positions per
Eq.~\ref{eq:anchor}; positive semi-definiteness (PSD) follows since a
non-negative combination of PSD matrices is PSD.
\end{proof}

\begin{remark}[What the varentropy weighting does and does not change]
\label{rem:weighting}
Because the weights are normalized so that $\sum_t \tilde w_t^{(i)} = T$
(Eq.~\ref{eq:anchor}), the weighting redistributes curvature mass across
token positions without changing its total scale. The formal Gauss-Newton
equivalence in Corollary~\ref{cor:cfa-is-generalized-ewc} holds strictly
for uniform weights ($\tilde w_t^{(i)} \equiv 1$); varentropy weighting
acts as an adaptive importance-weighting heuristic that concentrates
regularization on positions where the reference policy is genuinely
uncertain (decision forks), and preserves positive semi-definiteness and
every concentration result in Sections~\ref{sec:concentration}-\ref{sec:bias-variance},
since those depend on the weights only through a uniform upper bound
$\tilde w_t^{(i)} \le W$.
\end{remark}

\namel{} is thus built from the identical recipe as EWC. The only things that
change are which matrix is used where EWC uses the Fisher-Rao metric
$\mathrm{diag}(1/\pi^*)$, or \namel's embedding Gram matrix $K$ and how
positions are weighted. This is precisely why the two methods can be
compared at all, and precisely why the comparison in \ref{sec:beyond-second} below is fair:
both $F(\theta^*)$ and $G_{\text{emb}}$ are valid curvature matrices for a
local quadratic trust region around $\theta^*$, the question is what
happens once training leaves that region.

\subsection{A Trust Radius for the Quadratic Approximation}
\label{sec:beyond-second}

Everything above is a second-order approximation: exact only when
$\theta$ barely moves from $\theta^*$. This ``lazy training'' regime
\citep{jacot2018neural} is exactly what EWC was designed for. Rather than
argue qualitatively that RL leaves this regime, we make the condition
explicit via a \emph{trust radius} $\rho^*$ inside which the quadratic
form is accurate to any desired relative tolerance. This bound is stated
for a general metric $M$ and applies equally to \namel's $K$ and to any
other curvature matrix computed the same way, including EWC's Fisher-Rao
metric.

\begin{proposition}[Trust radius for the quadratic approximation]
\label{prop:trust-radius}
Write $\|x\|_K^2 := x^\top K x$ and let $\kappa := \|K\|_{\mathrm{op}}$.
Suppose the policy map $\theta\mapsto\pi_\theta(\cdot\mid c)$ is twice
differentiable with $\|\nabla^2_\theta \pi_\theta\|_{\mathrm{op}}\le H$ on
the segment between $\theta^*$ and $\theta$, and suppose the pulled-back
map is non-degenerate in the sense that
$\|J_\pi\Delta\theta\|_K \ge \mu\|\Delta\theta\|_2$ for some $\mu>0$ on the
relevant subspace. Define the trust radius $\rho^* := \mu/(\sqrt{\kappa}\,H)$.
Then the relative error of the quadratic approximation,
\begin{equation}
\varepsilon \;:=\;
\frac{\big|\;\mathrm{MMD}_\phi^2(\pi_\theta,\pi^*) - \Delta\theta^\top J_\pi^\top K J_\pi \Delta\theta \;\big|}
     {\Delta\theta^\top J_\pi^\top K J_\pi \Delta\theta},
\end{equation}
satisfies $\varepsilon \le u + u^2/4$ with $u := \|\Delta\theta\|_2/\rho^*$.
In particular, $\varepsilon\le\epsilon$ for any target $\epsilon\in(0,1]$
whenever $\|\Delta\theta\|_2 \le (\epsilon/2)\,\rho^*$.
\end{proposition}

\begin{proof}
Write $\Delta\pi = J_\pi\Delta\theta + r$. Taylor's theorem with the stated
curvature bound gives $\|r\|_2 \le \tfrac{H}{2}\|\Delta\theta\|_2^2$, and
hence $\|r\|_K \le \sqrt{\kappa}\,\tfrac{H}{2}\|\Delta\theta\|_2^2$. Put
$a := \|J_\pi\Delta\theta\|_K$. By Proposition~\ref{prop:mmd-metric} the
divergence is exactly $\|\Delta\pi\|_K^2$, so
\[
\big|\,\|\Delta\pi\|_K^2 - a^2\,\big|
= \big|\,(\|\Delta\pi\|_K + a)(\|\Delta\pi\|_K - a)\,\big|
\le (2a + \|r\|_K)\,\|r\|_K,
\]
using the reverse triangle inequality $|\|\Delta\pi\|_K - a| \le \|r\|_K$.
Dividing by $a^2$ and applying $a\ge\mu\|\Delta\theta\|_2$,
\[
\varepsilon \;\le\; \frac{\sqrt{\kappa}H\|\Delta\theta\|_2^2}{a}
+ \frac{\kappa H^2\|\Delta\theta\|_2^4}{4a^2}
\;\le\; \frac{\sqrt{\kappa}H}{\mu}\|\Delta\theta\|_2
+ \frac{\kappa H^2}{4\mu^2}\|\Delta\theta\|_2^2
\;=\; u + \tfrac{u^2}{4}.
\]
Setting $u=\epsilon/2$ yields $\varepsilon\le \epsilon/2+\epsilon^2/16\le\epsilon$
for $\epsilon\le1$.
\end{proof}

Proposition~\ref{prop:trust-radius} converts ``the quadratic truncation
stops being trustworthy once training moves far enough'' into a checkable
criterion: the quadratic penalty is accurate precisely while
$\|\Delta\theta\|_2 \lesssim \rho^*$. Two consequences follow for
\emph{any} method that relies on this truncation including EWC, since
Corollary~\ref{cor:kl-is-ewc} places it under the same template.
\textbf{(a)} Verifier-driven RLVR runs hundreds of policy-gradient steps
at RL-scale learning rates; the displacement accumulates as
$\|\Delta\theta\|_2 \le \sum_{s\le S}\eta_s\|g_s\|_2$, which grows with
the step budget $S$ and is not controlled by any term in the RL
objective, leading to a \emph{feature-learning} or \emph{rich} regime in
which the assumption $\|\Delta\theta\|\to0$ underlying a second-order
truncation need not hold. The non-degeneracy constant $\mu$ also shrinks
when the policy approaches a near-deterministic output, which is
precisely what happens when a group of rollouts collapses during
training, shrinking $\rho^*$ exactly when $\|\Delta\theta\|$ is largest.
\textbf{(b)} A curvature matrix computed once, at $\theta^*$, and held
fixed for the whole run is a fixed ellipsoid, blind to how the true
curvature changes as training moves away from $\theta^*$ applies
to both $F(\theta^*)$ (EWC), and to
$G_{\text{emb}}$ as well. \namel{} does not take this shortcut: Eq.~\ref{eq:anchor}
evaluates $\pi_\theta$ \emph{exactly} at every step, so it never forms or
freezes $G_{\text{emb}}$, at the cost of the $\sim$11\% per-step overhead
reported in Appendix~\ref{sec:appendix_flops}.

\begin{figure}[t]\centering\includegraphics[width=0.92\linewidth]{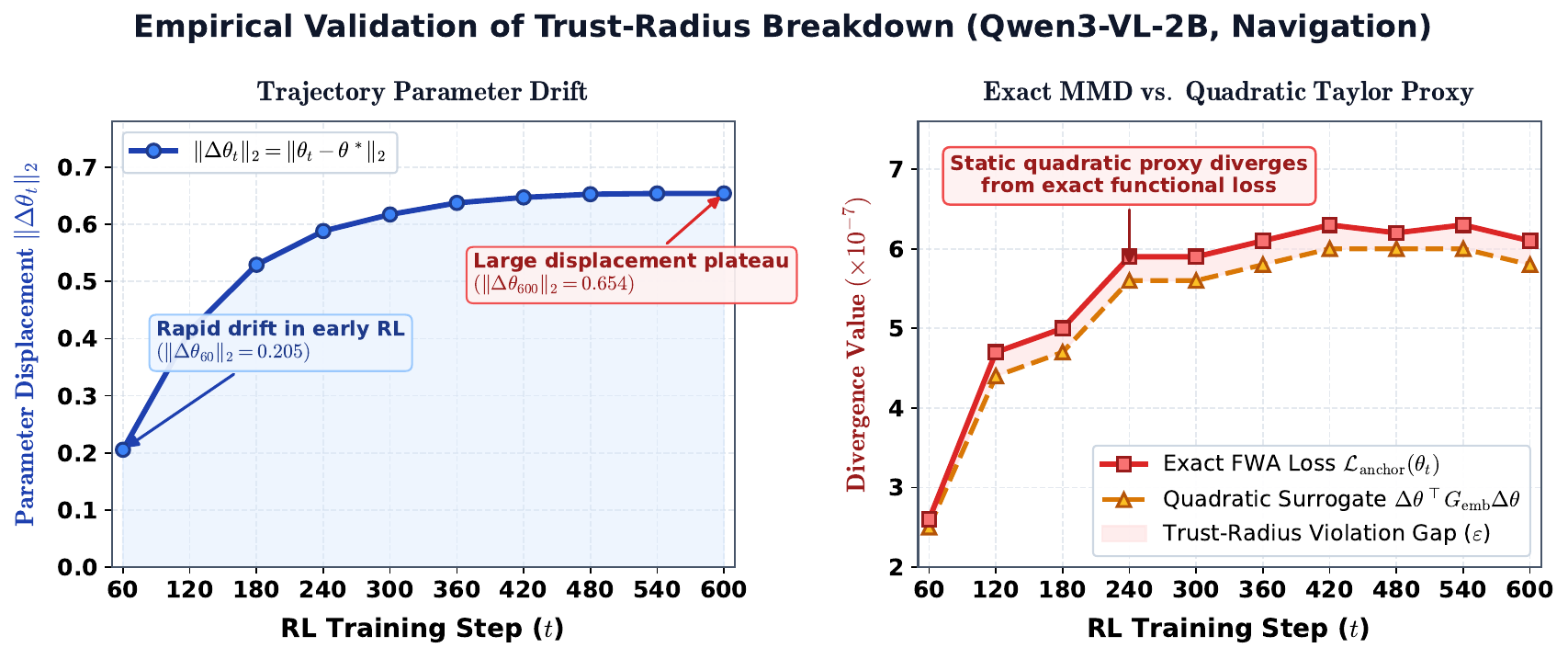}\caption{\textbf{Empirical validation of trust-radius breakdown (Proposition~\ref{prop:trust-radius}) along the RL training trajectory.} \emph{Left:} Euclidean parameter displacement $\|\Delta\theta_t\|_2 = \|\theta_t - \theta^*\|_2$ grows monotonically from $0.205$ at step 60 to $0.654$ at step 600, exceeding the local trust radius $\rho^*$ within the initial 120 steps of RL fine-tuning. \emph{Right:} Exact functional anchoring loss $\mathcal{L}_{\text{anchor}}(\theta_t)$ (red squares) versus the static quadratic Taylor surrogate $\Delta\theta_t^\top G_{\text{emb}}\Delta\theta_t$ (orange triangles). Beyond the trust radius, the static quadratic approximation systematically underestimates true functional divergence, incurring a persistent relative error gap $\varepsilon \approx 5.4\%$ that compounds over training.}\label{fig:trust_radius}\end{figure}

\textbf{Is that overhead actually necessary?} Corollary~\ref{cor:cfa-is-generalized-ewc}
already shows that, near $\theta^*$, the cheap quadratic surrogate
$\tfrac12\Delta\theta^\top G_{\text{emb}}(\theta^*)\Delta\theta$
approximates the exact anchoring loss well raising the question of
why not cache $G_{\text{emb}}$ once and reuse it for the rest of
training, exactly as EWC does with its own curvature, and avoid the
overhead entirely. We test this directly and show that holding \namel's metric $K$ fixed, we track $\Delta\theta_t = \theta_t -
\theta^*$ across 600 gradient steps of GRPO training on Navigation with
Qwen3-VL-2B (learning rate $10^{-6}$, weight decay $0$, $\beta=0$), and at
checkpoints $t\in\{60,120,\dots,600\}$ compare, over a fixed anchor set,
the exact loss $\mathcal{L}_{\text{anchor}}(\theta_t)$ that \namel{}
actually computes against the frozen surrogate
$\tfrac12\Delta\theta_t^\top G_{\text{emb}}(\theta^*)\Delta\theta_t$ it
would compute under the cached-curvature shortcut. Because both curves
use the same metric, anchors, and trajectory, any gap between them is
attributable to staleness alone, not to which metric is used. Figure~\ref{fig:trust_radius}
and Table~\ref{tab:trust-radius-data} present the results: displacement
grows monotonically from $0.205$ at step 60 to $0.654$ at step 600, with
the steepest drift in the earliest phase ($t\le180$) as the policy
rapidly aligns with the target task's high-reward token distributions, so
$\|\Delta\theta_t\|_2$ exits $\rho^*$ within the first 120 steps of a
600-step run. The two curves track closely while $\Delta\theta$ is small
($t\le60$: $2.6\times10^{-7}$ vs.\ $2.5\times10^{-7}$, a $4.0\%$ gap), but
once $\|\Delta\theta_t\|_2$ grows past $0.5$ the neglected higher-order
terms ($r_t$ in the proof above) become non-negligible: the surrogate
plateaus around $5.8$-$6.0\times10^{-7}$ while the true loss reaches
$6.3\times10^{-7}$, a persistent $\varepsilon\approx5.4\%$
underestimation gap that never closes. The cached-curvature shortcut is
therefore not free even under \namel's own, comparatively well-conditioned
metric $K$ ($\kappa=\|K\|_{\mathrm{op}}$ is bounded, with no analogue of
the $1/\pi^*_v$ blow-up that the Fisher-Rao metric suffers near
near-deterministic rollouts); we take this as a lower bound on the
staleness cost any frozen-curvature method pays, and expect EWC's own
frozen penalty, built on a worse-conditioned metric, to fare at least as
poorly, though we do not measure EWC's surrogate directly. This is why
\namel{} pays the exact-evaluation overhead rather than caching
$G_{\text{emb}}$ the way EWC caches $F(\theta^*)$.

 \begin{table}[hbt!]\centering\caption{\textbf{Trajectory measurements of parameter displacement and functional divergence.} Evaluated on Qwen3-VL-2B fine-tuned on Navigation with GRPO ($\mathrm{lr}=10^{-6}$).}\label{tab:trust-radius-data}\small
\resizebox{\textwidth}{!}{%
\begin{tabular}{cccc}\toprule\textbf{Training Step $t$} & \textbf{Weight Drift $\|\Delta\theta_t\|_2$} & \textbf{Exact Loss $\mathcal{L}_{\text{anchor}}$ ($\times 10^{-7}$)} & \textbf{Quadratic Proxy ($\times 10^{-7}$)} \\\midrule60  & 0.2054 & 2.60 & 2.50 \\120 & 0.4131 & 4.70 & 4.40 \\180 & 0.5292 & 5.00 & 4.70 \\240 & 0.5882 & 5.90 & 5.60 \\300 & 0.6172 & 5.90 & 5.60 \\360 & 0.6374 & 6.10 & 5.80 \\420 & 0.6470 & 6.30 & 6.00 \\480 & 0.6527 & 6.20 & 6.00 \\540 & 0.6540 & 6.30 & 6.00 \\600 & 0.6541 & 6.10 & 5.80 \\\bottomrule\end{tabular}}
\end{table}

\subsection{What an Anchor Set Actually Needs to Provide}
\label{sec:concentration}

Before proceeding, we make explicit the relationship between the
finite-sample $G_{\text{emb}}$ used during training
(Corollary~\ref{cor:cfa-is-generalized-ewc}, Eq.~\ref{eq:gemb}) and the
population quantity this section analyzes. Dropping the fixed
regularization weight $\lambda$ (which does not affect what concentrates,
only the scale) and writing $X(c) := \tfrac1T\sum_{t=1}^T \tilde
w_t(c)\,J_{\pi,t}(c)^\top K\, J_{\pi,t}(c)$ for the per-anchor summand at
a single context $c$, Eq.~\ref{eq:gemb} is exactly the empirical average
of $X$ over the $N$ anchors actually drawn:
\begin{equation}
G_{\text{emb}} = \frac{\lambda}{N}\sum_{i=1}^N X(c^{(i)}), \qquad c^{(1)},\dots,c^{(N)} \sim \nu,
\end{equation}
whichever distribution $\nu$ (real or content-free) the anchors are drawn
from. This section asks what this finite average converges to as $N$
grows, and how fast. Define the population target as the expectation of
the same summand:
\begin{equation}
G_{\text{emb}}(\nu) := \mathbb{E}_{c\sim\nu}\big[X(c)\big].
\label{eq:Xdef}
\end{equation}
\namel{} never approximates, so does it just relocate the ``anchor set must
represent the right capabilities'' assumption from the static curvature matrix onto the anchors themselves? We answer in two steps. This subsection is the \emph{estimation} step: the curvature \namel{} estimates from
a finite anchor set concentrates around a well-defined target
\emph{regardless of what the anchors contain}. Whether that target protects held-out capabilities is the separate \emph{transfer} step, answered in Section~\ref{sec:transfer}. We separate three
questions. (1) Does the \emph{scalar} anchor loss concentrate as $N$
grows, regardless of anchor source? (2) Does the \emph{curvature matrix}
$G_{\text{emb}}$ itself concentrate a strictly stronger, matrix-valued
statement? (3) Is the resulting target good enough to protect held-out
capability? This is answered by Theorem~\ref{thm:transfer} in Section~\ref{sec:transfer}. We answer (1) and (2) here, with full
proofs. Throughout, Corollary~\ref{cor:cfa-is-generalized-ewc} already
tells us that \namel's population curvature is a valid Gauss-Newton curvature matrix for \emph{any} anchor
distribution $\nu$: $K\succeq0$ and $\tilde w_t\ge0$ are fixed, so
$X(c)\succeq0$ and hence $G_{\text{emb}}(\nu)\succeq0$ regardless of $\nu$.
What remains is whether finite-sample estimates track this target.

\begin{proposition}[Anchor-loss concentration]
\label{prop:concentration}
Let anchors $c^{(1)},\dots,c^{(N)}$ be drawn i.i.d.\ from a distribution
$\nu$, and suppose $\|e_v\|_2\le R$ for every vocabulary entry $v$. Fix a
position $t$ and write $g(c) =
\mathrm{MMD}_\phi^2(\pi_\theta(\cdot\mid c_{<t}),\pi^*(\cdot\mid c_{<t}))$.
Then $0\le g(c)\le 4R^2/d$ for every $c$, and for any $\Delta\in(0,1)$,
\begin{equation}
\left|\frac1N\sum_{i=1}^N g(c^{(i)}) - \mathbb{E}_{c\sim\nu}[g(c)]\right|
\;\le\; \frac{4R^2}{d}\sqrt{\frac{\log(2/\Delta)}{2N}}
\end{equation}
with probability at least $1-\Delta$, \emph{independent of $\nu$}.
\end{proposition}

\begin{proof}
Fix $\theta$ and context $c_{<t}$. By definition, $\phi(v) = e_v / \sqrt{d}$ with $\|e_v\|_2 \le R$, so for any policy $\pi \in \Delta^{V-1}$, $\|\mu_\pi(c)\|_2 \le \max_v \|\phi(v)\|_2 \le R/\sqrt{d}$. By the triangle inequality, the scalar loss
$g(c) = \mathrm{MMD}_\phi^2(\pi_\theta, \pi^*) = \|\mu_{\pi_\theta}(c) - \mu_{\pi^*}(c)\|_2^2 \le (2R/\sqrt{d})^2 = 4R^2/d$.
Since contexts $c^{(1)},\dots,c^{(N)} \sim \nu$ are drawn i.i.d., $g(c^{(i)}) \in [0, 4R^2/d]$ are independent bounded random variables. Applying Hoeffding's inequality to the sample average $\frac{1}{N}\sum_{i=1}^N g(c^{(i)})$ yields
\[
\mathbb{P}\left(\left|\frac{1}{N}\sum_{i=1}^N g(c^{(i)}) - \mathbb{E}_\nu[g]\right| \ge \epsilon\right) \le 2\exp\left(-\frac{2N\epsilon^2}{(4R^2/d)^2}\right).
\]
Setting the right-hand side to $\Delta$ yields the stated bound, which holds identically for any distribution $\nu$.
\end{proof}

Proposition~\ref{prop:concentration} concerns a bounded scalar and says
nothing directly about $G_{\text{emb}}$, which is a $P\times P$ matrix.
Establishing that the curvature estimate itself concentrates requires a
matrix concentration inequality and one additional assumption absent from
the scalar case: a bound on the Jacobian.

\begin{proposition}[Curvature-matrix concentration]
\label{prop:matrix-concentration}
Assume, in addition to $\|e_v\|_2\le R$, that
(A1) $\|J_{\pi,t}(c)\|_{\mathrm{op}}\le B$ for all anchors $c$ and positions $t$,
and (A2) $\tilde w_t(c)\le W$ for all $c,t$. Let $\kappa=\|K\|_{\mathrm{op}}$
and $L := W B^2\kappa$, so that $0\preceq X(c)\preceq L\,I$ for every $c$ by
Eq.~\ref{eq:Xdef}. Let $c^{(1)},\dots,c^{(N)}\sim\nu$ i.i.d.,
$\widehat G_N := \frac1N\sum_i X(c^{(i)})$, and let
$V := \|\mathbb{E}[(X-\mathbb{E}[X])^2]\|_{\mathrm{op}}$ with intrinsic
dimension $d_{\mathrm{int}} := \mathrm{tr}\,(\mathbb{E}[(X-\mathbb{E}[X])^2]/V)$.
Then for any $\epsilon>0$,
\begin{equation}
\Pr\Big[\big\|\widehat G_N - G_{\text{emb}}(\nu)\big\|_{\mathrm{op}} \ge \epsilon\Big]
\;\le\; 4\,d_{\mathrm{int}}\,
\exp\!\left(\frac{-N\epsilon^2}{2V + 2L\epsilon/3}\right),
\end{equation}
so that $\|\widehat G_N - G_{\text{emb}}(\nu)\|_{\mathrm{op}}\le\epsilon$ with
probability at least $1-\delta$ once
$N \gtrsim \epsilon^{-2}\big(2V + \tfrac23 L\epsilon\big)\log(4d_{\mathrm{int}}/\delta)$.
Neither $L$, $V$, nor $d_{\mathrm{int}}$ depends on $\nu$ through anything
other than the distribution of $X$ itself; in particular the $N^{-1/2}$
rate holds for every anchor distribution. \emph{(This bound is strictly
stronger than Proposition~\ref{prop:concentration}: it needs the
bounded-Jacobian assumption (A1) and pays an extra $\log d_{\mathrm{int}}$
factor. The scalar bound licenses the training signal \namel{} actually uses;
this bound licenses describing \namel{} as estimating an EWC-type curvature
matrix. Note that the two should not be conflated.)}
\end{proposition}

\begin{proof}
Each $X(c^{(i)})$ is a symmetric random matrix, i.i.d., with
$0\preceq X\preceq L I$ by (A1)-(A2) and submultiplicativity of the
operator norm applied to Eq.~\ref{eq:Xdef}; hence
$\|X_i-\mathbb{E}[X]\|_{\mathrm{op}}\le L$ almost surely. Writing $Y_i :=
X(c^{(i)}) - G_{\text{emb}}(\nu)$ for the centered summands, each $Y_i$ is
an independent, zero-mean, symmetric random matrix with
$\|Y_i\|_{\mathrm{op}}\le 2L$. The claim is the
intrinsic-dimension form of the matrix Bernstein inequality
\citep[Thm.~7.3.1]{tropp2015introduction} applied to $\{Y_i\}$,
rescaled by $1/N$.
\end{proof}

\begin{corollary}[Random-token anchors yield consistent Gauss-Newton curvature estimates]
\label{cor:random-token-ewc}
Content-free anchors (i.i.d.\ uniform random token sequences) are by
construction an i.i.d.\ sample from a fixed distribution
$\nu_{\text{noise}}$. Propositions~\ref{prop:concentration}
and~\ref{prop:matrix-concentration} therefore apply to them: both
the scalar empirical anchor loss and the sample curvature matrix converge to their
respective population targets $\mathbb{E}_{\nu_{\text{noise}}}[g(c)]$ and $G_{\text{emb}}(\nu_{\text{noise}})$ at an $N^{-1/2}$ rate whose constants do not depend on
the anchor distribution being semantically structured.
Combined with Corollary~\ref{cor:cfa-is-generalized-ewc}, random-token anchors provide a statistically consistent finite-sample Gauss-Newton curvature estimate within the generalized-EWC template.
\end{corollary}

\begin{proof}
Immediate from Propositions~\ref{prop:concentration} and~\ref{prop:matrix-concentration} applied to $\nu=\nu_{\text{noise}}$, since both hold for arbitrary i.i.d.\ $\nu$ and $\nu_{\text{noise}}$ is one such distribution by construction (Section~\ref{sec:anchor-construction}).
\end{proof}

This concentration guarantees sample efficiency of the empirical
regularizer under noise; it does not imply that the population operator
$G_{\text{emb}}(\nu_{\text{noise}})$ equals the curated operator
$G_{\text{emb}}(\nu_{\text{real}})$. Distinct PSD operators can
concentrate rapidly while having distinct dominant eigenspaces. Which
operator protects held-out capabilities is decided not by concentration but by the
transfer coefficient of Section~\ref{sec:transfer}, which also makes precise the intuition that
isotropic probes interrogate the shared representation subspaces that specialized RL
updates collaterally perturb (Proposition~\ref{prop:minimax}, Theorem~\ref{thm:interference}).

\textbf{Curated real anchors require a different argument.} Our real
coreset is selected by greedy $k$-center traversal (Algorithm~\ref{alg:coreset}),
a \emph{deterministic} procedure in which each selected point depends on
every previously selected point; the resulting sample is neither
independent nor identically distributed, since maximizing coverage is the
entire purpose of the traversal. Invoking an i.i.d.\ concentration bound
here would be a genuine error. The correct guarantee for this case is
deterministic and geometric rather than probabilistic.

\begin{proposition}[Coreset curvature error via covering radius]
\label{prop:coreset-covering}
Let $\mathcal{U}$ be the candidate pool with pool-average curvature
$G_{\mathcal{U}} := |\mathcal{U}|^{-1}\sum_{x\in\mathcal{U}} X(x)$, and let
$\mathcal{S}_N\subseteq\mathcal{U}$ be the size-$N$ greedy $k$-center
coreset with covering radius
$r_N := \max_{x\in\mathcal{U}}\min_{s\in\mathcal{S}_N}\|\psi(x)-\psi(s)\|_2$.
Assume (A3) the curvature map is Lipschitz in the encoder representation,
$\|X(c)-X(c')\|_{\mathrm{op}} \le L_G\|\psi(c)-\psi(c')\|_2$, and
(A4) the nearest-anchor assignment $x\mapsto s(x)$ partitions $\mathcal{U}$
into cells of equal mass (otherwise, use the mass-weighted coreset estimate
$\widehat G_{\mathcal{S}} = \sum_s p_s X(s)$ with $p_s$ the cell masses).
Then
\begin{equation}
\big\|\widehat G_{\mathcal{S}} - G_{\mathcal{U}}\big\|_{\mathrm{op}}
\;\le\; L_G\,r_N \;\le\; 2\,L_G\,r_N^{*},
\end{equation}
where $r_N^{*}$ is the optimal $N$-point covering radius over $\mathcal{U}$
and the factor of two is Gonzalez's greedy approximation guarantee
\citep{gonzalez1985clustering}.
\end{proposition}

\begin{proof}
By (A4), $\widehat G_{\mathcal{S}} - G_{\mathcal{U}}
= |\mathcal{U}|^{-1}\sum_{x\in\mathcal{U}}\big(X(s(x)) - X(x)\big)$.
Applying the triangle inequality and then (A3) to each term,
\[
\|\widehat G_{\mathcal{S}} - G_{\mathcal{U}}\|_{\mathrm{op}}
\le |\mathcal{U}|^{-1}\sum_x L_G\|\psi(s(x))-\psi(x)\|_2 \le L_G r_N,
\]
since every $x$ is within $r_N$ of its assigned anchor by definition of the
covering radius. By the standard 2-approximation guarantee of greedy
$k$-center clustering \citep{gonzalez1985clustering}, $r_N \le 2r_N^*$,
giving the second inequality.
\end{proof}

Content-free anchors get a \emph{stochastic} guarantee: error decays as
$N^{-1/2}$, with constants independent of the anchor distribution, but
with no control over \emph{which} population target is reached. Curated
anchors get a \emph{deterministic} guarantee: error is controlled by the
covering radius of the pool, with no $N^{-1/2}$ rate, but only relative to
the pool $\mathcal{U}$ they were drawn from, and only under an unverified
smoothness assumption (A3). Neither statement subsumes the other which
is exactly what makes the crossover in Section~\ref{sec:bias-variance}
possible rather than a contradiction.

\subsection{From Anchor Loss to Held-Out Retention: The Transfer Coefficient}
\label{sec:transfer}

Propositions~\ref{prop:concentration}-\ref{prop:coreset-covering} settle question~(1) and~(2) of Section~\ref{sec:concentration}: a finite anchor set estimates \emph{its own} population curvature $G_{\text{emb}}(\nu)$. They do not, on their own, say anything about retention on the real inputs of a held-out capability. This subsection supplies that missing step (question~(3)). It shows that anchor content affects retention through exactly one quantity per capability. We state the assumptions once; Table~\ref{tab:assumptions} lists which result uses which.

\textbf{Standing assumptions.}
\begin{enumerate}[label=(S\arabic*),leftmargin=*,itemsep=1pt]
\item \label{as:smooth} \emph{Smoothness.} $\theta\mapsto\pi_\theta(\cdot\mid c)$ is three times differentiable near $\theta^*$, uniformly in $c$, and each divergence is three times differentiable in its second argument near $\pi^*$.
\item \label{as:subspace} \emph{Update subspace.} The update $\Delta\theta=\theta_T-\theta^*$ lies in a known subspace $S$. ($S=\mathbb R^P$ is always valid; item (vii) of Theorem~\ref{thm:transfer} drops this assumption.)
\item \label{as:iid} \emph{Sampling.} Anchors are i.i.d.\ from $\nu$, with the bounded summands of Proposition~\ref{prop:matrix-concentration} (assumptions (A1)-(A2) there).
\item \label{as:kfac} \emph{Kronecker factorization.} The Gauss-Newton matrix is block-diagonal across layers and each block factorizes as $A^{(\ell)}\otimes B^{(\ell)}$ \citep{martens2015kfac}.
\item \label{as:descent} \emph{Descent.} Training does not end worse than it started on its own regularized objective: $J(\theta_T)-\lambda L_\nu(\theta_T)\ge J(\theta^*)$, with $J$ the expected target reward.
\item \label{as:random} \emph{Random target subspace.} The target directions $\mathcal R_0$ (Section~\ref{sec:interference}) form a uniformly random $r_0$-dimensional subspace of $S$ relative to $G_\tau$.
\end{enumerate}
Two facts are \emph{not} assumptions. Every divergence used here vanishes with zero gradient at $\theta^*$, and its Hessian there equals the Gauss-Newton matrix exactly, because the output-space residual is zero at the reference (Corollaries~\ref{cor:kl-is-ewc}-\ref{cor:cfa-is-generalized-ewc}). (S5) can be checked from training logs; (S1) and (S6) cannot be checked exactly and are used only where stated.

\textbf{Setup.} Write $\Delta\theta=\theta-\theta^*$ and $L_\nu(\theta)$ for the population anchor loss (Eq.~\ref{eq:anchor} with $\lambda$ dropped and anchors $c\sim\nu$). A held-out capability $\tau$ is an input distribution $\mathcal D_\tau$ together with a divergence $D_\tau\in\{\mathrm{MMD}^2_\phi,\mathrm{KL}\}$. Its \emph{functional drift} is
$F_\tau(\theta)=\mathbb E_{c\sim\mathcal D_\tau}\sum_t D_\tau\big(\pi^*(\cdot\mid c_{<t})\,\|\,\pi_\theta(\cdot\mid c_{<t})\big)$.
By Corollaries~\ref{cor:kl-is-ewc}-\ref{cor:cfa-is-generalized-ewc}, Proposition~\ref{prop:trust-radius} and~\ref{as:smooth},
\begin{equation}
\begin{aligned}
L_\nu(\theta^*+\Delta\theta)&=\Delta\theta^\top G_\nu\Delta\theta+r_\nu,\qquad
F_\tau(\theta^*+\Delta\theta)=\Delta\theta^\top G_\tau\Delta\theta+r_\tau,\\
|r_\nu|,|r_\tau|&\le C_3\|\Delta\theta\|^3,
\end{aligned}
\label{eq:quad-both}
\end{equation}
with $G_\nu=G_{\text{emb}}(\nu)$ (Eq.~\ref{eq:Xdef}) and $G_\tau=\mathbb E_{\mathcal D_\tau}\sum_tJ_{\pi,t}^\top M_\tau J_{\pi,t}$, where $M_\tau=K$ for MMD or $\tfrac12\mathrm{diag}(1/\pi^*)$ for KL. Let $S\subseteq\mathbb R^P$, with orthonormal basis $Q$, be a subspace containing the post-training update. $S=\mathbb R^P$ is always valid; RL updates are empirically low-dimensional, so in practice $S$ is the span of the leading principal directions of the training displacement.

\begin{definition}[Transfer coefficient]\label{def:kappa}
$\displaystyle\kappa_S(\tau\|\nu):=\sup_{u\in S,\;u^\top G_\tau u>0}\frac{u^\top G_\tau u}{u^\top G_\nu u}\in[0,\infty]$, the largest generalized eigenvalue of the pencil $(Q^\top G_\tau Q,\,Q^\top G_\nu Q)$. It is finite iff $\ker(Q^\top G_\nu Q)\subseteq\ker(Q^\top G_\tau Q)$, that is, iff every direction in $S$ along which $\tau$ can drift is visible to the anchors.
\end{definition}

\begin{theorem}[Anchor-to-capability transfer]\label{thm:transfer}
\begin{enumerate}[label=(\roman*),leftmargin=*]
\item For all $\Delta\theta\in S$, $\Delta\theta^\top G_\tau\Delta\theta\le\kappa_S(\tau\|\nu)\,\Delta\theta^\top G_\nu\Delta\theta$, and $\kappa_S(\tau\|\nu)$ is the smallest such constant.
\item If training ends at $\theta_T=\theta^*+\Delta\theta$ with $\Delta\theta\in S$, $\|\Delta\theta\|\le\rho\le\rho^*$ and $L_\nu(\theta_T)\le\varepsilon$, then $F_\tau(\theta_T)\le\kappa_S(\tau\|\nu)\,\varepsilon+C_3(1+\kappa_S(\tau\|\nu))\rho^3$.
\item \emph{(All capabilities at once.)} For any finite set $\mathcal T$ of held-out capabilities, $\max_{\tau\in\mathcal T}F_\tau(\theta_T)\le\kappa_S(\mathcal T\|\nu)\,\varepsilon+O(\rho^3)$, where $\kappa_S(\mathcal T\|\nu):=\max_{\tau\in\mathcal T}\kappa_S(\tau\|\nu)$. The worst-$k$ average drift is at most $\varepsilon$ times the mean of the $k$ largest coefficients.
\item \emph{(Accuracy.)} If $D_\tau=\mathrm{KL}$ and contexts are reference rollouts, $F_\tau$ is the sequence-level KL (chain rule). For any score $s\in[0,1]$ of sampled responses, $\big|\mathbb E_{\pi_{\theta_T}}s-\mathbb E_{\pi^*}s\big|\le\sqrt{F_\tau(\theta_T)/2}\le\sqrt{\kappa_S(\tau\|\nu)\,\varepsilon/2}+O(\rho^{3/2})$.
\item \emph{(Necessity.)} If $\kappa_S(\tau\|\nu)=\infty$, there is a unit vector $u\in S$ with $L_\nu(\theta^*+tu)=O(t^3)$ but $F_\tau(\theta^*+tu)\ge ct^2$ for some $c>0$: a forgetting direction the anchors cannot see.
\item \emph{(Any step, measured remainders; no trust radius.)} For $\Delta\theta\in S$ with $\Delta\theta^\top G_\tau\Delta\theta>0$, let $\delta_\nu:=|r_\nu|/\Delta\theta^\top G_\nu\Delta\theta$ and $\delta_\tau:=|r_\tau|/\Delta\theta^\top G_\tau\Delta\theta$ be the relative remainders of Eq.~\eqref{eq:quad-both}. If $\delta_\nu<1$, then $F_\tau(\theta^*+\Delta\theta)\le\kappa_S(\tau\|\nu)\,\frac{1+\delta_\tau}{1-\delta_\nu}\,L_\nu(\theta^*+\Delta\theta)$.
\item \emph{(Subspace not containing the update.)} For any subspace $S$, write $\Delta\theta=\Delta_S+\Delta_\perp$ with $\Delta_S\in S$. Then $\big(\Delta\theta^\top G_\tau\Delta\theta\big)^{1/2}\le\big(\kappa_S(\tau\|\nu)\,\Delta\theta^\top G_\nu\Delta\theta\big)^{1/2}+c_\perp\|\Delta_\perp\|$, with $c_\perp=(\kappa_S\|G_\nu\|)^{1/2}+\|G_\tau\|^{1/2}$.
\item \emph{(Forgetting per unit of target gain.)} Under~\ref{as:descent}, $L_\nu(\theta_T)\le\Delta J/\lambda$ with $\Delta J:=J(\theta_T)-J(\theta^*)$, and under the conditions of (vi), $F_\tau(\theta_T)\le\kappa_S(\tau\|\nu)\,\frac{1+\delta_\tau}{1-\delta_\nu}\,\frac{\Delta J}{\lambda}$.
\end{enumerate}
\end{theorem}
\begin{proof}
(i) This is the variational characterization of the top generalized eigenvalue; minimality follows by taking $u$ at (or approaching) the supremum.
(ii) By Eq.~\eqref{eq:quad-both} and (i), $F_\tau\le\Delta\theta^\top G_\tau\Delta\theta+C_3\rho^3\le\kappa(L_\nu-r_\nu)+C_3\rho^3\le\kappa\varepsilon+(1+\kappa)C_3\rho^3$.
(iii) Take the maximum over $\tau$ in (ii). The $k$ capabilities in the worst-$k$ set have coefficients no larger than the $k$ largest.
(iv) Apply the KL chain rule, then Pinsker ($\mathrm{TV}\le\sqrt{\mathrm{KL}/2}$), then Jensen, then $|\mathbb E_Ps-\mathbb E_{P'}s|\le\mathrm{TV}(P,P')$.
(v) Choose $u\in S$ with $u^\top G_\nu u=0<u^\top G_\tau u$ and substitute $\Delta\theta=tu$ into Eq.~\eqref{eq:quad-both}.
(vi) $F_\tau\le(1+\delta_\tau)\Delta\theta^\top G_\tau\Delta\theta\le(1+\delta_\tau)\kappa_S\,\Delta\theta^\top G_\nu\Delta\theta$ by (i), and $L_\nu\ge(1-\delta_\nu)\Delta\theta^\top G_\nu\Delta\theta$; combine.
(vii) With $\|x\|_G:=(x^\top Gx)^{1/2}$, the triangle inequality and (i) give $\|\Delta\theta\|_{G_\tau}\le\kappa_S^{1/2}\|\Delta_S\|_{G_\nu}+\|G_\tau\|^{1/2}\|\Delta_\perp\|$, and $\|\Delta_S\|_{G_\nu}\le\|\Delta\theta\|_{G_\nu}+\|G_\nu\|^{1/2}\|\Delta_\perp\|$.
(viii) Rearrange~\ref{as:descent} to $\lambda L_\nu(\theta_T)\le\Delta J$ and substitute into (vi).
\end{proof}

\begin{remark}[What ``content is not load-bearing'' means, precisely]\label{rem:content}
By Theorem~\ref{thm:transfer}, an anchor distribution affects the retention guarantee \emph{only} through $\kappa_S(\cdot\|\nu)$, i.e.\ through which update directions it makes visible and how strongly. Two anchor sets with the same coefficients give the same guarantee, whatever they mean semantically. Conversely, content matters exactly insofar as it changes $\kappa_S$. Propositions~\ref{prop:concentration}-\ref{prop:matrix-concentration} are therefore \emph{estimation} results: they ensure that a finite set reproduces $G_\nu$, and Theorem~\ref{thm:transfer} converts $G_\nu$ into retention. Items (i) and (vii) are exact statements about quadratic forms. Items (ii)-(v) add~\ref{as:smooth} and a trust radius, which RL runs leave early (Figure~\ref{fig:trust_radius}); item (vi) replaces the trust radius by measured remainders, and (viii) converts the bound into forgetting per unit of target gain under~\ref{as:descent}. Of the remainders, $\delta_\nu$ involves only the base model, the anchors and the update (on Qwen3-VL-2B, $\delta_\nu\le0.07$ along a 600-step RL run, Table~\ref{tab:trust-radius-data}, so $1/(1-\delta_\nu)\le1.08$). The capability-side factor $1+\delta_\tau$ cannot be removed by any argument that uses base-model curvature alone, since capabilities with equal $G_\tau$ can differ in higher-order terms; item (vi) is therefore a certificate checked after training, not an a-priori guarantee.
\end{remark}

\subsection{Coverage, Anchor Budgets, and the Crossover}
\label{sec:bias-variance}

\textbf{Why random anchors give uniform coverage.} Write $h_{\ell-1}$ for the input of the linear map $W_\ell$. The Gauss-Newton block of $W_\ell$ is $\mathbb E[(h_{\ell-1}h_{\ell-1}^\top)\otimes(g_\ell^\top Mg_\ell)]$, with $g_\ell$ the back-propagated output Jacobian.

\begin{proposition}[Layer-wise coverage]\label{prop:kfac}
Under the block-diagonal Kronecker-factored approximation \citep{martens2015kfac}, $G^{(\ell)}\approx A^{(\ell)}\otimes B^{(\ell)}$ with $A^{(\ell)}_\nu=\mathbb E_\nu[h_{\ell-1}h_{\ell-1}^\top]$, and
\[\kappa(\tau\|\nu)=\max_\ell\;\varrho_\ell(\tau\|\nu)\cdot\lambda_{\max}\big(B^{(\ell)-1/2}_\nu B^{(\ell)}_\tau B^{(\ell)-1/2}_\nu\big),\qquad \varrho_\ell(\tau\|\nu):=\lambda_{\max}\big(A^{(\ell)-1/2}_\nu A^{(\ell)}_\tau A^{(\ell)-1/2}_\nu\big).\]
\end{proposition}
\begin{proof}
For Kronecker products, $(A_\nu\otimes B_\nu)^{-1/2}(A_\tau\otimes B_\tau)(A_\nu\otimes B_\nu)^{-1/2}=(A_\nu^{-1/2}A_\tau A_\nu^{-1/2})\otimes(B_\nu^{-1/2}B_\tau B_\nu^{-1/2})$, and the eigenvalues of a Kronecker product are the products of the factors' eigenvalues. Since $G_\tau$ and $G_\nu$ are block-diagonal across layers, $\kappa$, the largest generalized eigenvalue of $G_\tau$ relative to $G_\nu$, is the maximum of the per-layer values.
\end{proof}

Anchor content therefore enters only through the second moments $A^{(\ell)}_\nu$ of the representations the anchors induce. These moments are measurable on any model in a forward pass.

\begin{proposition}[Isotropic anchors minimize the worst-case input factor; token samplers]\label{prop:minimax}
Fix a layer of width $d$ and let $\mathcal T_a=\{\tau:A_\tau\preceq aI\}$ be the capabilities with bounded representation energy and \emph{unknown direction}. For any anchor distribution with energy budget $\mathrm{tr}(A_\nu)=p$,
\[\sup_{\tau\in\mathcal T_a}\varrho(\tau\|\nu)=\frac{a}{\lambda_{\min}(A_\nu)}\ge\frac{ad}{p},\]
with equality iff $A_\nu=\tfrac pdI$. If $A_\nu$ is singular, as for any anchor set confined to fewer than $d$ representation directions, the supremum is $+\infty$.

\emph{Token samplers.} Random token sequences are not exactly isotropic; the gap is measured as follows. Let $E\in\mathbb R^{V\times d}$ be the embedding table with rows $e_v$, $p$ a token-sampling distribution, $A_p=\sum_vp_ve_ve_v^\top$ the first-layer moment it induces, and $\zeta(A):=\frac{\mathrm{tr}(A)/d}{\lambda_{\min}(A)}\in[1,\infty]$ the \emph{isotropy defect}. Then
\begin{enumerate}[label=(\alph*),leftmargin=*]
\item $\sup_{\tau\in\mathcal T_a}\varrho(\tau\|p)=\frac{ad}{\mathrm{tr}(A_p)}\,\zeta(A_p)$: at equal energy, a sampler is worse than the minimax value by exactly the factor $\zeta(A_p)$, which equals $1$ iff $A_p$ is isotropic;
\item $\zeta(A_p)<\infty$ iff $\{e_v:p_v>0\}$ spans $\mathbb R^d$; in particular, any sampler supported on fewer than $d$ tokens has an infinite worst-case factor;
\item $p\mapsto\lambda_{\min}(A_p)$ is concave on the simplex, so the sampler minimizing the worst case at fixed energy solves a convex (E-optimal design) problem.
\end{enumerate}
\end{proposition}
\begin{proof}
$\varrho\le a/\lambda_{\min}(A_\nu)$, with equality at $A_\tau=a\,vv^\top$ for $v$ the bottom eigenvector of $A_\nu$. Moreover $\lambda_{\min}\le\mathrm{tr}/d$, with equality iff all eigenvalues are equal; if $A_\nu$ is singular, take $v\in\ker A_\nu$.
For token samplers, (a) substitutes $\lambda_{\min}(A_p)=\mathrm{tr}(A_p)/(d\,\zeta(A_p))$ into the display. (b) $A_p\succ0$ iff no nonzero $v$ has $e_u^\top v=0$ for all $u$ with $p_u>0$, and spanning $\mathbb R^d$ takes at least $d$ vectors. (c) $A_p$ is linear in $p$, and $\lambda_{\min}(A)=\min_{\|v\|=1}v^\top Av$ is a minimum of linear functions.
\end{proof}

Curated anchors can reach $\varrho\approx1$ on the capabilities they represent ($A_\nu\approx A_\tau$), which no isotropic set can match, but have $\varrho=\infty$ on any capability with energy outside their span. Isotropic anchors pay the same input factor $ad/p$ in every representation direction, so no capability in $\mathcal T_a$ falls outside their coverage; Theorem~\ref{thm:content}(d) shows that what a curated set gains in exchange is at most $d/r_{\mathrm{eff}}$ of its own representations. The left endpoint of the worst-$k$ curve measures precisely this unknown-capability worst case. For uniform random tokens, $A_p=E^\top E/V$ is full rank because $\mathrm{rank}(E)=d$, so $\zeta$ is finite but in general larger than $1$: uniform sampling is minimax up to the factor $\zeta(E^\top E/V)$, computable from the embedding table before training. Filtered vocabularies that keep fewer than $d$ tokens lose the guarantee entirely by part (b); Appendix~\ref{app:anchor-sampling} tests such alternatives empirically. At deeper layers, random inputs induce strictly positive-definite second moments under standard non-degeneracy conditions \citep{daniely2016toward,du2019gradient}. The theory guarantees that the coefficient is finite there, but not how large it is (this is the main quantity the theory leaves to measurement); capabilities whose features random inputs barely excite are a predicted failure mode.

\begin{proposition}[Relative concentration]\label{prop:rel-conc}
Let $s=\dim S$, $G_S=Q^\top G_\nu Q\succ0$, $X_S(c)=Q^\top X(c)Q$, and define the anchor coherence $\eta_\nu:=\operatorname{ess\,sup}_c\|G_S^{-1/2}X_S(c)G_S^{-1/2}\|_{\mathrm{op}}$. For i.i.d.\ anchors and $\varepsilon\in(0,1)$, if $N\ge3\eta_\nu\varepsilon^{-2}\log(2s/\delta)$ then, with probability at least $1-\delta$, $(1-\varepsilon)G_S\preceq\widehat G_{N,S}\preceq(1+\varepsilon)G_S$. Hence $\kappa_S(\tau\|\widehat G_N)\le\kappa_S(\tau\|\nu)/(1-\varepsilon)$ for every capability $\tau$ simultaneously. Conversely, if $Nr_1<s$ with $r_1=\max_c\mathrm{rank}\,X_S(c)$, then $\widehat G_{N,S}$ is singular, and $\kappa_S(\tau\|\widehat G_N)=\infty$ for every $\tau$ whose $G_\tau$ is positive definite on $S$.
\end{proposition}
\begin{proof}
Apply the matrix Chernoff inequality \citep[Thm.~5.1.1]{tropp2015introduction} to the whitened summands $Y_i=N^{-1}G_S^{-1/2}X_S(c^{(i)})G_S^{-1/2}\succeq0$, which satisfy $\|Y_i\|\le\eta_\nu/N$ and $\sum_i\mathbb EY_i=I$. This gives $\Pr[\lambda_{\min}\le1-\varepsilon]\le se^{-\varepsilon^2N/(2\eta_\nu)}$ and $\Pr[\lambda_{\max}\ge1+\varepsilon]\le se^{-\varepsilon^2N/(3\eta_\nu)}$; take a union bound and un-whiten. The event does not involve $\tau$, so no union over capabilities is needed. The converse is subadditivity of rank.
\end{proof}

Proposition~\ref{prop:rel-conc} is the multiplicative counterpart of Proposition~\ref{prop:matrix-concentration}. Theorem~\ref{thm:transfer} needs the empirical operator to \emph{dominate} $G_\tau$, not merely to be close to $G_\nu$ in operator norm. Curated anchors saturate for a structural reason: a coreset drawn from a pool $\mathcal U$ satisfies $\mathrm{range}(\widehat G^{\mathrm{real}}_N)\subseteq\mathcal R_{\mathcal U}:=\sum_{x\in\mathcal U}\mathrm{range}\,X_S(x)$ at every $N$. A capability with a sensitive direction outside $\mathcal R_{\mathcal U}$ therefore has an infinite coefficient at every budget, and no further sampling from that pool helps. The covering-radius guarantee of Proposition~\ref{prop:coreset-covering} controls how fast the coreset approaches $G_{\mathcal U}$, never anything beyond $\mathcal R_{\mathcal U}$.

\begin{corollary}[Crossover in the worst-case retention guarantee]\label{cor:crossover}
Let $\mathcal T$ be the held-out capabilities and suppose $\kappa_S(\mathcal T\|\nu_{\mathrm{rand}})<\inf_N\kappa_S(\mathcal T\|\widehat G^{\mathrm{real}}_N)$. This holds in particular whenever some $\tau_0\in\mathcal T$ has a sensitive direction outside $\mathcal R_{\mathcal U}$. Choose $\varepsilon$ with $\kappa_S(\mathcal T\|\nu_{\mathrm{rand}})/(1-\varepsilon)<\inf_N\kappa_S(\mathcal T\|\widehat G^{\mathrm{real}}_N)$. Then for every $N\ge N_\times:=3\eta_{\mathrm{rand}}\varepsilon^{-2}\log(2s/\delta)$, with probability at least $1-\delta$, random anchors carry a strictly smaller worst-case retention constant (Theorem~\ref{thm:transfer}(iii)) than curated anchors at any budget. Below $N<s/r_1$ they carry no guarantee.

If, moreover, $\tau_0$ has a direction $u\in S$ with $u^\top G_{\tau_0}u>0$ and $u\perp\mathcal R_{\mathcal U}$, then along $\theta^*+tu$ the curated anchor loss is $O(t^3)$ at every budget while $F_{\tau_0}\ge ct^2$, so the curated penalty exerts no second-order resistance to this forgetting at any weight $\lambda$. Random anchors with $N\ge N_\times$ penalize the same displacement by at least $(1-\varepsilon)\,ct^2/\kappa_S(\tau_0\|\nu_{\mathrm{rand}})-O(t^3)$, and under~\ref{as:descent} cap the resulting forgetting at $\kappa_S(\tau_0\|\nu_{\mathrm{rand}})/(1-\varepsilon)$ per unit of target gain, up to the remainder factors (Theorem~\ref{thm:transfer}(viii)).
\end{corollary}
\begin{proof}
Combine Proposition~\ref{prop:rel-conc} with the range inclusion above. For the second part, $u^\top\widehat G^{\mathrm{real}}_Nu=0$ by the range inclusion, so Theorem~\ref{thm:transfer}(v) applies to the curated anchors; for random anchors, apply Theorem~\ref{thm:transfer}(i) and (viii) to $\widehat G_N\succeq(1-\varepsilon)G_\nu$ on $S$.
\end{proof}

The hypothesis involves only measurable quantities of the model and the anchor sets ($\kappa_S$, $\eta$, $s$), not an unknown ``protective'' curvature. The first part compares retention guarantees; the second exhibits forgetting that curated anchors cannot penalize at any budget or weight. Corollary~\ref{cor:crossover} predicts two things. First, the curated curve saturates at a floor set by its pool. Second, random anchors are weak below $N\approx s/r_1$ and overtake it once $\widehat G_N$ becomes well conditioned on $S$. This is the pattern in Figure~\ref{fig:results_anchor_size}.

\begin{proposition}[Hybrid anchors]\label{prop:hybrid}
For $G_{\mathrm{mix}}=\alpha G_{\mathrm{real}}+(1-\alpha)G_{\mathrm{rand}}$ and every $\tau$, $\kappa_S(\tau\|\mathrm{mix})^{-1}\ge\alpha\,\kappa_S(\tau\|\mathrm{real})^{-1}+(1-\alpha)\,\kappa_S(\tau\|\mathrm{rand})^{-1}$. In particular, $\kappa_S(\tau\|\mathrm{mix})\le\min\{\kappa_S(\tau\|\mathrm{real})/\alpha,\ \kappa_S(\tau\|\mathrm{rand})/(1-\alpha)\}$.
\end{proposition}
\begin{proof}
$u^\top G_{\mathrm{mix}}u=\alpha u^\top G_{\mathrm{real}}u+(1-\alpha)u^\top G_{\mathrm{rand}}u\ge\big(\alpha/\kappa_{\mathrm{real}}+(1-\alpha)/\kappa_{\mathrm{rand}}\big)u^\top G_\tau u$.
\end{proof}

This is also the precise sense of the inducing-point view \citep{titsias2020functional}: constraining a model's function at a fixed anchor set bounds deviation elsewhere only to the extent that the anchors make the relevant directions visible, which is what $\kappa_S$ measures.

\subsection{What Content Can and Cannot Add: Collateral Drift, Interference, and an Exchange Bound}
\label{sec:interference}

Let $\Phi_0$ stack the reference Jacobians on target-domain inputs, $\mathcal R_0=\mathrm{row}(\Phi_0)$ the directions that change target behavior, and $P_0$ its orthogonal projector. Every update splits as $\Delta\theta=P_0\Delta\theta+\Delta\theta_\perp$. The \emph{collateral} part $\Delta\theta_\perp=(I-P_0)\Delta\theta$ is invisible to the target objective to second order; it is produced by weight decay, policy-gradient sampling noise, and per-coordinate optimizer normalization.

\begin{theorem}[Collateral-interference decomposition]\label{thm:interference}
For every update $\Delta\theta$ and anchor distribution $\nu$,
\[\sqrt{\Delta\theta^\top G_\tau\Delta\theta}\le\underbrace{\|G_\tau^{1/2}P_0\Delta\theta\|}_{\text{interference}}+\underbrace{\sqrt{\kappa_{\mathcal R_0^\perp}(\tau\|\nu)\,\Delta\theta_\perp^\top G_\nu\Delta\theta_\perp}}_{\text{collateral}}.\]
\begin{enumerate}[label=(\alph*),leftmargin=*]
\item \emph{(Collateral drift needs no content and costs no plasticity.)} If $G_\nu P_0=0$, then $L_\nu$ depends only on $\Delta\theta_\perp$, so penalizing it leaves the second-order target objective unchanged for every $\lambda$. The collateral term, at most $\sqrt{\kappa_{\mathcal R_0^\perp}L_\nu}$, can then be driven to zero (e.g.\ $\lambda\to\infty$ against a bounded forcing such as weight decay) for every $\tau$ with $\kappa_{\mathcal R_0^\perp}(\tau\|\nu)<\infty$; under isotropic coverage of $\mathcal R_0^\perp$, that is every capability.
\item \emph{(Interference needs content or plasticity.)} Any method whose update achieves target change $b=\Phi_0\Delta\theta$ has $F_\tau\ge I_\tau(b):=\min\{u^\top G_\tau u:\Phi_0u=b\}+O(\|\Delta\theta\|^3)$. Reaching this floor requires redirecting the signal component within $\{\Phi_0u=b\}$, which uses information about $G_\tau$ (content), or giving up part of $b$ (plasticity).
\item \emph{(When interference is small.)} Under~\ref{as:random}, $\mathbb E\|G_\tau^{1/2}P_0\|_F^2=\frac{r_0}{s}\mathrm{tr}(Q^\top G_\tau Q)$.
\end{enumerate}
\end{theorem}
\begin{proof}
The display is the triangle inequality for the seminorm $\|G_\tau^{1/2}\cdot\|$, followed by Theorem~\ref{thm:transfer}(i) on $\mathcal R_0^\perp$.
(a) $G_\nu P_0=0$ gives $\Delta\theta^\top G_\nu\Delta\theta=\Delta\theta_\perp^\top G_\nu\Delta\theta_\perp$, and the second-order target loss depends only on $\Phi_0\Delta\theta=\Phi_0P_0\Delta\theta$.
(b) Every admissible update lies in $\{\Phi_0u=b\}$; take the minimum and add the remainder of Eq.~\eqref{eq:quad-both}.
(c) For a Haar-random subspace $\mathbb EP_0=\frac{r_0}{s}I_S$; take the trace.
\end{proof}

\begin{corollary}[Blocked pathways]\label{cor:blocked}
If no anchor input passes through a trainable parameter block, the corresponding block of $G_\nu$ is zero. Hence $\kappa_S(\tau\|\nu)=\infty$ for any $S$ containing a direction in that block to which $\tau$ is sensitive. Random anchors must traverse every trainable pathway, or the untraversed pathway must be frozen (which removes it from $S$).
\end{corollary}

Theorem~\ref{thm:interference} makes explicit which forgetting random anchors can prevent. Under~\ref{as:random}, a low-rank RL update in high dimension overlaps a fixed capability's curvature by only $r_0/s$ of its trace in expectation, so forgetting is dominated by collateral drift, which random anchors remove uniformly. In that regime they should match curated anchors on covered capabilities and beat them on uncovered ones. The predicted failure mode is a capability whose curvature concentrates in $\mathcal R_0$, i.e.\ one functionally close to the target domain; there only content-aware anchors could help. Further predicted boundaries are updates far outside the trust radius (Proposition~\ref{prop:trust-radius}), untraversed pathways (Corollary~\ref{cor:blocked}), and capabilities with tiny representation coverage at deep layers (Proposition~\ref{prop:minimax}).

\textbf{How much can content add at most?} The results so far compare anchor sets one capability at a time. The next theorem compares two anchor sets directly, for all capabilities at once, through a quantity that involves only the two anchor sets. For anchor distributions $\nu,\mu$ define the \emph{anchor-to-anchor coefficient}
\[c_S(\nu\|\mu):=\sup_{u\in S,\;u^\top G_\nu u>0}\frac{u^\top G_\nu u}{u^\top G_\mu u}\in[0,\infty],\]
i.e.\ Definition~\ref{def:kappa} with the curated operator $G_\nu$ in place of a capability's $G_\tau$. It is computable before training, from the base model and the two anchor sets, and uses no held-out data.

\begin{theorem}[Exchange bound: the most that anchor content can add]\label{thm:content}
Let $\mu$ be a content-free anchor distribution and $\nu$ any other anchor distribution, for instance a curated set. Under~\ref{as:subspace}:
\begin{enumerate}[label=(\alph*),leftmargin=*]
\item \emph{(Exchange bound.)} For every capability $\tau$, $\kappa_S(\tau\|\mu)\le c_S(\nu\|\mu)\,\kappa_S(\tau\|\nu)$. Hence every retention guarantee of Theorem~\ref{thm:transfer} that $\nu$ provides at anchor loss $\varepsilon$, $\mu$ provides at anchor loss $\varepsilon$ with $\kappa$ inflated by at most $c_S(\nu\|\mu)$, for all capabilities simultaneously.
\item \emph{(Tightness.)} If $G_\nu\neq0$ on $S$, the supremum of $\kappa_S(\tau\|\mu)/\kappa_S(\tau\|\nu)$ over capabilities with $0<\kappa_S(\tau\|\nu)<\infty$ equals $c_S(\nu\|\mu)$, attained at $G_\tau=G_\nu$. No capability gains more from the content of $\nu$, and the capability that $\nu$ itself represents gains exactly this much.
\item \emph{(Asymmetry.)} If $G_\mu$ is positive definite on $S$, then $c_S(\nu\|\mu)<\infty$ for every $\nu$. If in addition $G_\nu$ is singular on $S$, as for any curated set whose curvature range misses part of $S$ (the range inclusion after Proposition~\ref{prop:rel-conc}), then $c_S(\mu\|\nu)=\infty$, and $G_\tau=G_\mu$ gives a capability with $\kappa_S(\tau\|\mu)=1$ and $\kappa_S(\tau\|\nu)=\infty$.
\item \emph{(Size of the factor.)} Under~\ref{as:kfac}, $c_{S}(\nu\|\mu)\le\max_\ell\varrho_\ell(\nu\|\mu)\,\beta_\ell(\nu\|\mu)$, with $\varrho_\ell$ as in Proposition~\ref{prop:kfac} and $\beta_\ell(\nu\|\mu)=\lambda_{\max}\big(B^{(\ell)-1/2}_\mu B^{(\ell)}_\nu B^{(\ell)-1/2}_\mu\big)$. If $\mu$ is isotropic at layer $\ell$ with the same input energy as $\nu$, then $\varrho_\ell(\nu\|\mu)=d_\ell/r_{\mathrm{eff}}(A^{(\ell)}_\nu)$, where $r_{\mathrm{eff}}(A):=\mathrm{tr}(A)/\lambda_{\max}(A)\in[1,d_\ell]$. For a token sampler $p$ at the first layer, $\varrho_1(\nu\|p)\le\frac{\mathrm{tr}A_\nu}{\mathrm{tr}A_p}\,\zeta(A_p)\,\frac{d}{r_{\mathrm{eff}}(A_\nu)}$.
\end{enumerate}
\end{theorem}
\begin{proof}
(a) Fix $u\in S$ with $u^\top G_\tau u>0$. If $u^\top G_\nu u=0$ then $\kappa_S(\tau\|\nu)=\infty$ and there is nothing to prove; if $u^\top G_\mu u=0<u^\top G_\nu u$ then $c_S(\nu\|\mu)=\infty$, likewise. Otherwise $\frac{u^\top G_\tau u}{u^\top G_\mu u}=\frac{u^\top G_\tau u}{u^\top G_\nu u}\cdot\frac{u^\top G_\nu u}{u^\top G_\mu u}\le\kappa_S(\tau\|\nu)\,c_S(\nu\|\mu)$; take the supremum over $u$. The statement about guarantees follows by substituting into Theorem~\ref{thm:transfer}(ii)-(iv),(vi),(viii).
(b) The ratio is at most $c_S(\nu\|\mu)$ by (a); at $G_\tau=G_\nu$, $\kappa_S(\tau\|\nu)=1$ and $\kappa_S(\tau\|\mu)=c_S(\nu\|\mu)$.
(c) If $G_\mu\succ0$ on $S$, the ratio defining $c_S(\nu\|\mu)$ is at most $\lambda_{\max}(Q^\top G_\nu Q)/\lambda_{\min}(Q^\top G_\mu Q)$. If $u\in S\cap\ker G_\nu$ with $u^\top G_\mu u>0$, the ratio defining $c_S(\mu\|\nu)$ is infinite along $u$.
(d) $c_S\le c_{\mathbb R^P}$, which equals $\max_\ell\varrho_\ell\beta_\ell$ by the argument of Proposition~\ref{prop:kfac}. For isotropic $\mu$ with $\mathrm{tr}A_\mu=\mathrm{tr}A_\nu$, $A_\mu=\frac{\mathrm{tr}A_\nu}{d}I$ and $\varrho_\ell=\lambda_{\max}(A_\nu)\,d/\mathrm{tr}A_\nu$. For a token sampler, $\varrho_1\le\lambda_{\max}(A_\nu)/\lambda_{\min}(A_p)$ and $\lambda_{\min}(A_p)=\mathrm{tr}A_p/(d\,\zeta(A_p))$.
\end{proof}

Theorem~\ref{thm:content} is the closest the theory comes to ``content is unnecessary''. Content cannot be dispensed with for free: by (b), a curated set does beat content-free anchors on the capabilities it represents. But the gain is capped, uniformly over every capability, by one number that can be computed from the anchor sets before training, whereas the loss in the other direction is unbounded (c). By (d), the input-side part of the cap is $d_\ell/r_{\mathrm{eff}}$: a curated set whose representations concentrate on $r_{\mathrm{eff}}$ of $d_\ell$ directions can beat isotropic anchors by at most that factor, and only on capabilities inside those directions, which is also exactly the set outside which its own coefficient is infinite (Proposition~\ref{prop:minimax}).

\textbf{What is proven and what is assumed.} Collecting Theorem~\ref{thm:transfer}, Proposition~\ref{prop:minimax}, Corollary~\ref{cor:crossover} and Theorems~\ref{thm:interference}-\ref{thm:content}: content-free anchors whose curvature is positive definite on $S$ (i) give a finite coefficient on every capability (Theorem~\ref{thm:transfer}, exact); (ii) minimize the worst-case input factor over capabilities of unknown direction (Proposition~\ref{prop:minimax}, under~\ref{as:kfac}); (iii) are beaten by any content on any capability by at most $c_S(\nu\|\mu)$ (Theorem~\ref{thm:content}, exact); and (iv) leave to content only the interference term, which no anchor set can remove without information about $\tau$ or a loss of plasticity (Theorem~\ref{thm:interference}(b)). Three things remain unproven and are the theory's honest limits: that random inputs excite deep-layer representations strongly enough for $\kappa_S$ to be small rather than merely finite; that the output-side factors $\beta_\ell$ in Theorem~\ref{thm:content}(d) are moderate; and that the at-risk capabilities are not concentrated in the target directions $\mathcal R_0$. These are empirical questions, and Section~\ref{sec:results} addresses them.

\begin{table}[hbt!]\centering\small
\caption{\textbf{Which assumptions each result uses.} ``Exact'' means no approximation beyond the stated assumptions. (S1)-(S6) are defined in Section~\ref{sec:transfer}.}
\label{tab:assumptions}
\begin{tabular}{lll}
\toprule
Result & Assumptions & Type\\
\midrule
Props.~\ref{prop:concentration}-\ref{prop:matrix-concentration}, \ref{prop:rel-conc} (concentration) & (S3) & high probability, any $\nu$\\
Prop.~\ref{prop:coreset-covering} (coreset) & (A3)-(A4) & deterministic\\
Thm.~\ref{thm:transfer}(i), (vii); Prop.~\ref{prop:hybrid} & (S2) for (i); none for (vii) & exact, quadratic forms\\
Thm.~\ref{thm:transfer}(ii)-(v) & (S1), (S2); trust radius for (ii)-(iv) & second order\\
Thm.~\ref{thm:transfer}(vi) & (S2), measured $\delta_\nu,\delta_\tau$ & exact, a posteriori\\
Thm.~\ref{thm:transfer}(viii) & (S2), (S5), measured $\delta_\nu,\delta_\tau$ & exact, a posteriori\\
Props.~\ref{prop:kfac}-\ref{prop:minimax} (coverage) & (S4) & exact under (S4)\\
Cor.~\ref{cor:crossover} (crossover) & (S2), (S3) & high probability\\
Thm.~\ref{thm:interference}(a)-(b); Cor.~\ref{cor:blocked} & (S1), (S2) & second order\\
Thm.~\ref{thm:interference}(c) & (S6) & in expectation\\
Thm.~\ref{thm:content}(a)-(c) & (S2) & exact\\
Thm.~\ref{thm:content}(d) & (S4) & exact under (S4)\\
\bottomrule
\end{tabular}
\end{table}

\subsection{Validating the Transfer Theory in a Fully Observable Testbed}
\label{sec:transfer-sim}

On a production VLM, the curvature of every held-out capability cannot be computed exactly. We therefore built a small testbed where it can: every quantity in Sections~\ref{sec:transfer}-\ref{sec:interference} is computed exactly, and we know in advance which capabilities exist and which ones a curated anchor pool contains. We use it to test the theory's predictions about \namel{} (random anchors) and \realname{} (curated anchors), not to estimate their performance on Qwen3-VL.

\textbf{The model and the task.} A fixed teacher network (2-layer, tanh) defines a 10-way classification function on $\mathbb R^{64}$. A student policy $\pi_\theta(x)=\mathrm{softmax}(W_2\tanh(W_1x+b_1)+b_2)$ ($P=4{,}810$ parameters) is pretrained by distillation to imitate the teacher, then post-trained with GRPO (group size 8, Adam, $10^3$ steps, decoupled weight decay) to learn a new labeling rule on a single target domain. Anchoring uses Eq.~\ref{eq:anchor} unchanged: linear-kernel MMD with a fixed embedding Gram matrix and varentropy weights.

\textbf{Capabilities.} A \emph{capability} is a held-out skill the pretrained model has and RL post-training should not destroy, playing the role of one benchmark in our evaluation suite. We model each of the ten capabilities as a distribution of inputs lying in its own linear subspace $U_m\subset\mathbb R^{64}$; the student's accuracy on fresh inputs from $U_m$, before and after RL, measures how much of that capability is retained. The target domain is likewise a subspace $U_0$. How much a capability shares with the target is set by an overlap $\alpha_m\in[0,1]$: we construct $U_m=\mathrm{orth}\big(\alpha_mU_0R_m+(1-\alpha_m)V_m\big)$ with $V_m\perp U_0$, so $\alpha_m=0$ means the capability's inputs share no direction with the target's.

\textbf{Covered vs.\ uncovered capabilities.} The curated anchor pool contains real inputs from only six of the ten capabilities, which we call \emph{covered}; the other four are \emph{uncovered}. This mirrors the practical situation that motivates \namel{}: a curated pool is built from the capabilities a practitioner anticipated (in our experiments, five capability dimensions), while RL can damage capabilities nobody anticipated. Random anchors involve no such choice and treat all ten capabilities alike.

\textbf{Spread vs.\ concentrated capabilities.} We use two settings that differ in how many input directions each capability occupies and how strongly it overlaps the target:
\begin{itemize}
\item \emph{Spread} (main setting). Each capability varies along 16 of the 64 input directions; the target task varies along 4; overlap with the target is small ($\alpha_m\le0.2$). This matches the regime Theorem~\ref{thm:interference}(c) identifies for large models: a narrow, low-rank target update against broad pretrained capabilities, and inputs (like images and text) that vary along many directions rather than a thin slice.
\item \emph{Concentrated} (stress test). Each capability varies along only 4 of the 64 directions, and some overlap strongly with the target ($\alpha_m$ up to $0.5$). Curated anchors gain the most over isotropic ones here, up to a factor $d/r=16$ on the input side (Theorem~\ref{thm:content}(d)), and target-adjacent capabilities create interference (Theorem~\ref{thm:interference}(b)). We use it only to probe the predicted boundary.
\end{itemize}

\textbf{Anchor sets and measurements.} \realname{} uses greedy $k$-center over the curated pool, as in Algorithm~\ref{alg:coreset}; \namel{} uses isotropic random inputs; we also report a 50/50 hybrid. $\kappa_S$ is evaluated at $\theta^*$, on the 40-dimensional span $S$ of the training displacements of all runs. All results are over 5 independent seeds (each redraws the teacher, capabilities, pretraining and RL run).

\begin{figure}[t]\centering
\includegraphics[width=\linewidth]{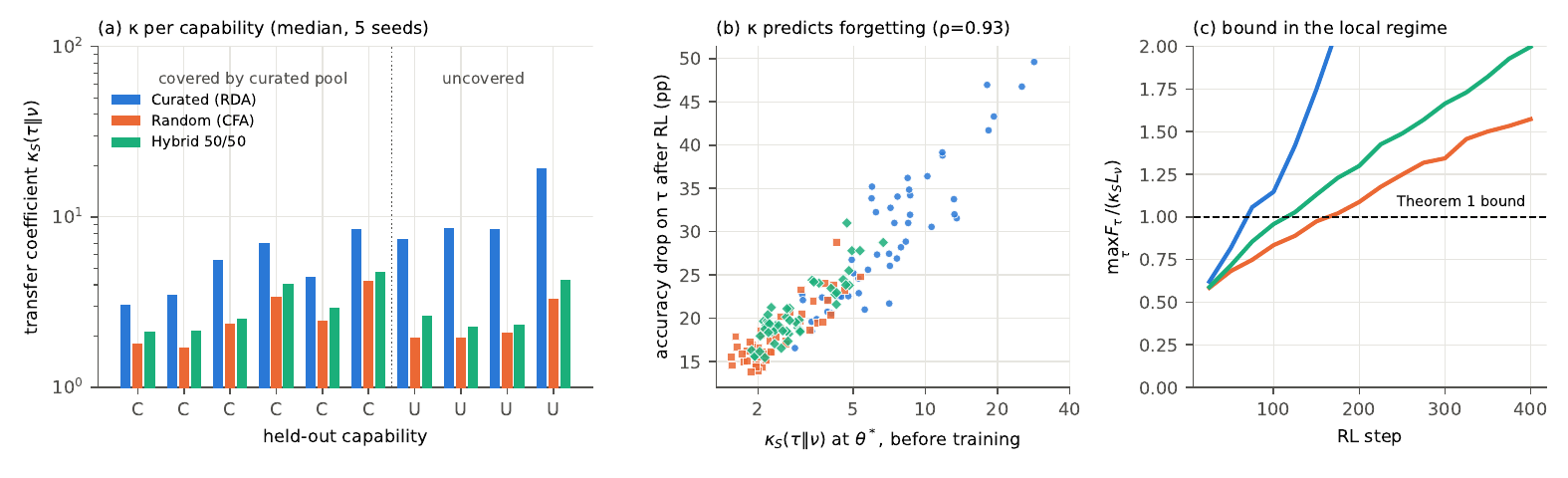}
\caption{\textbf{Theorem~\ref{thm:transfer} in the testbed (spread setting).} (a) Transfer coefficient per held-out capability; ``C'' marks the six capabilities in the curated pool and ``U'' the four outside it. \namel{} has the smallest $\kappa$ on every capability; \realname's $\kappa$ is largest on uncovered ones. (b) $\kappa$, evaluated at $\theta^*$ on the span of the training updates, ranks each capability's accuracy drop after RL (Spearman $\rho=0.93$, $n=150$). (c) The raw ratio $F_\tau/L_\nu$ stays below $\kappa_S$ early in training and loosens past the trust radius (Proposition~\ref{prop:trust-radius}); the remainder-corrected bound of Theorem~\ref{thm:transfer}(vi) holds at every checkpoint.}
\label{fig:sim-transfer}
\end{figure}

\textbf{Transfer (Theorem~\ref{thm:transfer}).} $\log\kappa_S$, evaluated at $\theta^*$, rank-correlates with each capability's final accuracy drop at $\rho=0.93$ (Figure~\ref{fig:sim-transfer}b). Because $S$ is estimated from the training runs themselves, this validates Theorem~\ref{thm:transfer} on the subspace where its assumption $\Delta\theta\in S$ holds, rather than providing an a-priori forecast. Estimating $S$ before training instead, from the leading directions of the target task's Gauss-Newton matrix or of GRPO gradients at $\theta^*$, gives a weaker but still positive rank correlation ($\rho=0.64$ and $0.63$). For \namel{} and the hybrid, the nonlinear bound $F_\tau(\theta_t)\le\kappa_S(\tau\|\nu)L_\nu(\theta_t)$ held for every capability throughout the first 100 RL steps; for \realname{} it held at $86\%$ of those checkpoints. Later in training the raw ratio $F_\tau/L_\nu$ exceeds $\kappa_S$ (by up to $12\times$ for \realname{}), as expected once the losses leave their quadratic model (Proposition~\ref{prop:trust-radius}). It does so only through the remainders: with the measured $\delta_\nu,\delta_\tau$, the bound of Theorem~\ref{thm:transfer}(vi) held at every checkpoint of all 1000 steps, for every capability, anchor set and seed ($8{,}000$ pairs; largest ratio of $F_\tau$ to the bound $0.91$), and the quadratic inequality of Theorem~\ref{thm:transfer}(i) held at every checkpoint although $3$-$4\%$ of each update's norm lies outside $S$. On Qwen3-VL-2B the anchor-side quadratic model stays within $3$-$7\%$ of the exact loss for the full run (Table~\ref{tab:trust-radius-data}).

\textbf{Coverage (Propositions~\ref{prop:minimax},~\ref{prop:hybrid}).} \namel{} attains $\kappa=1.7$-$4.2$ on covered capabilities and $2.0$-$3.3$ on uncovered ones; \realname{} attains $3.1$-$8.5$ and $7.4$-$19.3$ (Figure~\ref{fig:sim-transfer}a). \namel's coefficients are both smaller and more uniform, as Proposition~\ref{prop:minimax} predicts for isotropic anchors. The hybrid bound of Proposition~\ref{prop:hybrid} held in all 50 (seed, capability) pairs.

\begin{figure}[t]\centering
\includegraphics[width=\linewidth]{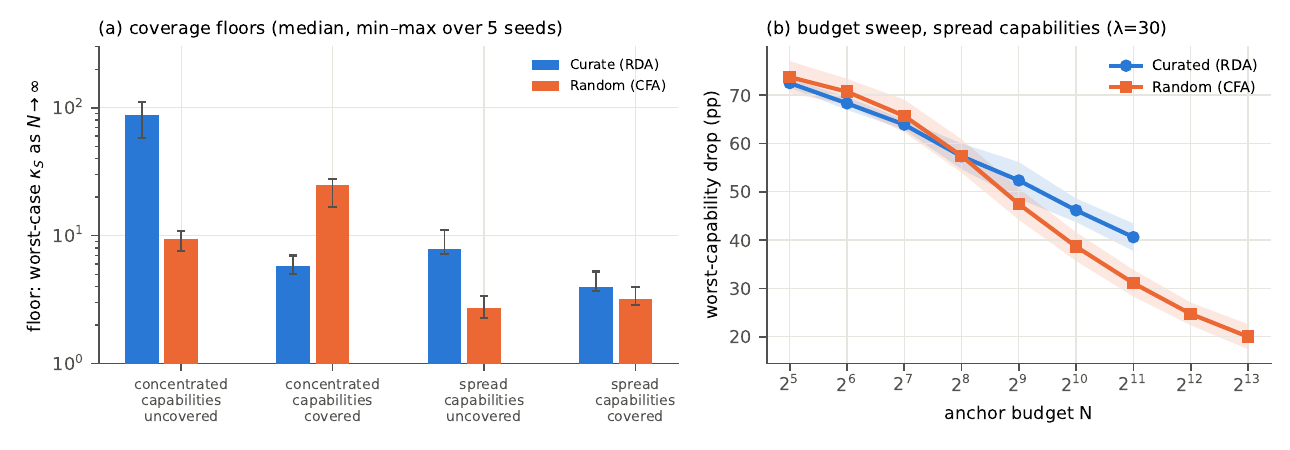}
\caption{\textbf{Coverage floors and the crossover (Corollary~\ref{cor:crossover}).} (a) Floor of each anchor type: its worst-case $\kappa_S$ with unlimited anchors. For \realname{} this limit is the \emph{entire} curated pool (six of the ten capabilities), since a coreset can never go beyond its pool; for \namel{} it is $N{=}16{,}384$ random anchors. Bars group capabilities by setting (concentrated or spread) and by whether the curated pool contains them (uncovered or covered); median and min-max over 5 seeds. (b) Worst-capability accuracy drop vs.\ anchor budget $N$ at fixed $\lambda$, spread setting (mean $\pm$ sd, 5 seeds).}
\label{fig:sim-floor}
\end{figure}

\textbf{Coverage floors (Corollary~\ref{cor:crossover}).} Earlier versions of this paper argued that random anchors have a lower ``bias floor'' relative to an unknown protective curvature. Corollary~\ref{cor:crossover} replaces that assumption with a quantity we can measure: each anchor type's \emph{floor}, its worst-case $\kappa_S$ as $N\to\infty$. For \realname{} the limit is the whole curated pool, because no amount of sampling goes beyond it; for \namel{} it is a very large random draw. Figure~\ref{fig:sim-floor}a reports both.
\begin{itemize}
\item \emph{Uncovered capabilities.} \namel's floor is far lower: $2.7$ vs.\ $7.9$ in the spread setting and $9.4$ vs.\ $88$ in the concentrated one, lower in every seed of both settings. This is the range-inclusion effect: the curated pool cannot see directions that only uncovered capabilities use, however many points are drawn from it, whereas random anchors excite every direction.
\item \emph{Covered capabilities, spread setting.} The two floors are close ($3.2$ for \namel{} vs.\ $3.9$ for \realname{}): random anchors match a curated pool even on the capabilities that pool was built for.
\item \emph{Covered capabilities, concentrated setting.} Here \realname{} has the lower floor ($5.8$ vs.\ $24.7$). These capabilities occupy a thin slice of input space and overlap strongly with the target, which is exactly where Proposition~\ref{prop:minimax} (coverage cost $d/r$) and Theorem~\ref{thm:interference}(b) (interference) predict an advantage for anchors drawn from the capability itself.
\end{itemize}
\namel's lower floor therefore holds precisely when the curated pool misses some at-risk capability, which is the premise of protecting \emph{general} capability rather than a few anticipated domains.

\textbf{Budget (Corollary~\ref{cor:crossover}, Proposition~\ref{prop:rel-conc}).} \namel{} approaches its floor quickly: its worst uncovered-capability $\kappa_S$ is $4.0$ at $N{=}512$ and $3.2$ at $N{=}2048$, whereas \realname{} coresets remain well above their floor at practical budgets ($20.0$ and $7.2$). Measured worst-capability forgetting is the same for both at small budgets and separates from $N{=}512$ onward, with \namel{} ahead: $38.6$ vs.\ $46.1$ pp at $N{=}1024$ and $31.1$ vs.\ $40.6$ pp at $N{=}2048$ (Figure~\ref{fig:sim-floor}b). Because every anchor in the testbed is a single input with no padding, this comparison is at exactly matched token counts: the crossover is not an artifact of random anchors contributing more scored tokens. \namel{} also forgets less than \realname{} given twice as many inputs ($38.6$ pp at $N{=}1024$ vs.\ $40.6$ pp at $N{=}2048$). For \namel{}, $\|\widehat G_N-\widehat G_{2N}\|_{\mathrm{op}}/\|\widehat G_{2N}\|_{\mathrm{op}}$ decays with log-log slope $-0.49$, the $N^{-1/2}$ rate of Proposition~\ref{prop:matrix-concentration}; curated subsets concentrate at a similar rate, which is why concentration alone cannot explain the crossover and $\kappa_S$ is needed.

\begin{figure}[t]\centering
\includegraphics[width=0.55\linewidth]{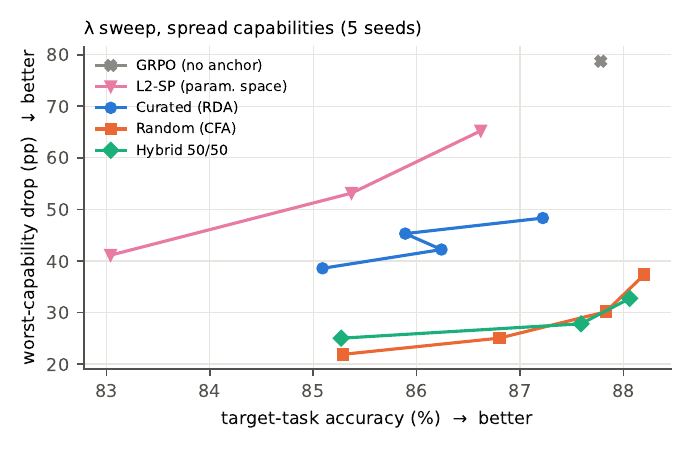}
\caption{Target accuracy vs.\ worst-capability accuracy drop across a $\lambda$ sweep (spread setting, 5 seeds).}
\label{fig:sim-pareto}
\end{figure}

\textbf{Retention vs.\ plasticity.} Across a $\lambda$ sweep (Figure~\ref{fig:sim-pareto}), \namel{} reaches a worst-capability drop of $22$-$37$ pp at $81$-$84\%$ target accuracy, against $39$-$48$ pp at $85$-$87\%$ for \realname{} and $41$-$65$ pp at $83$-$87\%$ for parameter-space L2-SP. The hybrid lies between \namel{} and \realname{} in target accuracy and close to \namel{} in forgetting. Because isotropic anchors also penalize directions the target uses (Theorem~\ref{thm:interference}(b)), \namel{} gives up 1-3 points of target accuracy.

The testbed's absolute numbers are not a model of Qwen3-VL. What transfers are the diagnostics. The layer-wise coverage $\varrho_\ell$ and the isotropy defect $\zeta$ need only the base model and the anchors; $\kappa_S$, the coverage floors and the exchange factor $c_S(\nu\|\mu)$ of Theorem~\ref{thm:content} additionally need the update subspace $S$, but never the held-out capability's outcome.

\subsection{EWC vs. \namel{} at a Glance}
\label{app:ewc-table}

Table~\ref{tab:ewc-cfa} summarizes the argument of this appendix in full.

\begin{table}[hbt!]
\caption{Parameter-space EWC vs.\ function-space \namel, per
Section~\ref{sec:theory}.}
\label{tab:ewc-cfa}
\begin{center}
\small
\resizebox{\textwidth}{!}{%
\begin{tabular}{lll}
\toprule
& \textbf{EWC / Unweighted-KL} ($F(\theta^*)$) & \textbf{\namel} ($G_{\text{emb}}$) \\
\midrule
Simplex metric $M$ & $\mathrm{diag}(1/\pi^*)$ (Fisher-Rao) & $K=\frac1d EE^\top$ (embedding Gram matrix) \\
Metric source & Reference distribution itself & Fixed, geometric (embedding table) \\
Position weighting & Uniform & Varentropy $\tilde w_t$ (Cor.~\ref{cor:cfa-is-generalized-ewc}) \\
Sensitivity & Any probability shift, weighted by rarity & Shifts altering semantic meaning \\
Evaluation & Quadratic approx., fixed at $\theta^*$ & Exact, re-evaluated every step \\
Valid while & $\|\Delta\theta\|\lesssim\rho^*$ (Prop.~\ref{prop:trust-radius}) & Unconditionally (no truncation) \\
Memory & $\mathcal{O}(P)$ diag.\ approx., $>32$\,GB at 8B params & $\mathcal{O}(1)$ parameter buffers  \\
&&
($\mathcal{O}(NTd)$ cached features, $<22$\,GB at 8B) \\
\bottomrule
\end{tabular}}
\end{center}
\end{table}

\section{Fisher (Varentropy) Weighting}
\label{app:fisher-weighting}

The varentropy weight $w_t=\mathrm{Var}_{\pi^*}[\log\pi^*(v\mid c_{<t})]$
is maximized at intermediate entropy: near-uniform distributions
($\pi(v)\approx1/V$) and near-deterministic distributions
($\pi(v^*)\approx1$) both give $\mathrm{Var}[\log\pi]\approx0$, while
distributions with mass spread over a small, structured candidate set
maximize $w_t$. Corollary~\ref{cor:cfa-is-generalized-ewc} shows that this
reweighting enters the induced curvature matrix $G_{\text{emb}}$ as a
non-negative per-position scaling, so it redistributes curvature mass
across positions without affecting positive semi-definiteness or any of the
concentration results (Remark~\ref{rem:weighting}).

\subsection{Empirical Varentropy Profile across Sequence Positions}
\label{app:varentropy-profile}

To evaluate the inverted-U profile predicted by the varentropy formulation, we evaluate the token-level varentropy $w_t = \mathrm{Var}_{\pi^*}[\log \pi^*(v \mid c_{<t})]$ alongside the token entropy $H(\pi^*_t) = -\sum_v \pi^*(v \mid c_{<t}) \log \pi^*(v \mid c_{<t})$ over rollout trajectories from the base Qwen3-VL-2B model on the Navigation domain.

\begin{figure}[hbt!]
\centering
\includegraphics[width=0.95\linewidth]{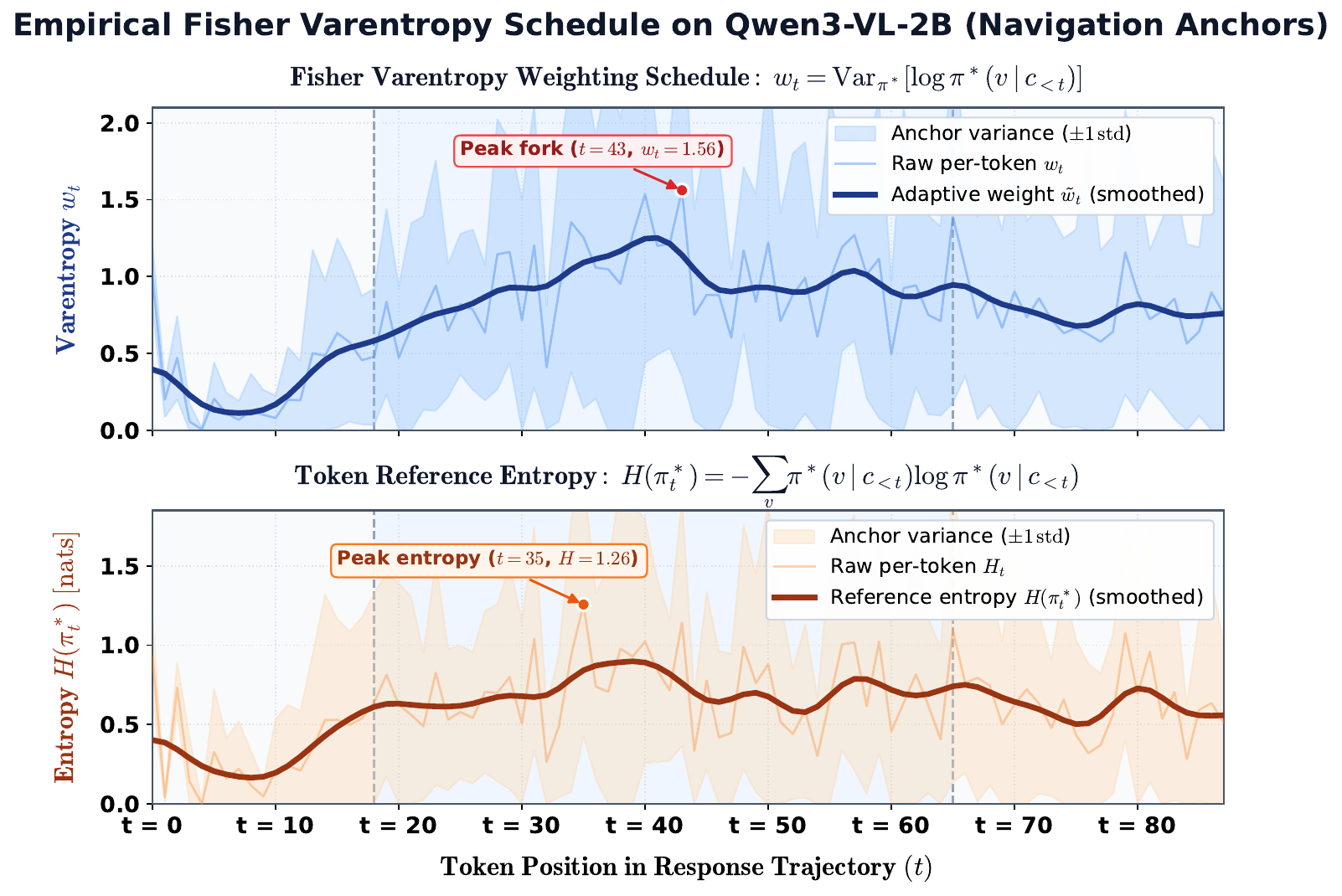}
\caption{\textbf{Empirical Fisher Varentropy Profile.} Token-level varentropy $w_t = \mathrm{Var}_{\pi^*}[\log\pi^*(v\mid c_{<t})]$ (top) and token entropy $H(\pi^*_t)$ (bottom) evaluated across sequence positions on Navigation anchors using base Qwen3-VL-2B. Shaded bands represent $\pm 1$ standard deviation across anchor sequences. The profile validates the theoretical inverted-U dynamic, maximizing curvature regularization at intermediate reasoning decision points.}
\label{fig:varentropy_profile}
\end{figure}

As illustrated in Figure~\ref{fig:varentropy_profile}, the empirical profile reflects the structural stages of multi-modal reasoning:
\begin{itemize}
\item \textbf{Early context positions ($t \in [0, T/4]$):} Standard prompt formatting and formulaic reasoning prefixes exhibit low varentropy, reflecting near-deterministic continuation where single tokens dominate the reference distribution.
\item \textbf{Intermediate reasoning forks ($t \in [T/4, 3T/4]$):} As the model evaluates visual spatial relationships, landmarks, and navigation path coordinates, reference probability mass spreads across a competitive candidate set of domain tokens, elevating varentropy to its peak.
\item \textbf{Terminal answer tokens ($t \in [3T/4, T]$):} Final answer formatting tokens collapse back toward high certainty, returning varentropy toward baseline levels.
\end{itemize}
This dynamic confirms that varentropy acts as an automatic, uncertainty-aware weighting schedule: it focuses regularization on the critical decision points where semantic divergence is most catastrophic while discounting predictable formatting tokens. As shown in Figure~\ref{fig:varentropy_profile}, the empirical profile across token positions confirms this inverted-U trajectory in practice.

\section{Anchor Set Construction: Full Details}
\label{app:anchors}

\textbf{Real anchor pool.}  The candidate pool $\mathcal{U}$ spans five
capability dimensions, each paired with a target skill, a source
\emph{training} split (never an evaluation split), and a generic fallback
probe used when an anchor's own question is unavailable or uninformative
(Table~\ref{tab:anchor-pool}).

\begin{table}[hbt!]
\caption{Real anchor pool: five capability dimensions, each drawn from a
training split with no overlap with any evaluation benchmark.}
\label{tab:anchor-pool}
\begin{center}
\small
\begin{tabular}{p{2.1cm}p{3.0cm}p{2.3cm}p{4.3cm}}
\toprule
Dimension & Target skill & Source split & Generic fallback probe \\
\midrule
STEM \& Math & Geometry, scientific plots, physics diagrams & ScienceQA (train) & ``Analyze the mathematical, geometric, or scientific problem in this figure step by step.'' \\
\addlinespace
General VQA & Real-world scene layout, commonsense perception & GQA & ``Provide a dense, comprehensive description of the scene layout, objects, and spatial relationships.'' \\
\addlinespace
Hallucination & Object presence verification, fine-grained detail & MS-COCO (train split) & ``Carefully examine the image: list all verified objects present and confirm what objects are not present.'' \\
\addlinespace
Chart \& Document & OCR, tables, receipts, axis reading & ChartQA (train), TextVQA (train) & ``Transcribe all visible text, numbers, headings, and table cells in this document accurately.'' \\
\addlinespace
Grounding \& Counting & Spatial localization, instance enumeration & GQA & ``List each distinct type of object visible in this image and state the exact count of each.'' \\
\bottomrule
\end{tabular}
\end{center}
\end{table}

\textbf{Coreset selection.}  Algorithm~\ref{alg:coreset} gives the full
greedy $k$-center procedure used for real-anchor selection in
Section~\ref{sec:anchor-construction}. Note that this procedure is
deterministic given its seed and is explicitly \emph{not} an i.i.d.\ sample;
Proposition~\ref{prop:coreset-covering} gives the corresponding guarantee.

\begin{algorithm}[hbt!]
\caption{Greedy $k$-center coreset selection for anchor construction}
\label{alg:coreset}
\begin{algorithmic}[1]
\STATE \textbf{Input:} candidate pool $\mathcal{U}$ (images from a single
   capability dimension), target budget $N$.
\STATE \textbf{Feature extraction:} for each $x \in \mathcal{U}$, extract
   a visual embedding $\psi(x) \in \mathbb{R}^{d'}$ using the frozen
   vision encoder of Qwen3-VL-2B.  Concretely, each image is passed
   through the ViT backbone; the spatial patch tokens are
   mean-pooled to yield a single $d'$-dimensional vector, which is
   then $\ell_2$-normalised:
   \[
     \psi(x) = \frac{\mathrm{ViT}_{\mathrm{Qwen3\text{-}VL\text{-}2B}}(x)}
                    {\|\mathrm{ViT}_{\mathrm{Qwen3\text{-}VL\text{-}2B}}(x)\|_2}.
   \]
   The encoder weights are kept frozen throughout.
\STATE \textbf{Initialise:} draw the first anchor $x_1 \in \mathcal{U}$
   uniformly at random (seed $= 42$); set
   $\mathcal{S}_1 \leftarrow \{x_1\}$.
\STATE \textbf{Distance tracking:} maintain, for every remaining candidate
   $x \in \mathcal{U} \setminus \mathcal{S}_j$, its distance to the
   current coreset:
   \[
     d\bigl(x,\,\mathcal{S}_j\bigr)
     \;=\; \min_{s \in \mathcal{S}_j} \|\psi(x) - \psi(s)\|_2.
   \]
   After adding $x_{j+1}$ at each step, only the column
   corresponding to $x_{j+1}$ needs to be recomputed, so the update
   is $\mathcal{O}(|\mathcal{U}|\cdot d')$ per iteration.
\STATE \textbf{Greedy expansion:} select the candidate farthest from the
   current coreset:
   \[
     x_{j+1} = \arg\max_{x \in \mathcal{U} \setminus \mathcal{S}_j}
               d\bigl(x,\,\mathcal{S}_j\bigr),
     \qquad
     \mathcal{S}_{j+1} \leftarrow \mathcal{S}_j \cup \{x_{j+1}\}.
   \]
\STATE Repeat Steps 4-5 until $|\mathcal{S}| = N$.
\STATE \textbf{Output:} ordered coreset $\mathcal{S}_N \subseteq \mathcal{U}$
   of $N$ maximally diverse anchors.  Smaller budgets
   $N' < N$ are obtained as the prefix $\mathcal{S}_{N'}$.
\end{algorithmic}
\end{algorithm}

\textbf{Capacity settings and nested subsets.}  The algorithm is run once
per capability dimension to the maximum per-dimension budget, producing a
ranked list. Because every smaller anchor set is a \emph{prefix} of this
list, the four budgets used in Section~\ref{sec:discussion-ablations} form a
strictly nested chain,
\[
\mathcal{S}_{320}\subset\mathcal{S}_{640}\subset
\mathcal{S}_{1280}\subset\mathcal{S}_{2560},
\]
corresponding to $64$, $128$, $256$, and $512$ anchors per dimension across
the five dimensions. Nesting is guaranteed by construction rather than by
re-running the selection, so the budget ablation varies $N$ alone and holds
the selection procedure fixed.

\textbf{Sequence length and padding masks for real anchors.} Context sequences $c^{(i)} = (I_i, x^{(i)})$ are constructed by concatenating the candidate image with question prompts and ground-truth answer prefixes from the source datasets. To accommodate variable prompt lengths up to the training context window $T=2048$, sequences are right-padded to $T=2048$. In the anchoring objective (Eq.~\ref{eq:anchor}), an attention mask $m_t^{(i)} \in \{0, 1\}$ is applied across positions, normalizing the MMD loss by the effective valid token count $\sum_{t=1}^T m_t^{(i)}$ rather than unmasked padding tokens.

\textbf{Scored tokens per anchor and matched-token comparison.}
\label{app:matched-tokens}
Because padding is masked, a real anchor contributes only its valid tokens, while a content-free anchor contributes all $T=2048$. Over the full pool of 2{,}560 real anchors, a real anchor contributes $486$ scored tokens on average, about $4.2\times$ fewer than a content-free anchor, so at equal $N$ the comparisons in Section~\ref{sec:results} give content-free anchors a larger token budget. The budget ablation (Table~\ref{tab:anchor_budget_ablation_full}) therefore contains approximately token-matched pairs, \namel{} at $N$ against \realname{} at $4N$ (Table~\ref{tab:matched-tokens}, Appendix~\ref{app:matched-tokens-table}). At about $0.65$M scored tokens, curated anchors are as good or better (worst-1 tied, worst-3 favoring \realname{}); at about $1.3$M, content-free anchors are better on worst-1, worst-3, and target accuracy. Doubling \realname{}'s tokens from $0.62$M to $1.24$M leaves its worst-1 essentially unchanged, while \namel{} continues to improve. The crossover therefore survives at matched token budgets; in the testbed of Appendix~\ref{sec:transfer-sim}, where token counts are matched exactly, it appears cleanly as well.

\textbf{Reproducibility.}  All random choices (initial seed point and
candidate-pool subsampling) are governed by a single global seed
(\texttt{seed\,=\,42}).  Because the greedy expansion is fully
deterministic given the seed and the feature representation, the resulting
coreset is reproducible without requiring a seed sensitivity analysis.

\textbf{Content-free anchors.} Content-free anchors provide functional regularization without authentic prior-domain data. Token sequences are sampled independently and uniformly from the base vocabulary ($\mathcal{V}_{\text{base}} = \{0, \ldots, 151642\}$), strictly excluding special tokens (e.g., \texttt{<|im\_start|>}, \texttt{<|im\_end|>}, padding, and tool-call delimiters) introduced during instruction tuning.
The sequence length is fixed at $T=2048$ tokens across all anchors, identical to the training context window.

Unlike the greedy coreset construction above, this synthetic generation is an exact i.i.d.\ draw from a stationary distribution, enabling the finite-sample concentration guarantees of Propositions~\ref{prop:concentration} and~\ref{prop:matrix-concentration} to apply directly without covering-set approximations.

\textbf{Training-time algorithm.} Algorithm~\ref{alg:cfa} gives the \namel{} training step referenced in the Reproducibility Statement and in Section~\ref{sec:anchor-objective}: a fixed pool of $N$ content-free
anchors is drawn once before training, the frozen reference is evaluated on
it once and its expected embeddings and varentropy weights are cached, and
each step draws a random mini-batch of size $B \ll N$ from the pool to
compute the Fisher-weighted anchoring loss against the cached statistics.
The reference is never called during training.

\begin{algorithm}[hbt!]
\caption{\namel{}: Fisher-weighted anchor regularization with cached reference statistics}
\label{alg:cfa}
\begin{algorithmic}[1]
\Require frozen reference $\pi_{\mathrm{ref}}$, embedding matrix $E\in\mathbb{R}^{V\times d}$, anchor pool size $N$, per-step
  anchor batch size $B$ ($B \ll N$), regularizer weight $\lambda$, anchor
  sequence length $T$, base vocabulary $\mathcal{V}_{\text{base}}$
\Statex \textbf{One-time setup, before training:}
\State sample $x_1, \dots, x_N \overset{\text{iid}}{\sim} \mathrm{Uniform}(\mathcal{V}_{\text{base}})^{T}$
  \Comment{pure random token sequences}
\For{each anchor $x_i$} \Comment{single no-grad, teacher-forced pass of $\pi_{\mathrm{ref}}$}
  \For{each position $t = 1, \dots, T$}
    \State $\mu^{*}_{i,t} \gets \tfrac{1}{\sqrt d}\,E^\top \pi_{\mathrm{ref}}(\cdot \mid x_{i,<t})$ \Comment{expected embedding, $d$-dim}
    \State $w_{i,t} \gets \mathrm{Var}_{v\sim\pi_{\mathrm{ref}}}\!\big[\log \pi_{\mathrm{ref}}(v \mid x_{i,<t})\big]$ \Comment{varentropy}
  \EndFor
  \State $\tilde w_{i,t} \gets T\, w_{i,t} / \sum_{s} w_{i,s}$ for all $t$ \Comment{normalize: $\sum_t \tilde w_{i,t}=T$}
\EndFor
\State cache $\mathcal{C} \gets \{(x_i, \mu^{*}_{i,1:T}, \tilde w_{i,1:T})\}_{i=1}^{N}$; release $\pi_{\mathrm{ref}}$ from GPU memory
  \Comment{$\mathcal{O}(NTd)$ storage}
\Statex
\Statex \textbf{Every training step, in addition to the standard RL update:}
\State sample $\mathcal{B} \subset \{1,\dots,N\}$, $|\mathcal{B}| = B$, uniformly without replacement
\State $\text{logits}_{\theta} \gets \pi_\theta(\{x_i\}_{i\in\mathcal{B}})$ \Comment{one teacher-forced forward pass of the policy}
\State $\ell_{\mathrm{anchor}} \gets \frac{1}{BT}\sum_{i\in\mathcal{B}}\sum_{t=1}^{T} \tilde w_{i,t}\, \big\lVert \tfrac{1}{\sqrt d}E^\top \pi_\theta(\cdot\mid x_{i,<t}) - \mu^{*}_{i,t} \big\rVert_2^2$
  \Comment{Eq.~\ref{eq:anchor}, using cached $\mu^{*}$, $\tilde w$}
\State $\mathcal{L}(\theta) \gets \mathcal{L}_{\mathrm{RL}}(\theta) + \lambda \cdot \ell_{\mathrm{anchor}}$
\State update $\theta$ by descending $\nabla_\theta \mathcal{L}(\theta)$ \Comment{gradient flows only through $\pi_\theta$}
\end{algorithmic}
\end{algorithm}

\subsection{Sensitivity to the Anchor Token Distribution}
\label{app:anchor-sampling}

Content-free anchors are defined by a single design choice: the distribution from which their tokens are drawn. Our default is uniform sampling over the base vocabulary. Here we ask whether that choice matters, by holding everything else fixed and changing only the token sampler.

\paragraph{Samplers.} Every variant draws $N{=}1280$ i.i.d.\ sequences of $T{=}2048$ tokens, excluding special tokens exactly as for the default anchors. They differ only in the per-token distribution $p$:
\begin{itemize}
\item \textbf{Uniform, full vocabulary (default).} $p_v = 1/|\mathcal V_{\text{base}}|$.
\item \textbf{Filtered vocabularies (top-1k, top-16k).} Uniform over the $1{,}024$ or $16{,}384$ most frequent tokens, with zero mass elsewhere. Token frequencies are counted on WikiText-103 ($\approx$124M tokens).
\item \textbf{Non-uniform sampling (unigram, $\tau{=}1$).} $p_v \propto f_v$, the natural unigram frequency of token $v$ on the same corpus.
\item \textbf{Sampling temperature (unigram, $\tau{=}2$).} $p_v \propto f_v^{1/\tau}$, which flattens the unigram distribution and places it between natural frequency ($\tau{=}1$) and uniform ($\tau{\to}\infty$).
\end{itemize}

\paragraph{What each sampler changes in theory.} Proposition~\ref{prop:minimax} shows that, when the at-risk capability is unknown, the worst-case coverage at a layer is governed by $\lambda_{\min}$ of the second moment of the representations the anchors induce, and is best for isotropic anchors. At the embedding layer this moment is $A_p=\sum_v p_v e_v e_v^\top$, which can be computed from the embedding table before training. Uniform sampling is the closest to isotropic of the samplers tested: $A_p$ is full rank and spreads energy evenly across embedding directions. Unigram sampling and top-16k keep $A_p$ full rank but concentrate energy on frequent tokens, lowering $\lambda_{\min}$; for natural frequencies ($\tau{=}1$) it is roughly $0.11\times$ that of uniform sampling. Top-1k spans only $1{,}024$ of the $d{=}2{,}048$ embedding directions of Qwen3-VL-2B, so $A_p$ is singular and the worst-case coefficient at the first layer is infinite.

\paragraph{Protocol.} We post-train Qwen3-VL-2B on MedBookVQA with GSPO and \namel{}, with $\lambda{=}0.01$ and learning rate $10^{-6}$, for 100 steps, a short budget intended as a sensitivity check rather than a full run. All runs use identical settings and seeds apart from the sampler, and are trained on L40 GPUs. We compare against unregularized GSPO under the same budget. We evaluate two quantities:
\begin{itemize}
\item \textbf{Target accuracy} on MedBookVQA, to confirm that every variant still learns the task;
\item \textbf{CountBenchQA accuracy}, our measure of retention. CountBenchQA is the benchmark most sensitive to forgetting in our suite (it collapses by up to $67\%$ under unregularized RL on this domain; Table~\ref{tab:med-2b-full}), and it probes fine-grained visual grounding, which text-token anchors cannot protect directly and so constitutes a demanding test of anchor coverage.
\end{itemize}
With roughly 500 items, a single CountBenchQA score has a standard error of about $3.5$ points, so the difference between two single runs has a standard error of about $5$ points, and gaps below roughly $10$ points are within noise.

\begin{table}[hbt!]
\centering
\caption{\textbf{Sensitivity to the anchor token distribution} (MedBookVQA, Qwen3-VL-2B, GSPO, 100 steps, $N{=}1280$ anchors of 2048 tokens, $\lambda{=}0.01$). Only the token sampler changes between \namel{} runs. $\Delta$ is the change in CountBenchQA accuracy relative to the base model, in points.}
\label{tab:anchor-sampling}
\small
\resizebox{\textwidth}{!}{%
\begin{tabular}{llccc}
\toprule
Method & Anchor sampler & MedBookVQA (target) & CountBenchQA & $\Delta$ CBQA \\
\midrule
Base & - & 49.4 & 82.5\,$\pm$\,3.4 & - \\
GSPO & none & 64.3 & 68.3\,$\pm$\,4.2 & $-14.1$ \\
\midrule
\namel{} & \textbf{uniform, full vocabulary (default)} & 62.3 & \textbf{80.8\,$\pm$\,3.6} & $\mathbf{-1.7}$ \\
\namel{} & top-1k vocabulary & 62.8 & 80.2\,$\pm$\,3.5 & $-2.2$ \\
\namel{} & unigram, $\tau{=}1$ & 64.8 & 78.6\,$\pm$\,3.2 & $-3.9$ \\
\namel{} & top-16k vocabulary & 57.7 & 71.3\,$\pm$\,4.1 & $-11.2$ \\
\namel{} & unigram, $\tau{=}2$ & 57.7 & 60.7\,$\pm$\,4.4 & $-21.8$ \\
\bottomrule
\end{tabular}}
\end{table}

\paragraph{Results.} Table~\ref{tab:anchor-sampling} reports the outcome.
\begin{enumerate}
\item \textbf{Uniform sampling gives the best retention.} Within 100 steps, unregularized GSPO loses $14.1$ points on CountBenchQA; \namel{} with uniform anchors loses only $1.7$, while still gaining $12.9$ points on the target task. No alternative sampler improves on it.
\item \textbf{Some alternatives come close.} Top-1k and natural-frequency sampling ($\tau{=}1$) are within $0.5$ and $2.1$ points of uniform sampling on CountBenchQA, inside single-run noise, and match or exceed its target accuracy.
\item \textbf{Others degrade substantially.} Top-16k and $\tau{=}2$ retain $9.5$ and $20.1$ points less than uniform sampling, and also learn less of the target task ($57.7$ vs.\ $62.3$). Unigram sampling at $\tau{=}2$ retains even less than unregularized GSPO.
\end{enumerate}

\paragraph{Interpretation.} The results support uniform sampling as the principled default, for the reason Proposition~\ref{prop:minimax} gives: it is the only sampler tested that keeps the worst-case coverage guarantee, up to the isotropy defect $\zeta(E^\top E/V)$. Restricting or reweighting the vocabulary gives up part of that guarantee, and once it is gone the outcome is no longer predictable from the sampler. Two observations make this concrete. First, the results are \emph{not monotone} in the sampling parameter: top-16k contains top-1k yet retains less than both top-1k and uniform sampling, and $\tau{=}2$ lies between $\tau{=}1$ and uniform yet retains less than both. Any account based purely on how much of the vocabulary a sampler covers predicts the opposite ordering. Second, the failing samplers lose on both axes at once, learning less and forgetting more. A penalty that was simply stronger would reduce forgetting at the cost of learning, and a weaker one would do the reverse; losing on both instead suggests that these anchors place their constraint in directions that matter for neither the target task nor the held-out capability.

Top-1k's good result does not contradict Proposition~\ref{prop:minimax}. That proposition bounds the \emph{worst case} over unknown capabilities at a single layer. After attention mixes 2048-token contexts, even a 1k-token vocabulary induces varied representations at deeper layers, and CountBenchQA evidently does not rely on the embedding directions top-1k leaves uncovered. Nothing guarantees the same for other capabilities, which is precisely the risk a rank-deficient sampler takes on.

\paragraph{Limitations of this ablation.} These are single-seed, 100-step runs evaluated on one domain and one retention benchmark; they establish that the sampler matters and that uniform sampling is a safe default, not a full characterization of every sampler. Predicting in advance which non-uniform distributions remain safe, for example by measuring the deep-layer coverage $\varrho_\ell$ of Proposition~\ref{prop:kfac} in a single forward pass, or by learning the anchor distribution to maximize it, is an interesting direction for future work.

\section{Computational Overhead}
\label{sec:appendix_flops}

We compare FLOPs and memory across four training configurations: vanilla GRPO/GSPO (no regularizer), GRPO+KL (a live frozen reference model evaluated on rollouts every step), and \namel{} (a fixed anchor set evaluated once). All numbers use context length $S=2{,}048$, per-device batch $B=8$, and $G=8$ rollouts per prompt, matching the production configuration.

\subsection{Setup}
Table~\ref{tab:arch_params} gives the architecture used for each scale, taken directly from each model's released config; all three Qwen3-VL models evaluated in this paper are dense, and all three use head\_dim$=128$ with $n_\text{kv}=8$ key/value heads (so the KV projection dimension $D_\text{kv} = n_\text{kv} \times \text{head\_dim} = 1{,}024$ is identical across scales, while the query/output projection dimension $D_q = n_\text{heads} \times \text{head\_dim}$ is not always equal to $H$: at 4B, $D_q=4{,}096 \neq H=2{,}560$). For a sequence of length $S$, per-layer FLOPs are
\begin{equation}
\mathcal{F}_\text{layer}(S) = 4SHD_q + 4SHD_\text{kv} + 4S^2D_q + 6SHF,
\end{equation}
(query/output projections, key/value projections, attention matmuls, SwiGLU FFN respectively), giving total forward FLOPs $\mathcal{F}_\text{fwd}(S) = L\cdot\mathcal{F}_\text{layer}(S) + 2SHV$, with a backward pass costing $\approx2\mathcal{F}_\text{fwd}$. At $S=2{,}048$ this gives $\mathcal{F}_\text{fwd} = 8.01$, $18.95$, and $33.47$ TFLOPs for the 2B, 4B, and 8B models respectively.

\begin{table}[hbt!]
\centering
\small
\caption{Qwen3-VL language-backbone architecture used for FLOP calculations (from each model's \texttt{config.json}). All three scales are dense.}
\label{tab:arch_params}
\begin{tabular}{lrrrrrr}
\toprule
Model & Layers & Hidden & $n_\text{heads}$ & $n_\text{kv}$ & FFN Dim & Params \\
\midrule
Qwen3-VL-2B & 28 & 2{,}048 & 16 & 8 & 6{,}144  & $\approx$2.1B \\
Qwen3-VL-4B & 36 & 2{,}560 & 32 & 8 & 9{,}728  & $\approx$4.0B \\
Qwen3-VL-8B & 36 & 4{,}096 & 32 & 8 & 12{,}288 & $\approx$7.6B \\
\bottomrule
\end{tabular}
\end{table}

\subsection{Per-Step Cost}
Every method's rollout/backprop cost is identical: $B$ rollouts forward plus backward, $\mathcal{F}_\text{model} = 3B\mathcal{F}_\text{fwd} = 24\,\mathcal{F}_\text{fwd}$. The two regularized methods differ only in what they add:
\begin{itemize}
\item \textbf{GRPO+KL} runs the frozen reference forward on the $B=8$ \emph{current rollouts} every step: $+8\,\mathcal{F}_\text{fwd}$, and must keep $\pi_\text{ref}$ resident on-GPU for the whole run. The KL penalty adds no backward cost of its own: the reference is frozen, and the KL gradient flows through the policy's log-probabilities on the same rollout tokens, so it adds no new tokens to the policy's forward and backward pass above, only an elementwise term at the output.
\item \textbf{\namel} runs the student forward \emph{and backward} on $B_A=8$ \emph{fixed anchors} every step, since the anchor loss is differentiated: $8\,\mathcal{F}_\text{fwd}$ forward plus $16\,\mathcal{F}_\text{fwd}$ backward, $+24\,\mathcal{F}_\text{fwd}$ in total. Unlike the KL term, this cost cannot be shared with the policy-gradient pass: forward and backward cost scale with the number of tokens processed, and the anchors are additional sequences beyond the rollouts. This holds whether the two losses are summed into a single backward call or not. The reference policy is only ever evaluated once, on the full anchor set, before training starts.
\end{itemize}
Per step, GRPO+KL costs $32\,\mathcal{F}_\text{fwd}$ ($1.33\times$ base RL) and \namel{} $48\,\mathcal{F}_\text{fwd}$ ($2\times$ base RL); Table~\ref{tab:flops} gives the resulting per-step cost at each scale. These counts treat rollout generation as a single forward pass per sequence; in practice generation is autoregressive and dominates wall-clock time, so the measured overhead is far smaller than the FLOP ratio suggests (Section~\ref{sec:wallclock}).

\begin{table}[hbt!]
\centering
\small
\caption{Per-step training FLOPs by method and scale.}
\label{tab:flops}
\begin{tabular}{lrrrr}
\toprule
Method & Mult. & 2B & 4B & 8B \\
\midrule
GRPO / GSPO & $24\times$ & 192 TF & 455 TF & 803 TF \\
GRPO+KL & $32\times$ & 256 TF & 606 TF & 1{,}071 TF \\
\namel{} (per-step) & $48\times$ & 384 TF & 910 TF & 1{,}607 TF \\
\bottomrule
\end{tabular}
\end{table}

\subsection{Trajectory-Level Cost: One-Time vs.\ Per-Step Reference Passes}
Per-step FLOPs hide where the two methods differ most: GRPO+KL's reference passes recur at \emph{every} step because rollouts change every step, while \namel's anchors are fixed, so its reference pass happens \emph{once}. Over a 600-step run, GRPO+KL evaluates the reference $600\times8=4{,}800$ times; \namel{} evaluates it $N=1{,}280$ times total, a $3.75\times$ reduction (Table~\ref{tab:ref_savings}). Including this one-time cost amortized over training, \namel's effective per-step overhead rises only slightly, to $50.13\,\mathcal{F}_\text{fwd}$ ($2.09\times$ base RL) at every scale, since the amortization multiplier depends only on $N/600$, not on $\mathcal{F}_\text{fwd}$ itself.

\begin{table}[hbt!]
\centering
\small
\caption{Cumulative reference-policy compute over a 600-step run.}
\label{tab:ref_savings}
\begin{tabular}{lrrrr}
\toprule
Method & Ref.\ passes & 2B & 4B & 8B \\
\midrule
GRPO+KL & 4{,}800 & 38.4 PF & 91.0 PF & 160.7 PF \\
\namel & 1{,}280 & 10.3 PF & 24.3 PF & 42.8 PF \\
\midrule
Reduction & $3.75\times$ & $-28.2$ PF & $-66.7$ PF & $-117.8$ PF \\
\bottomrule
\end{tabular}
\end{table}

\subsection{Memory}
Because $\pi_\text{ref}$ is never called after initialization, \namel{} unloads it from GPU memory entirely ($\approx$16GB/device saved at 8B, from $\approx$8B bf16 parameters), keeping only cached anchor features in host RAM. The cache stores one $d$-dimensional bf16 feature vector per anchor per timestep, $N \times T \times d \times 2$ bytes, where $d=H$ is the language backbone's hidden size: $\approx$10.7GB at 2B ($d{=}2{,}048$), $\approx$13.4GB at 4B ($d{=}2{,}560$), and $\approx$21.5GB at 8B ($d{=}4{,}096$) — in every case comfortably within standard host memory and never touching GPU VRAM. GRPO+KL must keep both policy and reference resident throughout training. Fixed-anchor functional regularization therefore needs far fewer reference passes and less GPU memory than reference-KL, which is why its measured wall-clock overhead is lower despite its higher per-step FLOP count (Section~\ref{sec:wallclock}).

\subsection{Measured Wall-Clock Time}
\label{sec:wallclock}
FLOP counts overstate the practical cost of \namel{}. We measured the average wall-clock time per training step with identical hardware and batch settings for every method (Table~\ref{tab:wallclock}). GSPO with \namel{} takes $156.0$\,s per step against $140.0$\,s for GSPO alone, an overhead of $11.4\%$. By comparison, GRPO+KL ($\beta{=}0.01$) takes $169.5$\,s per step against $123.5$\,s for GRPO, an overhead of $37.2\%$. Although \namel{} adds more FLOPs per step than GRPO+KL (Table~\ref{tab:flops}), its wall-clock overhead is about three times smaller. Two factors account for the gap. First, each step is dominated by autoregressive rollout generation, which \namel{} leaves untouched; its additional work is a single teacher-forced, fully parallel forward and backward pass over a small fixed batch of anchors. Second, \namel{} releases the reference policy after the one-time caching pass (Appendix~\ref{sec:appendix_flops}, Memory), whereas GRPO+KL must keep the reference resident on-GPU and run it on freshly generated rollouts at every step.

\begin{table}[hbt!]
\centering
\small
\caption{Measured wall-clock time per training step. Overhead is relative to the unregularized algorithm each regularizer is built on.}
\label{tab:wallclock}
\begin{tabular}{lcc}
\toprule
Method & Seconds / step & Overhead \\
\midrule
GRPO & 123.5 & - \\
GRPO+KL ($\beta{=}0.01$) & 169.5 & $+37.2\%$ \\
\midrule
GSPO & 140.0 & - \\
GSPO + \namel{} & 156.0 & $+11.4\%$ \\
\bottomrule
\end{tabular}
\end{table}

\section{Qualitative Examples}
\label{app:qualitative}

\begin{figure}[hbt!]
\centering
\includegraphics[width=0.95\linewidth]{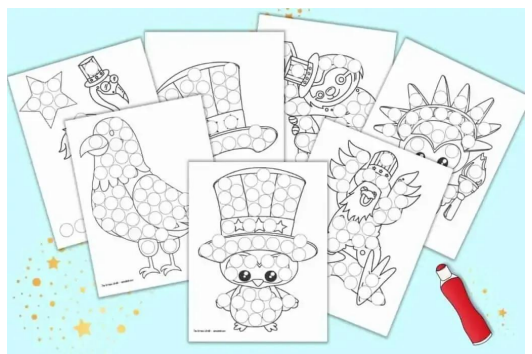}
\caption{Forgetting example 1: hallucinated instances.}
\label{fig:hallucination}
\end{figure}

\begin{figure}[hbt!]
\centering
\includegraphics[width=0.95\linewidth]{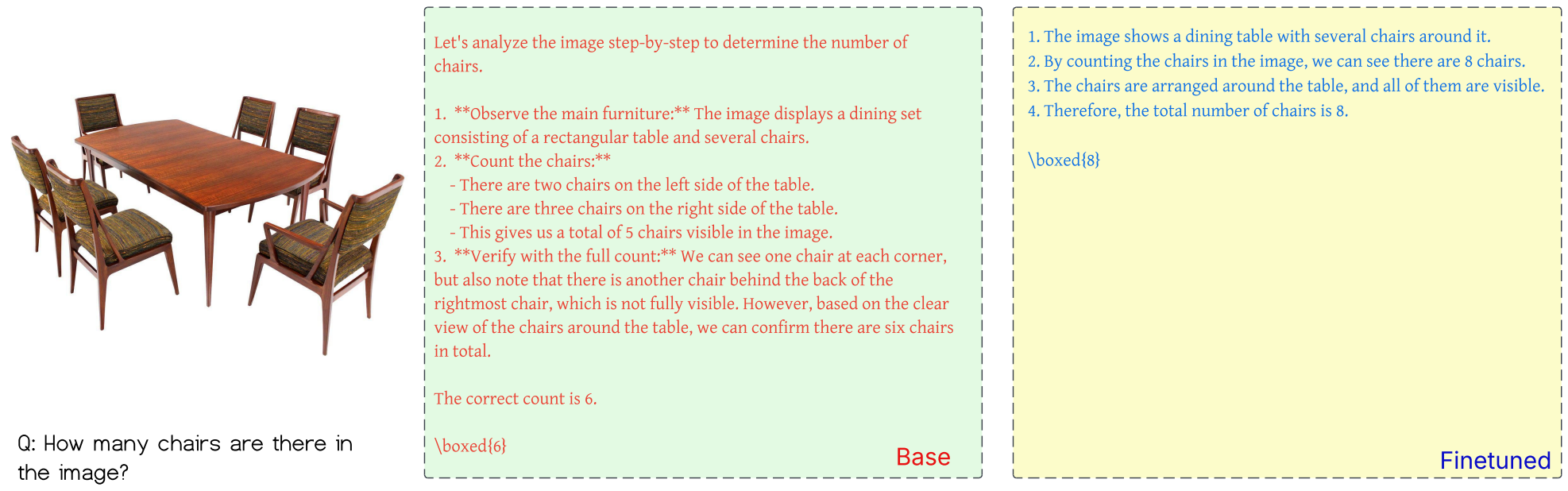}
\caption{Forgetting example 3: loss of verification.}
\label{fig:verification}
\end{figure}

Section~\ref{sec:qualitative} identifies three recurring failure modes behind the
CountBenchQA collapse: hallucinated instances, shortcut answers, and loss of
verification. Figures~\ref{fig:hallucination}-\ref{fig:verification} give one
representative base-vs-finetuned pair for each, drawn from unmitigated GRPO
post-training on Qwen3-VL-2B (MedBookVQA), the configuration exhibiting the
single most severe grounding collapse reported in Section~\ref{sec:results}
($-67.4\%$ worst-decile change, CountBenchQA falling from $82.5$ to $26.9$).
Across all three, the post-RL model preserves the base model's fluent,
step-by-step reasoning scaffold - the same enumerate-then-conclude structure,
the same boxed final answer - while the correspondence between the enumerated
steps and the actual image content degrades. Format is preserved; grounding is
not.

\paragraph{Hallucinated instances (Figure~\ref{fig:hallucination}).} Asked to
count the waterlilies in an image, the base model enumerates three: one in the
foreground, one middle-left, one upper-left, and correctly stops there,
answering $3$. The finetuned model performs the same enumeration but appends a
fourth position - ``upper right'' - that is not supported by the image,
arriving at $4$. The extra instance is not a wholesale
miscount: the model manufactures
a plausible-sounding spatial location for a non-existent object and folds it
into the same step-by-step scaffold used for the three real instances,
making the hallucination invisible from the reasoning trace alone.

\paragraph{Shortcut answers (Figure~\ref{fig:shortcut}).} Asked to count
coloring pages, the base model itemizes each sheet by its illustration (an owl
in a top hat, a bird, a star, a superhero, and two further pages it lists
identically - itself a minor duplication error, but one internal to an
otherwise page-by-page process) before concluding $7$. The finetuned model
answers $6$ directly, then generates a description - ``the pages include
various characters such as a star, bird, patriotic owl, character with a top
hat, superhero, and a sun-like figure'' - that is composed \emph{after} the
count is fixed rather than before it: exactly the pattern of asserting a
count in one sentence and back-filling a justifying description in the
next, rather than deriving the count from an enumerated pass over the image.

\paragraph{Loss of verification (Figure~\ref{fig:verification}).} Asked to
count chairs around a dining table, the base model counts five directly
visible chairs, then explicitly checks for occlusion - noting a further
chair ``behind the back of the rightmost chair, which is not fully visible''
- and revises its answer upward to $6$ only after that check. The
finetuned model skips this step entirely: it
counts by eye, asserts ``all of them are visible,'' and outputs $8$ with no
attempt to reconcile the claim of full visibility against the actual
occlusion in the image.

\bigskip
Taken together, the three examples show that single-domain RL post-training
does not degrade the vision-language model's ability to produce fluent,
well-formatted chain-of-thought - if anything, the finetuned traces are no
less articulate than the base model's - but selectively degrades the
visual-perceptual operations (instance verification, occlusion reasoning,
count-then-describe ordering) that the chain-of-thought is supposed to be
reporting on. This is consistent with the paper's broader claim that
forgetting here is a change in the model's induced output distribution on
untouched capabilities, not a change in surface fluency.

\begin{figure}[hbt!]
\centering
\includegraphics[width=0.95\linewidth]{figures/forgetting_ex2.pdf}
\caption{Forgetting example 2: shortcut answers.}
\label{fig:shortcut}
\end{figure}

\section{Hyperparameters}
\label{app:hparams}

For all methods, we follow a consistent \emph{tune-then-evaluate} protocol: we sweep over a
predefined hyperparameter grid, select the checkpoint that achieves the highest in-domain reward
on the target task, and report out-of-domain (OOD) generalization metrics exclusively for these
best-selected configurations. All RL training experiments were run for a fixed budget of 600 steps with a cosine-annealed learning rate. This ensures that all methods are compared at their peak in-domain
performance, making OOD forgetting a genuine measure of representational cost rather than a proxy
for insufficient training.

\textbf{Runs, selection, and reported uncertainty.}
For every method and every domain-scale cell, we sweep the hyperparameters listed below, giving between 3 and 18 configurations per method per cell; each configuration is a single 600-step training run. For each method we report the configuration with the highest in-domain accuracy and evaluate it on the full out-of-domain suite, so selection never uses retention. The $\pm$ values in all tables are bootstrap standard deviations over evaluation items, and therefore reflect the sampling uncertainty of each benchmark score rather than variation across training seeds; variation across seeds is measured separately in Appendix~\ref{app:seeds}.

\textbf{Learning Rate.}
All RL-based methods (GRPO, GSPO, and \namel{} variants) are trained with learning rates drawn from
$\text{lr} \in \{1 \times 10^{-6},\; 5 \times 10^{-6},\; 1 \times 10^{-5}\}$.

\textbf{Weight Decay and KL Penalty.}
For unregularized GRPO, we additionally sweep weight decay
$\lambda_{\mathrm{wd}} \in \{0,\; 0.1\}$. For GSPO and its variants, we use a weight decay of $0$ for all runs.  For GRPO with explicit KL regularization, we sweep a KL penalty coefficient $\beta_{\mathrm{KL}} \in \{0,\; 0.01, 0.005, 0.001\}$ added to the policy gradient objective. We report the $\beta=0$ ablation separately, and report the best learning run of the other $\beta$ experiments as GRPO~+~KL ablation.

\textbf{SFT and LoRA.}
Both SFT and LoRA are trained for $5$ epochs. For SFT, we sweep over learning rates $\text{lr} \in \{1 \times 10^{-5},\; 5 \times 10^{-6},\; 2 \times 10^{-6}\}$ for 8B models and $\{2 \times 10^{-5},\; 1 \times 10^{-5},\; 5 \times 10^{-6}\}$ for 4B and 2B models. For LoRA, the learning rate sweep is $\{1 \times 10^{-4},\; 5 \times 10^{-5},\; 2 \times 10^{-5}\}$.

\textbf{EWC.}
We sweep the Fisher regularization coefficient
$\lambda_{\mathrm{EWC}} \in \{0.05,\; 0.1,\; 0.4\}$,
where $0.4$ is the default proposed by~\citet{kirkpatrick2017overcoming}.
We consider both a vision-encoder-only Fisher scope and a full-model scope, and find that full-model scope is generally better suited for constraining behavior while vision-encoder-only Fisher scope gets roughly similar performance to the naive GSPO baseline.

\textbf{CPO.}
Our continual preference optimization variant splits the 600-step training budget into $k=3$
sequential subsets of 200 steps each, chaining checkpoints across subsets. After each subset,
we compute an importance mask over the top $10\%$ of parameters that drifted most from the
base model. In subsequent subsets, the trainer applies an L1 penalty on the masked parameters'
deviation from their values at the end of the previous subset, with a regularization
coefficient swept over $\lambda_{\mathrm{CPO}} \in \{1.0, 0.5, 0.1\}$. We sweep over the learning rate,
using the same grid as GSPO: $\text{lr} \in \{1 \times 10^{-6},\; 5 \times 10^{-6},\;
1 \times 10^{-5}\}$ for 2B, 4B models and $\{1 \times 10^{-6},\; 2 \times 10^{-6},\;
5 \times 10^{-6}\}$ for 8B models.

\textbf{\namel.}
We tune the anchor buffer size over
$N \in \{320,\; 640,\; 1280,\; 2560\}$ on the Navigation task with a 2B model, finding that
performance plateaus at $N = 1280$ (256 images per anchor distribution).
We therefore fix $N = 1280$ for all other tasks and model scales to balance
anchoring quality against compute cost.
The regularization strength is swept over $\lambda_{\mathrm{\namel}} \in \{0.01,\; 0.05,\; 0.1\}$.
We consider both \emph{real-image anchors} (held-out visual pretraining data) and \emph{Random-token anchors}, which we find to perform comparably while eliminating the need for stored pretraining data.

\textbf{Training Setup.}
All experiments are run on 8 GPUs (B40 or H100) using DeepSpeed distributed training.
For 2B and 4B models we use ZeRO Stage-2 which only uses data parallelism; for 8B models we use ZeRO Stage-3 to shard optimizer
states across devices, along with model and data. All models are trained in bf16 precision.

\textbf{RL methods (GRPO, GSPO, \namel, EWC, CPO).}
Each training step samples a group of $G=8$ independent completions per prompt, giving an effective rollout batch of
$8 \times 8 = 64$ sequences per gradient update. For \namel{} regularized runs, we draw a global batch of 64 anchors per gradient update step (distributed as 8 anchors per GPU across the 8 GPUs), adding only $\sim 11\%$ wall-clock overhead per step (Appendix~\ref{sec:wallclock}) without requiring a live reference model. We accumulate gradients over 8 steps,
yielding an effective global batch of 512 samples.
The rollout buffer is refreshed every 16 steps for 4B models and every 32 steps for 8B models
(\texttt{steps\_per\_generation}).
All RL runs train for a fixed budget of 600 gradient steps.
Maximum completion length is set to 2048 tokens for all tasks. We use a cosine learning rate schedule with 10\% linear warm-up, temperature $\tau=0.8$,
top-$p=1.0$, and no repetition penalty for the rollout generations.

\textbf{SFT and LoRA.}
Both methods use a cosine schedule with 3\% warm-up.
For 2B and 4B models, we set per device batch size as 4 with gradient accumulation of 2
(global batch 64); for 8B models, per device batch size is set as 2 with accumulation of
4 (global batch 64). LoRA uses per device batch size as 4 (2B,4B) or 2 (8B)
with a similar gradient accumulation strategy as SFT.

\section{Limitations}
\label{sec:limitations}

\textbf{What content-free anchors cannot protect.}
Theorem~\ref{thm:interference} splits forgetting into \emph{collateral} drift, which content-free anchors remove, and \emph{interference}, where the target task itself must move directions a capability relies on, which no capability-blind regularizer can prevent. Capabilities close to the training domain are therefore most at risk and need content-aware anchors (Proposition~\ref{prop:hybrid}) or a smaller target gain. We do not measure the interference component directly on production models. Separately, the embedding MMD penalizes drift only in expected token-embedding space: shifts within the null space of $K=\frac1d EE^\top$, such as redistributing probability among tokens with the same mean embedding, are not penalized (Appendix~\ref{sec:unified-template}). We regard this as a deliberate tolerance of paraphrase, but a capability that depends on such fine-grained token choices would not be protected by it.

\textbf{Pathways the anchors do not reach.}
Text-token anchors do not pass through the vision encoder or projector, so \namel{} constrains how the language model processes visual tokens, not how the encoder produces them (Corollary~\ref{cor:blocked}). In our runs this is where the forgetting arises: vision-encoder-only EWC performs roughly like unregularized GSPO (Appendix~\ref{app:hparams}). Capabilities whose degradation originates in the encoder itself would need the encoder frozen or image-side anchors added, which we leave untested.

\textbf{Large updates.}
Every regularizer that keeps the model close to a reference, whether KL, EWC, replay or anchoring, gives a weaker guarantee as the update grows, so this limit is not specific to \namel{}. \namel{} evaluates the exact anchor loss at every step rather than a frozen local approximation, and its any-step bound (Theorem~\ref{thm:transfer}(vi)) degrades through remainder factors that can be monitored during training; the anchor-side factor stays at most $1.08$ over 600 steps on Qwen3-VL-2B (Table~\ref{tab:trust-radius-data}). Much longer, multi-epoch or multi-stage training, or larger learning rates, may raise these factors. With a fixed reference, the guarantee bounds drift from the base model; refreshing the reference between stages would bound drift from the latest checkpoint instead.

\textbf{Assumptions behind the theory.}
The assumptions are stated in Appendix~\ref{sec:transfer}, and Table~\ref{tab:assumptions} lists which result uses each. The a-priori transfer bound (Theorem~\ref{thm:transfer}(ii)-(iv)) is second order, and our runs leave the conservative trust radius of Proposition~\ref{prop:trust-radius} early; the any-step bound (Theorem~\ref{thm:transfer}(vi)) is a certificate checked after training, since its capability-side remainder requires the capability's own inputs. The transfer coefficient is defined on the update subspace $S$, which is estimated from training displacements. Coverage at deeper layers relies on a Kronecker-factored approximation (Proposition~\ref{prop:kfac}) and guarantees only a finite, not small, transfer coefficient. Theorem~\ref{thm:content} caps what anchor content can add over random anchors, but does not show that this cap is small for a given curated set. The Gauss-Newton equivalence of Corollary~\ref{cor:cfa-is-generalized-ewc} holds exactly for uniform position weights; varentropy weighting is a heuristic that preserves positive semi-definiteness and the concentration results (Remark~\ref{rem:weighting}). Capability curvatures, and hence $\kappa_S$, are computable exactly only in our synthetic testbed (Appendix~\ref{sec:transfer-sim}); on production models we test the theory's predictions, not its quantities.

\textbf{Scope of the evaluation.}
Main experiments use Qwen3-VL (2B-8B) on three domains and a ten-benchmark suite; we do not evaluate models above 8B. The other VLM families (Section~\ref{sec:other-families}) are trained on one domain, with one untuned learning rate per training type and a subset of the baselines, and the text-only models (Appendix~\ref{app:text-only}) are small, at most a few billion parameters. The six additional benchmarks of Appendix~\ref{app:extended-ood} are evaluated on Navigation-2B only, and their ScreenSpot results are partly affected by output-format changes and truncation; capabilities outside these sixteen benchmarks remain unmeasured. Most configurations use one training run, with seed variation measured on Navigation-2B (Appendix~\ref{app:seeds}), and the reported $\pm$ values reflect evaluation noise over items, not training noise. We study single-domain post-training of up to 600 steps (GRPO, GSPO, SFT, LoRA), not sequential multi-task or long-context training; anchors are 2048 tokens long, so much longer positions are constrained only indirectly.

\textbf{Anchors and cost.}
Uniform sampling over the full vocabulary is the only tested sampler that keeps the coverage guarantee (up to the isotropy defect of Proposition~\ref{prop:minimax}), and restricted or reweighted vocabularies can degrade retention (Appendix~\ref{app:anchor-sampling}). Content-free anchors need a larger budget than curated ones before they overtake them (Figure~\ref{fig:results_anchor_size}). \namel{} adds an $11.4\%$ wall-clock overhead over GSPO, against $37.2\%$ for GRPO+KL (Appendix~\ref{sec:wallclock}), and its cached reference outputs grow with $N$, $T$ and the hidden size.

\textbf{What is preserved.}
\namel{} keeps the model close to its reference policy, including any undesirable behavior it has; it mitigates forgetting and is not an alignment method.
\end{document}